\documentclass[12pt,singlespace]{mitthesis}

\usepackage{lgrind}
\usepackage{cmap}
\usepackage[T1]{fontenc}
\usepackage{hyperref}
\hypersetup{ colorlinks, citecolor=blue, linkcolor=blue!80!black!80,urlcolor=blue}

\usepackage{multibib}
\usepackage{xcolor}
\newcites{pers}{Personal References}
\usepackage{graphicx}
\usepackage{subfigure}
\usepackage{booktabs} 

\usepackage{amssymb,amsmath,amsfonts,amscd,amstext,bbm, bm, enumerate, dsfont, mathtools}
\usepackage{enumitem}
\usepackage{microtype}
\usepackage{epstopdf}
\usepackage{wrapfig}
\usepackage{verbatim}
\usepackage{etoc}
\usepackage{adjustbox}
\usepackage{color,colortbl}

\usepackage{algorithm,algpseudocode}
\usepackage{verbatim}
\usepackage{tcolorbox}
\usepackage{import}
\usepackage{tikz,pgfplots}
\pgfplotsset{compat=1.17}
\usetikzlibrary{tikzmark,shapes}
\usetikzlibrary{matrix}
\usepgfplotslibrary{groupplots}
\usepackage{tabularx}
\usepackage{array}
\usepackage{csquotes}
\usepgfplotslibrary{statistics}
\newtheorem{definition}{Definition}
\newtheorem{assumption}{Assumption}

\newtheorem{theorem}{Theorem}

\newenvironment{Proof}{\par\noindent{\bf Proof.}}{\hfill$\square$}
\usepackage{multicol}
\usepackage{multirow}
\usepackage{url}

\usepackage{setspace}
\usepackage{xcolor}
\usepackage{microtype}
\usepackage{cmap}
\usepackage[final]{pdfpages}
\usepackage{etoolbox}
\usepackage{breakcites}

\DeclareMathOperator*{\argmin}{arg\,min}
\DeclareMathOperator{\Var}{Var}
\DeclareMathOperator{\nb}{nb}

\newcommand{\Loss}{\mathcal{L}}

\newcommand{\userS}{\mathcal{U}}
\newcommand{\itemS}{\mathcal{I}}
\newcommand{\vecU}{\mathbf{U}}
\newcommand{\vecI}{\mathbf{V}}
\newcommand{\prefu}{\renewcommand\arraystretch{.2} \begin{array}{c}
  {\succ} \\  \mbox{{\tiny {\it u}}}
  \end{array}\renewcommand\arraystretch{1ex}}
\newcommand{\posI}{\mathcal{I}^{+}}
\newcommand{\negI}{\mathcal{I}^{-}}

\newcommand{\NetF}{\textsc{Netflix}}
\newcommand{\RecS}{\textsc{RecSys'16}}
\newcommand{\Out}{\textsc{Outbrain}}
\newcommand{\ML}{\textsc{ML}}
\newcommand{\R}{\mathbb{R}}
\newcommand{\kasandr}{\textsc{Kasandr}}
\newcommand{\PANDOR}{\textsc{Pandor}}

\newcommand{\GCN}{\texttt{LightGCN}}
\newcommand{\SASR}{\texttt{SASRec}}

\newcommand{\MostPop}{\texttt{MostPop}}

\newcommand{\MF}{\texttt{MF}}
\newcommand{\GRU}{\texttt{GRU4Rec}}
\newcommand{\ProdVec}{\texttt{Prod2Vec}}
\newcommand{\BPR}{\texttt{BPR}}

\newcommand{\SO}{\texttt{SAROS}}
\newcommand{\batch}{\texttt{BPR$_b$}}

\newcommand{\caser}{\texttt{Caser}}
\newcommand{\mapk}{\texttt{MAP@K}}
\newcommand{\apk}{\texttt{AP@K}}
\newcommand{\mapfive}{\texttt{MAP@5}}
\newcommand{\mapten}{\texttt{MAP@10}}
\newcommand{\ndcgfive}{\texttt{NDCG@5}}
\newcommand{\ndcgten}{\texttt{NDCG@10}}
\newcommand{\ndcgk}{\texttt{NDCG@K}}
\newcommand{\dcgk}{\texttt{DCG@K}}
\newcommand{\idcgk}{\texttt{IDCG@K}}
\newcommand{\w}{\boldsymbol{\omega}}

\newcommand{\calI}{\mathcal{I}} 
\newcommand{\calS}{\mathcal{U}} 
\newcommand{\calB}{\mathcal{B}} 
\newcommand{\calL}{\mathcal{L}} 
\newcommand{\weight}{\omega}

\newcommand{\MOSAIC}{\texttt{MOSAIC}}

\newcommand{\bloc}{\mathcal B}

\usepackage{fancyhdr}
\AtBeginEnvironment{subappendices}{%
\counterwithin{figure}{section}
\counterwithin{table}{section}
}

\begin{document}

\title{\LARGE Large-Scale Sequential Learning for Recommender and Engineering Systems}

\author{Aleksandra Burashnikova}

\department{doctoral committee}

\degree{Doctor of Philosophy in Computer Science}

\degreemonth{May 1st, }
\degreeyear{2022}
\thesisdate{Ph.D. Candidate}


\chairman{Clement Fortin}{Chairman, Associate Provost, Dean of Education}

\maketitle

\cleardoublepage

\setcounter{savepage}{\thepage}

\cleardoublepage







	

	
	

	



	

 \begingroup
 \let\cleardoublepage\clearpage
 \chapter*{Abstract}
 In this thesis, we focus on the design of an automatic algorithms that provide personalized ranking by adapting to the current conditions. To demonstrate the empirical efficiency of the proposed approaches we investigate their  applications for decision making in recommender systems and energy systems domains. 

For the former, we propose novel algorithm called SAROS that take into account both kinds of feedback for learning over the sequence of interactions. The proposed approach consists in minimizing pairwise ranking loss over blocks constituted by a sequence of non-clicked items followed by the clicked one for each user. We also explore the influence of long memory on the accurateness of predictions. SAROS shows highly competitive and promising results based on quality metrics and also it turn out faster in terms of loss convergence than stochastic gradient descent and batch classical approaches.

Regarding power systems, we propose an algorithm for faulted lines detection based on focusing of misclassifications in lines close to the true event location. The proposed idea of taking into account the neighbour lines shows statistically significant results in comparison with the initial approach based on convolutional neural networks for faults detection in power grid.

 \chapter*{R\'esum\'e}
 
Dans cette thèse, nous nous intéressons à l'étude des algorithmes d'apprentissage qui fournissent un ordonnancement personnalisé en s'adaptant au contexte. Pour démontrer l'efficacité empirique des approches proposées, nous étudions leurs applications pour la prise de décision dans les domaines des systèmes de recommandation et des systèmes énergétiques.

Pour notre étude concernant les systèmes de recommandation, nous proposons un nouvel algorithme appelé SAROS qui prend en compte les deux types de retours utilisateurs pour apprendre leurs appétances. L'approche proposée consiste à minimiser la perte d'ordonnancement par paires de produits dans des blocs d'interactions constitués d'une séquence d'éléments avec un retour négatif (ou non cliqués) suivis d'un produit avec un retour positif pour chaque utilisateur. Nous explorons également l'influence de la mémoire longue sur la précision des prédictions. Notre algorithme montre des résultats hautement compétitifs et prometteurs basés sur des métriques de qualité et il s'avère également plus rapide en termes de convergence par rapport à  la descente de gradient stochastique et les approches classiques par lots.

En ce qui concerne les réseaux électriques, nous proposons un algorithme de détection des lignes défaillantes basé sur la focalisation des erreurs de classification dans les lignes proches du véritable emplacement de l'événement. L'idée proposée de prendre en compte les lignes voisines montre des résultats statistiquement significatifs par rapport à l'approche initiale basée sur les réseaux de neurones convolutifs pour la détection des défauts dans le réseau électrique.


 \pagestyle{plain}
\tableofcontents

 \listoftables
 \listoffigures

 \clearpage

\chapter{General Introduction}\label{ch:intro}
\chaptermark{Introduction}

The various scientific communities such as computer scientists and statisticians have been interested for many years in the problems of data analysis. The different currents from these communities focused on a set of specific issues and created scientific fields that quickly evolved independently. This is for example the case of Information Retrieval (IR), Information Extraction, or in the case of statisticians, data science, etc. In recent years, the field of data analysis has undergone a rapid evolution, with in particular the development of large-scale collections. The boundaries that had been drawn between the different traditional domains of data analysis are currently largely redrawn to create a large domain that we designate here by \textit{information access}. New problems appear, to which the various communities try to provide answers by adapting the existing tools, or by developing new tools. In particular, it has become important to be able to process huge amounts of data, to provide diversified solutions to new user demands, and to automate the tools that make it possible to exploit textual or image information.

More recently, the rapid development of techniques for the acquisition and storage of digital information has favored the explosion of the quantities of information to be processed, but also the diversity of their content. Thus, information to be processed takes forms as diverse as sequences of interactions, textual documents, images, music videos or even music. User needs have also evolved. Information systems must not only help them find the information they are looking for, but also advise them or make new suggestions. This is the main purpose of recommendation systems, which suggest to their users items likely to interest them: books, films, music albums, etc.

Machine learning offers a range of tools to move in these directions. It is within this framework that our work is situated, which aims to explore the potential of learning techniques to meet the needs of users and to detect fault lines in energy systems. In the case of textual information retrieval, for example, machine learning models are based on the assumption that it is possible to perform many textual information processing tasks by fairly crude analyzes of the text. Thus, any learning algorithm works from an initially known and fixed data representation. It is common to pre-process the data in order to modify this initial representation. The learning algorithm is then used on the new representation obtained. Learning about this new representation has several advantages: gains in algorithmic complexity and memory space, as well as the possibility of interpreting or visualizing the data. On the other hand, the influence of the new representation on the performance in prediction is more difficult to analyze. In the ideal case, we of course want the new representation to improve the prediction performance of the learning algorithms. In recommender systems for example, the new representation of users and items should make it possible to order higher  items that are of the interest of  users with respect to the others using the dot product in the latent learned space. In supervised learning, the new representation of the data should make it possible to make fewer prediction errors. The choice of the new representation, and therefore the choice of the method used to modify the initial representation, thus seems essential in learning.

\section{Contributions}

In this thesis we first propose to learn the palatability of users for items by exploiting their sequences of interactions using ranking models that take into account both positive and negative feedback. Most of the state-of-the-art systems consider only the items in the sequence of interactions for which a user has provided positive feedback. The incentive of using not only positive feedback is that it difficult to understand which items a user really likes and determine the characteristics of his or her consumption/action based only on the information about the clicks/purchases/likes/views etc. We suggest to avoid this problem and to increase the quality of the ranking by considering also negative interactions of user with the system. The proposed approach hence constructs a ranking model by taking user' negative and positive sequence of interactions.  

We tackle this problem from a learning to rank perspective, which involves sorting instances in relation to a demand. In any case, we are all confronted with task ranking in our everyday lives. We make decisions all of the time by intuitively constructing a scale of preferences for ourselves, based on which we choose one instance over another. For example, we may go to the automobile service for maintenance and then decide, based on the outcome, whether we would repair everything the service personnel suggests or only the most urgent problems at the time, given our budget. There are many different types of ranking systems that we encounter, such as document search engines or recommender systems.  

Companies may offer individualized and relevant suggestions using a properly set ranking model. Because a consumer spends the least amount of time searching for the right things and getting what he wants, his loyalty to the platform increases when he spends the least amount of time searching for the right things and getting what he wants with their help, the search time for the necessary items is reduced, and the likelihood of performing related targeted actions increases.

At the same time, development of a personalized large-scale ranking system is a serious and complex task. Formally, as in any supervised machine learning task, we need to build a function that fit the data in the best way. The input data for the training such function are the features of the system. For example, in case of fault detection problem in power systems, voltage information could be used to detect a location of contingencies, if any exists. Considering recommender system task, its possible to apply the user description or item descriptions to identify relevant items.

The important factors that affect the quality of ranking are how to take into account the features extracted for training. That is why its crucial to define what kind of feedback or characteristics of system to use. Another key point for building relevant ranking model is time. To provide fast predictions for ranking especially for predicting abnormal/dangerous behavior of the system is critical.

The detection of faults in power networks is the second subject we looked at in this thesis. In this scenario, we show that by utilizing grid's topology, significant characteristics for forecasting faults in its line may be deduced.

In resume the applications we studied in this thesis are
\begin{itemize}
    \item Recommender systems: 
    We proposed an algorithm that learns user and item representation over time while taking into account users' negative and positive feedback. As a result, the suggested method constructs blocks over the input sequence of feedbacks, which is composed of a series of negative items followed by positive items. We proved the convergence of the algorithm in the general case of non-convex loss functions and showed its efficiency compared to the state-of-the-art models over six large-scale benchmarks. A hybrid technique was also presented to speed up the algorithm in practice, including pre-filtering of input users.

    \item Power grids: We also considered the problem of faults detection in  power grids. The idea to improve the quality of predictions lies on taking into account the structure or the topology of the power grid. 
\end{itemize}

\section{Thesis Structure}

This thesis is organized in two parts. In the first part, we present state-of-the-art  frameworks and approaches related to our study. In the second part, we present our contributions.

\noindent The first part of this thesis consists of two chapters. 
\begin{itemize}
    \item[$\ast$] In \textbf{chapter \ref{ch:0}}, we present the main statistical supervised learning frameworks which are classification and ranking. For each learning framework, we present the important concepts and algorithms that will be useful to us in the second part of the thesis. 
    \item[$\ast$] In \textbf{chapter \ref{ch:1}}, we present recent approaches in recommender systems.
\end{itemize}

\noindent The second part of this thesis consists of three chapters. 
\begin{itemize}
    \item[$\ast$] In \textbf{chapter \ref{ch:2}}, we present \SO$_{}$ which is a sequential ranking algorithm for recommendation. Based on the assumption that users are shown a set of items sequentially, and that positive feedback convey relevant information for the problem in hand, the proposed algorithm updates the weights of a scoring function whenever an active user interacts with the system, by clicking on a shown item. We prove that these sequential updates of the weights converge to the global minimal of a convex surrogate ranking loss estimated over the total set of users who interacted with the system.  \item[$\ast$] In \textbf{chapter \ref{ch:3}}, we propose a unified framework for convergence analysis of \SO{}, in the general case of non-convex ranking losses. Furthermore, we study the effect of non-stationarities and memory in the learnability of a sequential recommender system that exploits user's implicit feedback.
    \item[$\ast$] In \textbf{chapter \ref{ch:5}} we present our work for the problem of fault detection in power grids. The main idea is first to characterize a power grid by exploiting its topology then to learn a prediction function by minimizing a loss where the errors that are farther away from the true location are penalized more than errors that are nearer to the true location. This is done by considering additional terms in the loss function that take into account the neighbours of the faulted line. Finally, using the statistical Mann–Whitney U-test we show the efficiency of the proposed approach.
\end{itemize}

\noindent Finally, in \textbf{chapter \ref{ch:6}} we conclude our work and present directions for future work.

\section{Corresponding articles}

The contributions of this manuscript are based on the following articles, prepared in scope of the research made during this Ph.D.

\begin{itemize}
    \item[] \textbf{Chapter \ref{ch:2}} is based on the paper \citepers{Burashnikova19} published at the \textit{European Conference in Machine Learning  and Principles and Practice of Knowledge Discovery in Databases} (ECML-PKDD 2019)
    
    In this paper, we proposed a theoretically founded sequential strategy for training large-scale Recommender Systems (RS) over implicit feedback mainly in the form of clicks. 
    The proposed approach consists in minimizing pairwise ranking loss over blocks of consecutive items constituted by a sequence of non-clicked items followed by a clicked one for each user. 
    Parameter updates are discarded if for a given user the number of sequential blocks is below or above some given thresholds estimated over the distribution of the number of blocks in the training set.
    This is to prevent from updating the parameters for an abnormally high number of clicks over some targeted items, mainly due to bots; or very few user interactions. Both scenarios affect the decision of RS and imply a shift over the distribution of items that are shown to the users. We provide a proof of convergence of the algorithm to the minimizer of the ranking loss,  in the case where the latter is convex.  Furthermore, experimental results on five large-scale collections demonstrate the efficiency of the proposed algorithm concerning the state-of-the-art approaches, both regarding different ranking measures and computation time.
    \item[] \textbf{Chapter \ref{ch:3}}, is based on two papers published respectively in \textit{Journal of Artificial Intelligence Research} (JAIR 2022) \citepers{DBLP:journals/jair/BurashnikovaMCL21} and the \textit{European Confernence in Information Retrieval} (ECIR 2022) \citepers{DBLP:journals/corr/abs-2112-02242}. 
    
    The journal paper is a continuation of the paper \citepers{Burashnikova19}. Here, additionally to the gradient-based strategy proposed in \SO{}, we present the momentum method for updating the parameters. Furthermore, we provide a convergence analysis of both algorithms for the general case, when ranking loss is non-convex, whereas in \citepers{Burashnikova19} we made it just for the convex loss function. The set of benchmarks also was extended by \RecS{}, that is a fairly large dataset and completely satisfy the task we are solving. The set of baselines algorithms also was increased by powerful graph-neural network based approach for impartial comparison of our algorithm with "fresh" state-of-the-arts. 
    
     In \citepers{DBLP:journals/corr/abs-2112-02242} we studied the effect of long memory over the user interactions in large-scale recommender systems. In essence, the paper proposes the idea of filtering the training data based on the concept of memory. Our finding led to an improvement of the empirical results, that confirmed our idea about the redundancy of information in the input data affecting the learning process. 
    
    \item[] \textbf{Chapter \ref{ch:5}} is supposed to be submitted at IEEE Control Systems Letters soon.
    
    Climate change increases the number of extreme weather events (wind and snowstorms, heavy rains, wildfires) that compromise power system reliability and lead to multiple equipment failures. Real-time and accurate detecting of potential line failures is the first step to mitigating the extreme weather impact, followed by activating emergency controls. Power balance equations non-linearity, increased uncertainty in renewable generation, and lack of grid observability compromise the efficiency of traditional data-driven failure detection methods. At the same time, modern problem-oblivious machine learning methods based on neural networks require a large amount of data to detect an accident, especially in a time-changing environment. In this paper, we propose a Topology-Aware Line failure Detector (TALD) that leverages grid topology information to reduce sample and time complexities and improve localization accuracy. Finally, we illustrate superior empirical performance of our approach compared to state-of-the-art method over various IEEE test cases.  
    
\end{itemize}

~\\~\\~\\~\\~\\~\\~\\~\\~\\~\\~\\~\\~\\
\addcontentsline{toc}{chapter}{Part I: State-of-the-art}
\begin{center}{\Huge \textsc{\underline{Part I}}\\~\\ \textsc{State-of-the-art}}\end{center}\normalsize
\chapter{Supervised Learning}
\label{ch:0}
\section{Introduction}

In this chapter, we present the two main frameworks in supervised learning which are classification and ranking. For each of these two frameworks, we present the important concepts and the algorithms that will be useful to us for the rest of our work.

In supervised learning, the goal is to learn the probabilistic relation (or joint distribution)  between the examples (mostly in vector form) $x\in \mathcal{X}$ and their outputs $y\in~\mathcal{Y}$. This training is done using a training set which contains labeled examples  $\{(x_i,y_i) \mid i=1,...,m \}$ that are supposed to be identically and independently distributed with respect to a joint probability $\mathcal D_{\mathcal{X},\mathcal{Y}}$. In classification and ranking, the output set $\mathcal {Y}$ is  discrete and they have two distinct goals. In classification, the aim is to predict the class label of an example, while in ranking the aim is to rank examples with respect to their outputs.  

We note another important framework in supervised learning that is not covered here and which is regression. The main difference between regression and the two aforementioned frameworks is that the output set is, in the case of regression, continuous (i.e. $\mathcal {Y} \subset \mathbbm{R}$).

Currently, a large number of algorithms for supervised learning have been developed for solving both classification and ranking problems, each of them has its own strengths and weaknesses. That goes without saying that the most of the recent proposed state-of-the-art approaches are based on neural networks.

This chapter is composed of three sections. We start by briefly presenting the supervised learning theory in section \ref{Sec:AppSup}. Following that, we present classification in section \ref{Sec:classif}, and then ranking in section \ref{sec:ranking}. Note that this chapter is not an exhaustive description of the algorithms developed in these two frameworks. We therefore focus our presentation on the concepts and algorithms that we will need in our contributions (see part II).

\section{A brief presentation of the learning theory}
\label{Sec:AppSup}
A supervised learning algorithm learns a prediction function from a set of examples, called training set. Each example is composed of a pair \textit{(observation, output)}. The goal of learning is to find a prediction function able of predicting the outputs associated with new examples, i.e. examples that do not belong to the training set.

In practice, a loss function measures the (dis)agreement between prediction and desired output (also called \textit{label}). The smaller the error, the better the prediction. Thus, the learning algorithm chooses the prediction function that minimizes the average error on the training examples, called \textit{empirical risk}. This is the \textit{Empirical Risk Minimization} (ERM) principle. By minimizing the empirical risk, we hope that the prediction function will have a low \textit{generalization error}, i.e it will make few errors on average on new examples. The underlying assumption is that the new examples are identical, in one way or another, to the training examples that were used to find the prediction function. The study of the link between empirical error and error in generalization is at the heart of the theory of statistical learning \cite{Vapnik2000}. The main result of this study is that learning is a compromise between a low empirical error and a high complexity of the class of functions where the prediction function is to be found. This is called the \textit{Structural Risk Minimization} (SRM) principle. In the following we will describe in detail these notions.

\paragraph*{Definitions and notations} We begin by giving some definitions and notations that we will use in the remainder of this thesis. An example is a pair consisting of an observation and a desired output. Observations have a numerical representation in a vector space $\mathcal X$, typically $\mathcal X \subset \mathbb{R}^d $ for fixed $d$. The response will be called the desired output, and it is assumed to be part of an output set $\mathcal Y$.  A pair $(x, y)$ will designate an element of $ \mathcal X \times\mathcal Y $.

\paragraph*{Central Assumption} The fundamental assumption of statistical learning theory is that all examples are independently and identically distributed (i.i.d.) by a fixed but unknown probability distribution $ \mathcal D$ . Thus for any set $S$, the examples $(x_i, y_i)\in S$ are generated i.i.d; according to $ \mathcal D $. We then say that $S$ is a  i.i.d. sample following $ \mathcal D$. Informally, this hypothesis defines the notion of representativeness of a training set
or test in relation to the problem: the training examples, as well as
future observations and their desired output, come from a given source.

\paragraph*{Loss functions} The second fundamental notion in learning is the notion of error, also called risk or loss. Given a prediction function $f$, the agreement between the prediction $f(x)$ and the desired output $y$ for a pair $(x, y)$ is measured using a function $\ell_c: \mathcal Y\times \mathcal Y \rightarrow \mathbb R_+ $. Intuitively, $\ell_c(f(x), y)$ measures the similarity between the
predicted and the desired output. It is therefore generally a distance over the set of outputs $ \mathcal Y$. In classification, the error generally considered is the 0/1 loss:
\begin{align*}
\ell_c(f(x), y) = \mathbbm{1}_{f(x) \neq y}.
\end{align*}

Where $\mathbbm{1}_{\pi}$ is $1$ if the predicate $\pi$ is true and $0$ otherwise. In other words the loss of a prediction error on the label of an example $x$ is worth $1$.

In bipartite ranking, which consists in assigning a higher score to a relevant observation $x$ (i.e. having a positive output $y=+1$) than to an irrelevant one (i.e. having a negative output $y=-1$), a classical ranking loss $\ell_r:(\mathcal Y\times \mathcal Y)^2 \rightarrow \mathbb R_+$ is to count an error when the ordering induced by a scoring function $f:\mathcal{X}\rightarrow \mathbb{R}$ is reversed. Hence 
\begin{align*}
\ell_r((f(x), y),(f(x'),y') = \mathbbm{1}_{(y-y')(f(x)-f(x')) \leq 0}.
\end{align*}

\paragraph*{Generalization error and empirical error} We are now able to give the definition of the error associated with a prediction function $f$ on all examples $ (x,y)$ from $ \left( \mathcal X \times \mathcal Y \right) $. This quantity is called generalization error which in the case of classification can be written as:

\begin{equation}
 \label{eq:GenError}
\mathcal{L}( f ) = \mathbb E_{ \mathcal D } \left[ \ell(f(x),y ) \right] = \int_{\mathcal X \times \mathcal Y} \ell( f(x),y ) d\mathcal D( x,y) 
\end{equation}

The function $f$ that is of interest  is the one that makes the fewest prediction errors on new examples, it is therefore the one that minimizes $ \mathcal{L}( f )$. However, as the probability distribution $ \mathcal D $ is unknown, this error in generalization cannot be directly estimated. \cite{Vapnik2000} has shown that the search for the function $f$ can be done in a consistent way by optimizing the average error of $f$ on a training set $S=\left((x_i,y_i)\right)_{1\leq i \leq m}$. This quantity is an unbiased estimator of generalization error and is commonly called the empirical risk of $f$ on $S$:

\begin{equation}
 \label{eq:EmpError}
\hat{\mathcal{L}}_m( f,S ) = \frac{1}{m} \sum_{i=1}^m \ell( f(x_i),y_i ) 
\end{equation}
Vapnik's ERM principle as well as the concepts mentioned above will be explained in the following paragraphs.

\paragraph*{Learning algorithm and ERM principle} A learning algorithm takes as input a training set $S$, and returns a prediction function $ f_S: \mathcal X \rightarrow \mathcal Y $ . In formal terms, a learning algorithm is a function $\mathcal{A}$ that looks for the function $ f_S $ inside a set of functions $\mathcal{F}$; called a class of functions. Intuitively, the ERM algorithm is understood as follows. If the training examples contained in $S$ are sufficiently representative of the distribution $ \mathcal D$, then (under certain conditions to be specified) the empirical error $ \hat{\mathcal{L}}_m( f,S ) $ is a good estimate of the generalization error  $ \mathcal{L}( f ) $. To minimize the generalization error, we will therefore minimize the empirical error on a given training set $S$. Given an error function $\ell:\mathcal{Y}\times\mathcal{Y}\rightarrow\mathbb{R}^+$, a training set $S$ containing $m$ examples and a class of functions $\mathcal F$, the ERM principle returns then the function $f_S$ verifying:

\begin{equation}
\label{eq:ERM}
f_S = \textrm{arg} \min_{f \in \mathcal F} \frac{1}{m} \sum_{ (x_i, y_i) \in S} \ell( f(x_i),y_i ) 
\end{equation}

\paragraph*{Generalization and consistency of a learning algorithm}

Let us underline that the minimization of the empirical error is not an end in itself, what interests us being the minimization of  generalization error. Thus, the ERM algorithm is of no use to us if the function learned $f_S$ has a low empirical error and a high generalization error. We will therefore expect from the ERM algorithm that it \textit{generalizes}, ie that the empirical error of $f_S$ is a good estimator of its  generalization error. If this property of \textit{generalization} holds, then we know that if ERM returns the low empirical error function $f_S$, then its generalization error will probably be low too.

Let us also insist on the fact that the ERM algorithm works in a known and fixed space of functions $\mathcal F$. The functions considered for the search for the lowest generalization error are elements of $\mathcal F$. Thus, a second naturally desirable property of the ERM algorithm is that it eventually finds the best function of $\mathcal F$ (for error in generalization) provided it has enough examples to learn from. This property is called the \textit{consistency}.

For learning to have meaning, the ERM algorithm must therefore verify the two previous properties. However, studies show that generalization and consistency are closely linked to the notion of complexity of the class of functions $\mathcal F$ considered.

\paragraph*{Overfitting and complexity of a class of functions} Let us first focus on the generalization property of the ERM algorithm. For certain classes of functions $\mathcal F$ whose empirical error of the learned function $ f_S $ is not a good estimator of its  generalization error. We guess that it is better to avoid that the learned function is too \textit{complex} compared to the training samples. Indeed it is easy to find a  function having a null empirical loss on a training set, and arbitrarily a high generalization error.  This phenomenon is called \textit{overfitting} and is illustrated in figure \ref{fig:Overfitting}.

\begin{figure}[H]
\centering
\includegraphics[width=.6\textwidth]{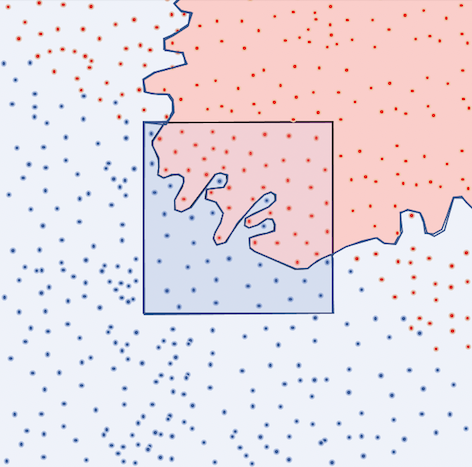}
\caption[Overfitting in classification]{Illustration of the overfitting phenomena on a classification problem where the aim is to separate between blue and red points. A complex model will fit exactly the data on the training set (square middle) by having an empirical error equal to $0$, but on other points outside the square (test points) it does a lot of mistake.}
\label{fig:Overfitting}
\end{figure}

So we want the ERM algorithm to learn simple functions. A way to impose simplicity is to constrain the class of functions $\mathcal F$ to contain only simple functions (the notion of simplicity remains to be defined). By doing so, it is possible to show that for ERM, the properties of generalization and consistency are equivalent: by limiting the complexity of the class of functions $\mathcal F$, we therefore guarantee the generalization and the consistency of the ERM algorithm. \\

On the other hand, if $ \mathcal F $ is too simple compared to the distribution $ \mathcal D $, then the learned function will probably not have good performance in generalization. Both its empirical error and its error in generalization will be high. We thus see that the choice of the space of hypotheses $\mathcal F$ is crucial: it must be neither too complex to avoid the problem of overfitting, nor too simple in order to avoid the problem of \textit{underfitting} and to achieve good performance in generalization anyway. This trade-off between low empirical error and a complex class of functions, also known as the \textit{bias-variance trade-off}, is fundamental in machine learning. This tradeoff is illustrated in figure \ref{fig:Tradeoff}.

\begin{figure}[H]
\centering
\begin{tikzpicture}
\begin{axis}%
[
    xmin=0,
    xmax=5,
    axis x line=bottom,
    ymax=3,
    xmax=6,
    xtick distance = 2,
    xtick={7,8},
    ytick={-7,-6},
    axis y line=middle,
    samples=100,
    legend style={at={(1,0.9)}}     
]
\addplot[domain=0.02:5, blue, ultra thick,samples=500] {1.3/(x)};
\addplot[domain=0.02:5, blue, ultra thick,samples=500] {(0.78*x-1.8)*(0.78*x-1.8)+1};
\end{axis}
\node at (5.5,2.5) (A) {true risk};
\node at (-1.8,1) (C) {empirical risk};
\node at (7.5,-0.5) (E) {Complexity};
\node at (-0.8,5.8) (F) {Error};
\node at (3.8,3) (B) {};
\node at (1.2,2) (D) {};
\draw [->] (A) -- (B);
\draw [->] (C) -- (D);
\node at (2.6,0.5) (G) {};
\node at (2.6,-0.5) (H) {};
\draw [-] (G) -- (H);
\node [text width=1cm] at (2.6,-1)  {Best Model};
\node at (2.7,-0.3) (I) {};
\node at (5,-0.3) (J) {};
\draw [->] (I) -- (J);
\node at (4.8,-0.7)  {overfitting};
\node at (2.3,-0.3) (K) {};
\node at (0.2,-0.3) (L) {};
\draw [->] (K) -- (L);
\node at (0,-0.7)  {underfitting};

\end{tikzpicture}
\caption[The phenomenon of overfitting]{The prediction error is displayed with regard to the complexity of the class of hypotheses in this illustration of the bias-variance trade-off. Both the generalization error and the empirical error are large for a class of hypotheses with little complexity. This is referred to as underfitting. When the complexity of the class of functions grows, both the empirical and true errors decline until a point at which the generalization error starts to increase while the empirical error continues to fall. This is referred to as overfitting. On the class of hypotheses with the lowest generalization error, the best model may be identified.}
\label{fig:Tradeoff}
\end{figure}
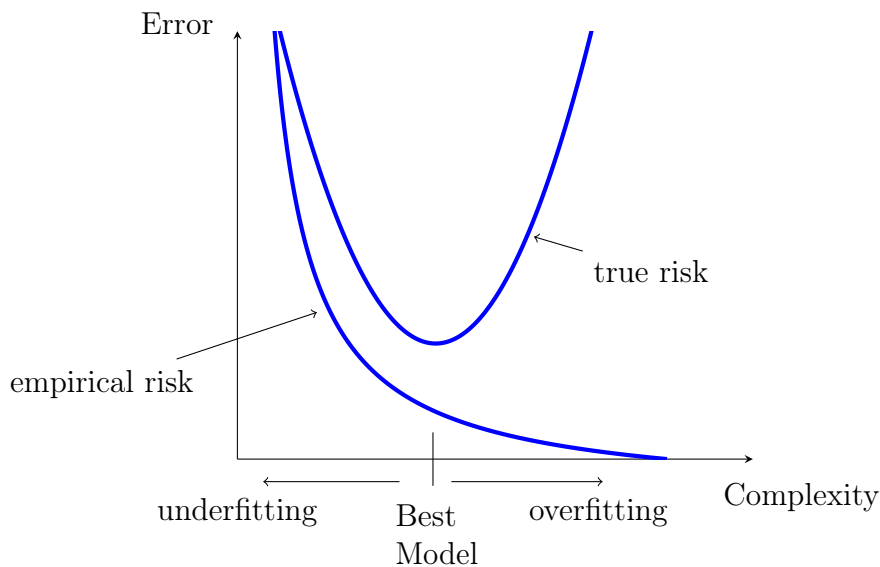

\paragraph*{VC Dimension} We now know that in learning it is crucial to be able to limit the complexity of the class of functions considered. For this, we must first define a way to define this complexity. In the case of binary classification, a fundamental measure of the generalization capacity developed by \cite{Vapnik2000} is the Vapnik-Chervonenkis dimension; or the VC dimension in short.

Let $ \mathcal F $ be a class of functions from $\mathcal X$ into $\mathcal Y = \{ -1, 1 \}$ and $X = \left( x_1, \ldots, x_m \right) $ a set observations in $\mathcal X$. Consider
\[
\mathcal S = \{ \left( ( x_1, y_1 ), \ldots, ( x_m, y_m ) \right) | ( y_1, \ldots, y_m ) \in \mathcal Y^m \} 
\]

in other terms $ \mathcal S $ is the set of observations with all possible labelings. The class of functions $ \mathcal F $ \textit{shatters} the set of observations $ X $ if whatever the set of examples $ S \in (\mathcal X\times\mathcal{Y})^m $, there is a classifier $ f \in \ mathcal F$ able to correctly classify all examples of $S$. The VC dimension of $ \mathcal F$ is the maximum number of points such that the function class can generate all possible classifications on this set of points. We then say that the set is \textit{shattered} by $\mathcal F$. The notion of complexity of a class of functions defined by its dimension VC is therefore linked to this notion of \textit{shattering}: the more a class of functions is capable of \textit{shattering} a large number of points, the more complex it is.

\paragraph*{Rademacher complexity} Another classical measure of the complexity of a class of functions is the Rademacher complexity \cite{BartlettM02}. This measure estimates how well a class of functions $\mathcal{F}=\{f:\mathcal{X}\rightarrow \{-1,+1\}\}$ can learn over a randomly noisy training set. Consider $\boldsymbol{\sigma}=\{\sigma_1,\ldots,\sigma_m\}$ a set of $m$ binary random variables where each $\sigma_i\in\boldsymbol{\sigma}$, called the Rademacher variable, takes a value $-1$ or $+1$ with probability $\frac{1}{2}$; i.e. $\forall i\in\{1,\ldots,m\}; \mathbb{P}(\sigma_i=-1)=\mathbb{P}(\sigma_i=-1)=\frac{1}{2}$. Then the empirical Rademacher complexity of $\mathcal{F}$ over a training set $S=(x_i,y_i)_{1\leq i\leq m}$ of size $m$ is defined as:
\begin{equation*}
    \hat{\mathfrak{F}}_m(\mathcal{F},S)=\frac{2}{m}\mathbb{E}_{\boldsymbol{\sigma}}\left[\mathop{sup}_{f\in\mathcal{F}}\left|\sum_{i=1}^m\sigma_i f(x_i)\right|\right],
\end{equation*}
Hence the higher the Rademacher complexity, the higher the ability of the class of function $\mathcal{F}$ to fit random (Rademacher) noise. 
The corresponding Rademacher complexity is then defined as 
\begin{equation*}
\mathfrak{F}_m(\mathcal{F})=\mathbb{E}_{S}[\hat{\mathfrak{F}}_m(\mathcal{F},S)].
\end{equation*}
The main difference between VC diemension and the Rademacher complexity is that the latter can be easily upper-bounded for some class of functions.

\paragraph*{Generalization bounds}
The study of the relationship between empirical error, error in generalization and complexity of the class of functions is at the heart of the  statistical learning theory. Most of these works take the form of probabilistic error bounds providing an upperbound of the generalization error that holds with high probability with respect to the empirical error, the complexity of the considered class of  an some residual term that controls the precision of the bound; as the following Rademacher generalization bound.
\begin{theorem}[Generalization bound \cite{BartlettM02}]
\label{thm:RadGenBound}

Let $\mathcal X\in\mathbb{R}^d$ be a vectorial space and  $\mathcal Y=\{-1,+1\}$ an output space. Suppose that the pairs of examples $(x,y)\in\mathcal X\times\mathcal Y$ are generated i.i.d. with respect to the distribution probability $\mathcal D$. Let $\mathcal{F}$ be a class of functions having values in  $\mathcal Y$ and $\ell:\mathcal Y\times \mathcal Y\rightarrow [0,1]$ a given instantaneous loss. Then for all $\delta\in ]0,1]$, we have with probability at least  $1-\delta$ the following inequality :
\begin{equation}
\label{eq:BrnGenRado}
\forall f\in \mathcal{F}, \Loss(f)\leq \hat{\mathcal{L}}_m( f,S )+\mathfrak{F}_m(\ell\circ\mathcal{F})+\sqrt{\frac{\ln\frac{1}{\delta}}{2m}}
\end{equation}
Using the same steps we can also show that with probability at least $1-\delta$
\begin{equation}
\label{eq:BrnGenRadoEmp}
 \Loss(f)\leq \hat{\mathcal{L}}_m( f,S )+\hat{\mathfrak{F}}_m(\ell\circ \mathcal{F},S)+3\sqrt{\frac{\ln\frac{2}{\delta}}{2m}}
\end{equation}
\end{theorem}
Where $\ell\circ \mathcal{F}=\{(x,y)\mapsto \ell(f(x),y)\mid f\in \mathcal{F} \}$.  It is therefore clear that to have a low risk, the two terms on the right of these inequalities \eqref{eq:BrnGenRado} or \eqref{eq:BrnGenRadoEmp} must be low: the empirical error which depends on the prediction function $f$, and the second term which depends on the complexity of the class of functions $ \mathcal F $. To have a theoretical guarantee on the error in generalization, it is therefore not only necessary to minimize the empirical error, it is also necessary to choose a class of functions that is not too complex (having a high complexity term). But it should not be too simple, otherwise the empirical error will be high. We therefore find the bias-variance trade-off, which we have already highlighted previously. We also mention some theoretical results in multi-class (extreme) classification establishing state-of-the-art bounds~\cite{maximov2016tight,yin2019rademacher}. Another line of research proposes reduction from multi-class to binary classification~\cite{joshi2017aggressive,rifkin2004defense}. We also mention a few extensive surveys on semi-supervised classification~\cite{amini2022self,van2020survey,maximov2018rademacher} and co-training~\cite{amini2022self}.

\paragraph*{Structural risk minimization} We previously underlined that the main difficulty in supervised learning resides in the choice of the class of functions, because it is this choice that implements the bias-variance trade-off. However, the previous generalization bound suggests a simple strategy to determine the adequate class of functions. Consider several classes of candidate functions $ \mathcal F_1, \ldots , \mathcal F_N $ whose Rademacher complexity we know. For each class, we can find a function by the ERM algorithm, then calculate the value of the bound on the generalization error. The class of functions which minimizes this bound obtains the best theoretical guarantee on the error in generalization among the classes of candidate functions. It is therefore naturally this one that we want to select. This is exactly the principle of structural risk minimization (SRM) \cite{Vapnik2000}.

The two principles of empirical risk minimization and risk minimization structural risk are at the origin of a large number of learning algorithms, and can explain the algorithms that existed before the establishment of this theory. This is particularly the case of support vector machines (SVM), whose empirical success could be justified after the fact thanks to the SRM principle.

\section{First-order methods}
\label{GDA}
Minimization problems related to the ERM or SRM principles are solved using optimization techniques, whose development is sometimes strongly tied to those of the Machine Learning field. Without doubt, the Gradient Descent (GD) algorithm is the most widely used of the several optimization techniques employed in Machine Learning. GD is a first-order optimisation procedure that iteratively finds the (local) minimum of a convex differentiable surrogate function of the (regularized) 0/1 loss. 

The algorithm is based on the observation that if the loss function $\hat{\mathcal{L}}$ to be minimized is defined and differentiable in a neighborhood of a weight vector $\boldsymbol{w}^{(t)}$ then the loss decreases if one goes from the actual value of the loss $\hat{\mathcal{L}}(\boldsymbol{w}^{(t)})$, one step $\eta_t\in\mathbb{R}_+$ - called the learning rate, following a descente direction $\mathbf{p}_t$ defined as $\mathbf{p}_t^\top \nabla\hat{\mathcal{L}}(\boldsymbol{w}^{(t)})\leq 0$. 

It then comes that for a small learning rate $\eta_t$ if we define the new weight vector  $\boldsymbol{w}^{(t+1)}$ as
 \begin{equation}
    \label{eq:GD}
\boldsymbol{w}^{(t+1)}=\boldsymbol{w}^{(t)} -\eta_t \mathbf{p}_t,
\end{equation}
then 
\begin{equation}
\label{eq:GD2}
    \hat{\mathcal{L}}(\boldsymbol{w}^{(t+1)})\leq \hat{\mathcal{L}}(\boldsymbol{w}^{(t)}).
\end{equation}
It is obvious that if $\eta_t$ is too small then the decreasing condition \eqref{eq:GD2} does not guarantee to reach a local minima, or the true minimum, of  $\hat{\mathcal{L}}$. At each iteration of GD, the following sufficient conditions, known as Wolfe conditions \cite{Wolfe69}, have been proposed in order to ensure the convergence of the algorithm.

\begin{itemize}
    \item The decreasing of $\hat{\mathcal{L}}$ should not be too small with respect to the length of the jumps. Hence for  a given $\alpha\in(0,1)$, 
\begin{equation}
\label{eq:Armijo}
\forall t\in\mathbb{N}^*, \hat{\mathcal{L}}(\boldsymbol{w}^{(t)}+\eta_t \mathbf{p}_t)\leqslant \hat{\mathcal{L}}(\boldsymbol{w}^{(t)})+\alpha\eta_t\mathbf{p}_t^{\top} \nabla \hat{\mathcal{L}}(\boldsymbol{w}^{(t)})
\end{equation} 
This is known as the Armijo condition.
\item There should be a change in the curvature of the loss function after each update. Or equivalently the slope has decreased sufficiently; i.e. $\exists \beta\in(\alpha,1)$ such that  
\begin{equation}
\label{eq:Curvature}
\forall t\in\mathbb{N}^*, \mathbf{p}_t^{\top} \nabla\hat{\mathcal{L}}(\boldsymbol{w}^{(t)}+\eta_t \mathbf{p}_t) \geqslant \beta  \mathbf{p}_t^{\top} \nabla\hat{\mathcal{L}}(\boldsymbol{w}^{(t)})
\end{equation} 
This is known as the curvature condition.
\end{itemize}

These conditions are shown in Figure \ref{fig:Wolfe} which for a given weight $\boldsymbol{w}^{(t)}$ and a descent direction $\mathbf{p}_t$ depicts the loss $\hat{\mathcal{L}}(\boldsymbol{w}^{(t)}+\eta \mathbf{p}_t)$ with respect to the learning rate $\eta_t$.  At $\boldsymbol{w}^{(t)}$, the objective is to find a learning rate $\eta_t$ which guarantees that the decreasing of $\hat{\mathcal{L}}$ is not too small with respect to the length of the jumps  of the update; and that the slope has been reduced sufficiently. At $\hat{\mathcal{L}}(\boldsymbol{w}^{(t)})$ (i.e. $\eta_t=0$) the equation of the tangent to the loss with respect to $\eta$ is 
$\eta\mapsto \hat{\mathcal{L}}(\boldsymbol{w}^{(t)})+\eta\mathbf{p}_t^{\top} \nabla \hat{\mathcal{L}}(\boldsymbol{w}^{(t)})$. For $\alpha\in(0,1)$ , the line $\eta\mapsto \hat{\mathcal{L}}(\boldsymbol{w}^{(t)})+\alpha\eta\mathbf{p}_t^{\top} \nabla \hat{\mathcal{L}}(\boldsymbol{w}^{(t)})$ has a slope smaller in absolute value than the one of the tangent; hence providing an upper bound on the value of admissible learning rate.
\begin{figure}[t!]
\centering
\includegraphics[width=.8\textwidth]{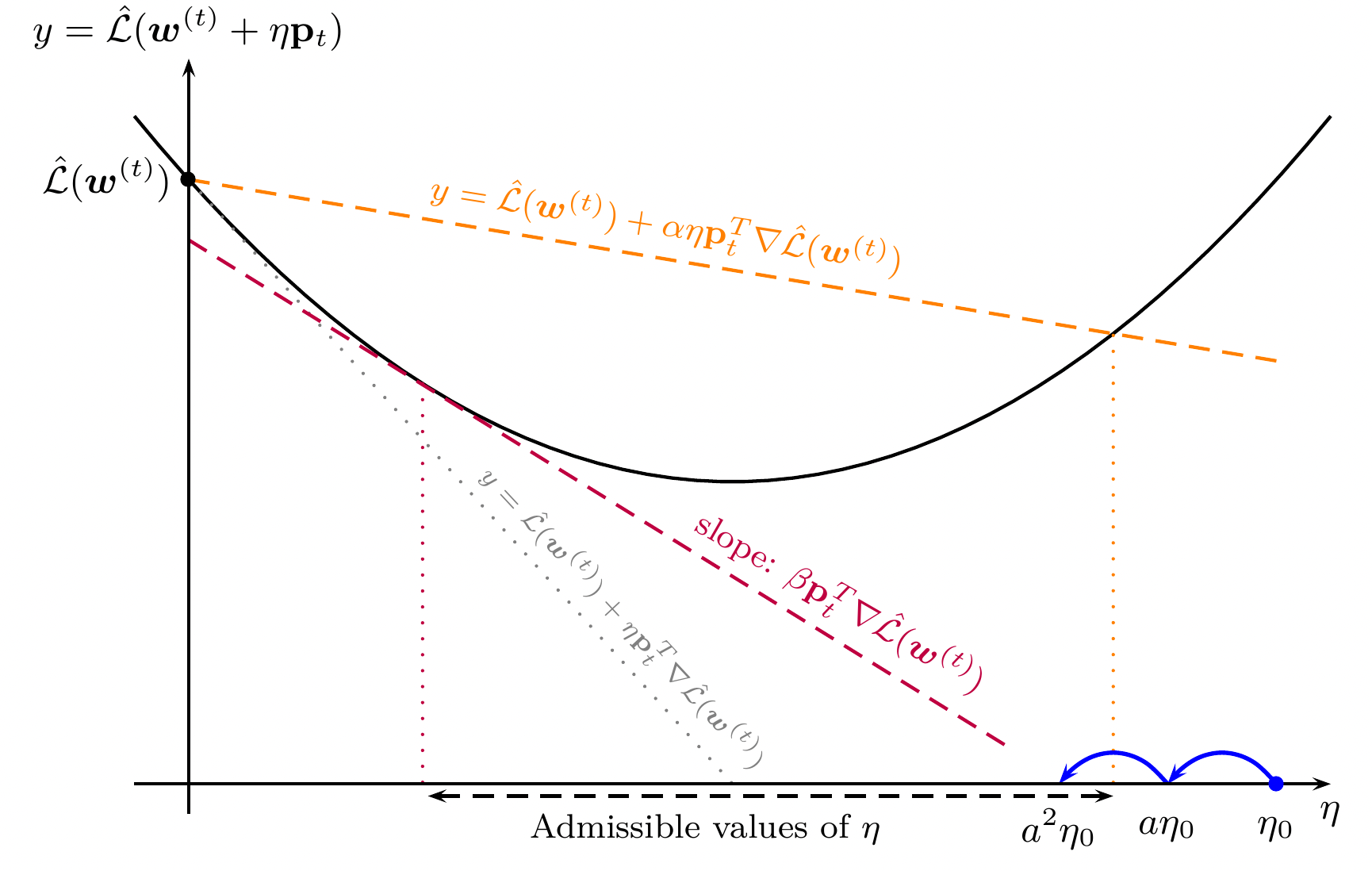}
\caption{Depiction of Wolfe conditions and the backtracking line-search strategy.}
\label{fig:Wolfe}
\end{figure}
The Armijo condition stipulates that the value of $\eta$ should be lower than this upper-bound. For a given $\beta\in(\alpha,1)$; the slope $\beta  \mathbf{p}_t^{\top} \nabla\hat{\mathcal{L}}(\boldsymbol{w}^{(t)})$ will be between $\mathbf{p}_t^{\top} \nabla\hat{\mathcal{L}}(\boldsymbol{w}^{(t)})$ and $\alpha\mathbf{p}_t^{\top} \nabla\hat{\mathcal{L}}(\boldsymbol{w}^{(t)})$ in absolute value. The curvature condition then ensures that the curvature of the loss on the new weight vector should be smaller than $\beta  \mathbf{p}_t^{\top} \nabla\hat{\mathcal{L}}(\boldsymbol{w}^{(t)})$. In practice, a line search method is used to determine the learning rate at each iteration.  It entails starting with a large value of the learning rate $\eta_0$ and then shrinking it iteratively by multiplying the current value with a factor $1>a>0$ (i.e., \textit{backtracking}) until the Armijo condition is met.

In the case where the loss function is convex and differentiable and that its gradient is Lipschitz continuous with parameter $L > 0$ defined as:
\begin{definition}
\label{def:Lipschitz}
The gradient of $\hat{\mathcal{L}}$ is Lipschitz continuous with parameter $L > 0$ if
\begin{equation}
\forall \boldsymbol{w}, \boldsymbol{w}';   \|\nabla\hat{\mathcal{L}}(\boldsymbol{w})-\nabla\hat{\mathcal{L}}(\boldsymbol{w}')\|_2\leq L\|\boldsymbol{w}-\boldsymbol{w}'\|_2
\end{equation} 
\end{definition}
Then the gradient descent algorithm is ensured to converge to the local minima of $\hat{\mathcal{L}}$ as stated in the following theorem.

\begin{theorem}[\cite{Zoutendijk:1966}]
Let $\hat{\mathcal{L}}$ be a differentiable objective function with a Lipschtiz continuous gradient and lower bounded. Suppose that the GD algorithm generates $(\boldsymbol{w}^{(t)})_{t\in\mathbb{N}}$ defined by $\forall t\in\mathbb{N}, \boldsymbol{w}^{(t+1)}=\boldsymbol{w}^{(t)}+\eta_t\mathbf{p}_t$; where $\mathbf{p}_t$ is a descent  direction of $\hat{\mathcal{L}}$ and $\eta_t$ a learning rate verifying both Wolfe conditions \eqref{eq:Armijo} and \eqref{eq:Curvature}. By considering the angle $\theta_t$ between the  descent  direction $\mathbf{p}_t$ and the direction of the gradient $
\cos(\theta_t)=\frac{\mathbf{p}_t^{\top}\nabla \hat{\mathcal{L}}(\boldsymbol{w}^{(t)})}{||\nabla\hat{\mathcal{L}}(\boldsymbol{w}^{(t)})||\times ||\mathbf{p}_t||}$; the following series is convergent
\[
\sum_{t} \cos^2(\theta_t) ||\nabla \hat{\mathcal{L}}(\boldsymbol{w}^{(t)})||^2
\] 
\end{theorem}

Various improvements to the gradient approach have recently been developed. The group of accelerated gradient methods is made up of these approaches, which the most popular ones are:

\begin{itemize}
    \item \underline{Classical Momentum}  \cite{polyak:64}
    
    Instead of using the true gradient this technique accumulates the gradients with the decaying parameter $\mu$ into momentum vector:
    \begin{equation}
        \boldsymbol{g}_t = \nabla \hat{\mathcal{L}}(\boldsymbol{w}^{(t-1)})
    \label{eq:cl_m_1}
    \end{equation}
    \begin{equation}
    \boldsymbol{m}_t = \mu \boldsymbol{m}_{t-1}+\boldsymbol{g}_t
    \label{eq:cl_m_2}
    \end{equation}
    \begin{equation}
    \boldsymbol{w}^{(t)} = \boldsymbol{w}^{(t-1)}-\eta \boldsymbol{m}_t
    \label{eq:cl_m_3}
    \end{equation}
    \item \underline{Nesterov's accelerated gradient} \cite{nesterov1983method}
    
    By plugging \eqref{eq:cl_m_2} into \eqref{eq:cl_m_3} we get that $    \boldsymbol{w}^{(t)} = \boldsymbol{w}^{(t-1)}-\eta \mu \boldsymbol{m}_{t-1} - \eta \boldsymbol{g}_t$. Nesterov momentum suggests the computation of the gradient immediately at the point $\boldsymbol{w}^{(t-1)}-\eta \mu \boldsymbol{m}_{t-1}$.
    
    \begin{equation*}
    \boldsymbol{g}_t = \hat{\mathcal{L}}(\boldsymbol{w}^{(t-1)})-\eta \mu \boldsymbol{m}_{t-1}
    \label{eq:n_m_1}
    \end{equation*}
    \begin{equation*}
    \boldsymbol{m}_t = \mu \boldsymbol{m}_{t-1}+\boldsymbol{g}_t
    \label{eq:n_m_2}
    \end{equation*}
    \begin{equation}
    \boldsymbol{w}^{(t)} = \boldsymbol{w}^{(t-1)}-\eta \boldsymbol{m}_t
    \label{eq:n_m_3}
    \end{equation}
    
    \item \underline{AdaGrad} \cite{Duchi11}
    
Another modified  gradient descent algorithm with per-parameter learning rate is \textit{adaptive gradient algorithm} (AdaGrad). Informally, this strategy raises the learning rate for sparser parameters while decreasing the rate for less sparse ones. In situations when data is sparse and sparse parameters are more useful, like in Natural language processing and image recognition applications, this technique often outperforms ordinary  gradient descent in terms of convergence. 
    \begin{equation*}
    \boldsymbol{g}_t = \nabla \hat{\mathcal{L}}(\boldsymbol{w}^{(t-1)})
    \label{eq:adagrad_1}
    \end{equation*}
    \begin{equation}
    \boldsymbol{w}^{(t)} = \boldsymbol{w}^{(t-1)}-\frac{\eta}{\|\boldsymbol{g}_t\|} \boldsymbol{g}_t
    \label{eq:adagrad_3}
    \end{equation}
    
    \item \underline{RMSProp} \cite{Hinton20}
    
Root Mean Square Propagation (RMSProp) is another method in which the learning rate is adjusted for each parameter. The aim is to divide a weight's learning rate by a running average of recent gradient magnitudes for that weight. As a result, the running average is first calculated in terms of the square root of the means.
    
    \begin{equation*}
    \boldsymbol{g}_t = \nabla \hat{\mathcal{L}}(\boldsymbol{w}^{(t-1)})
    \label{eq:rmsprop_1}
    \end{equation*}
    \begin{equation*}
    n_t =\nu n_{t-1}+(1-\nu)\boldsymbol{g}_{t}^\top  \boldsymbol{g}_t
    \label{eq:rmsprop_2}
    \end{equation*}
    where, ${\nu}$  is the forgetting factor. And the parameters are updated as,
    \begin{equation}
    \boldsymbol{w}^{(t)} = \boldsymbol{w}^{(t-1)}-\frac{\eta}{{\sqrt{n_t}+\epsilon}} \boldsymbol{g}_t
    \label{eq:rmsprop_3}
    \end{equation}
    Here $\epsilon$  is a small scalar in the order of $10^{-8}$ used to prevent division by 0.
    \item \underline{Adam} \cite{KingmaB14}
    
  The RMSProp optimizer has been updated with Adam (for Adaptive Moment Estimation). Running averages of both the gradients and the second moments of the gradients are used in this optimization process:
    
    \begin{equation*}
    \boldsymbol{g}_t = \nabla \hat{\mathcal{L}}(\boldsymbol{w}^{(t-1)})
    \label{eq:adam_1}
    \end{equation*}
    \begin{equation*}
    \boldsymbol{m}_t =\mu \boldsymbol{m}_{t-1}+(1-\mu)\boldsymbol{g}_{t}
    \label{eq:adam_2}
    \end{equation*}
    \begin{equation*}
    \hat{\boldsymbol{m}}_t =\frac{\boldsymbol{m}_t}{1-\mu}
    \label{eq:adam_3}
    \end{equation*}
    \begin{equation*}
    n_t =\nu n_{t-1}+(1-\nu)g_{t}^\top g_t
    \label{eq:adam_4}
    \end{equation*}
    \begin{equation*}
    \hat{n}_t =\frac{n_t}{1-\nu}
    \label{eq:adam_5}
    \end{equation*}
    \begin{equation}
    \boldsymbol{w}^{(t)} = \boldsymbol{w}^{(t-1)}- \frac{\eta}{\sqrt{\hat{n}_t}+\epsilon}\hat{\boldsymbol{m}}_t
    \label{eq:adam_6}
    \end{equation}
    where $\mu$ and $\nu$  are the forgetting factors for gradients and second moments of gradients, respectively.
\end{itemize}

\section{Classification}
\label{Sec:classif}
In this section, we present SVMs in the case of binary classification and neural networks. Both classifiers are undoubtedly the most popular classification algorithms in the field of Machine Learning, mainly due to the theoretical justifications for SVMs, and, their wide applications in different problems for neural networks. In Section \ref{sec:SVM}, we begin by presenting the notions of hyperplane separator of a set of examples and kernels, then the principle of support vector machines \cite{Vapnik2000}, which allows one to find a separating hyperplane thanks to a method which can be interpreted as a minimization of the structural risk presented in the previous section. We then describe neural networks in Section \ref{sec:nn_supervise}.

\subsection{Support Vector machines}
\label{sec:SVM}
Consider an input space $ \mathcal X \subset \mathbb R^d $; a linear classifier is a function from $ \mathbb R^d $ into $ \{ -1, 1 \} $ of the form $ f(x) = \textrm{sgn}(\langle\boldsymbol{w}, x \rangle + w_0 ) $ with $ \boldsymbol{w} \in \mathbb R^d$ and $ w_0 \in \mathbb R $, and $ \textrm{sgn}(t) = 1$ if $ t > 0 $, $-1$ otherwise. We notice that the hyperplan $ h(x) = \langle\boldsymbol{w}, x \rangle + w_0 $ divides $ \mathcal X $ into two subspaces which are the sets $ \{ x \in \mathcal X | \langle\boldsymbol{w}, x \rangle + w_0< 0 \} $ and $ \{ x \in \mathcal X |\langle\boldsymbol{w}, x \rangle + w_0 > 0 \} $.

\begin{figure}[t!]
\centering
\includegraphics[width=.6\textwidth]{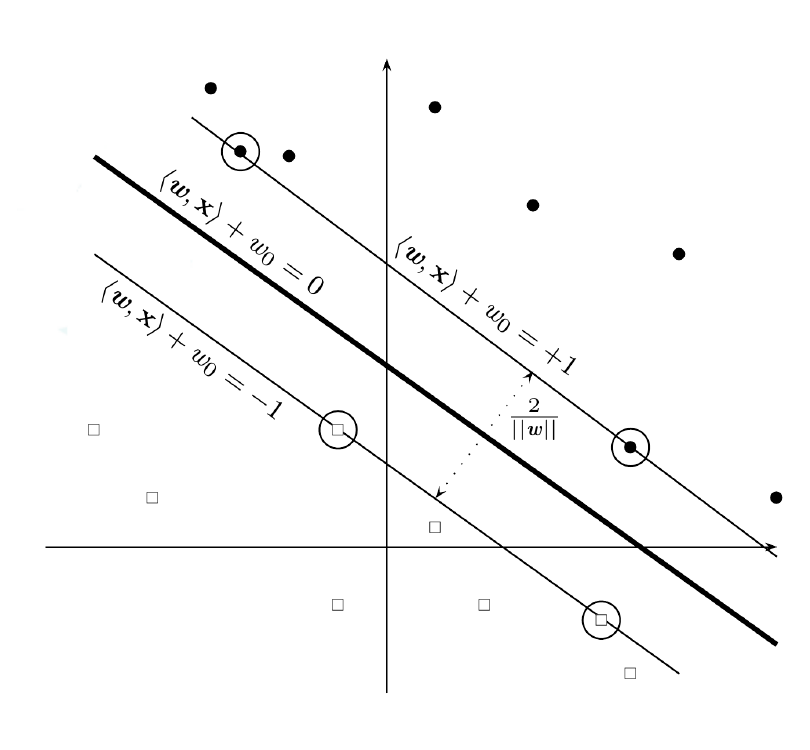}
\caption[Linear separation]{Illustration of the linear hyperplane serparator (in bold) perfectly separating the examples of the positive class (circle) and the negative class (square) as well as the margin. }
\label{img_svm_max_margin}
\end{figure}

Let us now consider a classifier $ h(x) = \langle\boldsymbol{w}, x \rangle+ w_0$ that correctly classifies all the examples of $S$.  In this case, there exists then  a scalar such that the examples $ ( x_i, y_i ) $ closest to the hyperplane satisfy $ | \langle\boldsymbol{w}, x \rangle + w_0| = $1. Now consider two observations $x_1$ and $x_2$ of different classes, such that $ \langle\boldsymbol{w}, x \rangle + w_0 = +1 $ and $\langle\boldsymbol{w}, x \rangle + w_0 = -1 $. The margin is defined as the distance between these two points, measured perpendicular to the hyperplane. In other words, the margin is $ \frac{\boldsymbol{w}}{ \|w_0\|} ( x_1 - x_2 ) = \frac{2}{\|\boldsymbol{w}\| }$. This notion of margin is illustrated in the figure \ref{img_svm_max_margin}.

\paragraph*{Hard margin SVM}
We have seen previously that provided that the examples closest to the hyperplane satisfy $ | \langle\boldsymbol{w}, x \rangle + w_0 | = 1$, then the margin is related to the norm $ \boldsymbol{w} $ by the relation $ \gamma = \frac{2}{ \|\boldsymbol{w}\| }$. This result therefore suggests minimizing the norm of $\boldsymbol{w}$ in order to determine a maximum margin hyperplane. Noting that the constraints $ | \langle\boldsymbol{w}, x \rangle + w_0| = 1$ can be written $ y ( \langle\boldsymbol{w}, x \rangle + w_0 ) = 1 $ for the examples close to the hyperplane, this amounts to solving the following optimization problem:
 
\begin{align*}
\min_{ \boldsymbol{w} \in \mathbb R^d, w_0 \in \mathbb R} & \frac{1}{2} \| \boldsymbol{w}\| ^2 \\
\textrm{s.c.} ~ & \forall i, y_i ( \langle\boldsymbol{w}, x \rangle + w_0 ) \geq 1
\end{align*}

We recognize a quadratic optimization problem with linear constraints
linear. In general we will not try to solve this problem directly, but rather we will be interested in the dual problem \cite{FerrisM02}:

\begin{align*}
\max_{ ( \alpha_1, ..., \alpha_m) \in \mathbb R^m } & \sum_{i=1}^m \alpha_i - \frac{1}{2} \sum_{i=1} ^m \sum_{j=1}^m y_i y_j \alpha_i \alpha_j x_i^\top x_j \\
\textrm{s.c.} ~ & \sum_{i=1}^m y_i \alpha_i = 0 \\
& \forall i, \alpha_i \geq 0
\end{align*}

This formulation has the advantage of expressing the vector $b$ solution of the optimization problem
initial in the following form:
\begin{align*}
\boldsymbol{w} = \sum_{i=1}^m y_i \alpha_i x_i
\end{align*} 
where $ ( \alpha_1, ..., \alpha_m ) $ is an optimal solution of the dual problem. Moreover, it can be shown
that $ \alpha_i > 0 $ if and only if $ y_i ( \langle\boldsymbol{w}, x \rangle + w_0 ) = 1 $. Thus, the normal vector of the optimal hyperplane
 can be decomposed as a linear combination of the input vectors
which are at the minimum distance from this hyperplane. These input vectors are called the
\textit{support vectors}. Thus, the maximum-margin hyperplane has the property of only depending on a subset of examples. These examples lie exactly on the margin and are called the support vectors. The other examples could be anywhere outside the margin without changing the solution. We would therefore find the same solution if the training set $S$ contained only these support vectors.

\paragraph*{Theoretical justification}
Support vector machines have been used successfully in many fields, but it is not immediately clear how good they perform from a theoretical point of view. \cite{Vapnik2000} provides an explanation by linking the notion of separation margin to that of VC dimension. In particular, he proves the following theorem: 

\begin{theorem}[\cite{Vapnik2000}]

Let $ \boldsymbol{w} \in \mathbb R^d $ be such that $ \| \boldsymbol{w}\| = 1 $, $ c_{\boldsymbol{w},w_0, \gamma} $ the classifier defined by the following relation:
$ c_{\boldsymbol{w},w_0, \gamma}( x) = 1 $ if $ \langle\boldsymbol{w}, x \rangle + w_0 \geq \gamma $, and $ c_{\boldsymbol{w},w_0, \gamma}( x) = -1 $ if $ \langle\boldsymbol{w}, x \rangle + w_0 \leq -\gamma $. This classifier
is called a $\gamma$-margined separator hyperplane. In cases where $x$ does not match any of the
two conditions, we consider it to be ignored. Then, if the space of observations
$ \mathcal X $ is included in a ball of radius $B$, the dimension VC of the set of 
margin separator hyperplanes $ \gamma $ over $ \mathcal X $ is less than $ \left\lceil R^2 / \gamma^2 \right\rceil + 1$, where $ \left\lceil t \right\rceil $ is the upper integer part of $t$.
\end{theorem}

The separation margin is therefore related to the VC dimension: the more one
separating hyperplane achieves a wide separation margin on a training set $S$, the more it
can be considered as part of a set of low-dimensional VC functions. However, we have seen that the principle of structural risk minimization suggests favoring low-dimensional VC binary classifiers. Thus by maximizing the margin, support vector machines minimize the VC dimension and can therefore be seen as \textit{implementations of the structural risk minimization principle}.

\paragraph*{Soft Margin SVM}
The hard-margin SVM presented in the previous section can only apply when $S$ is linearly separable. In practice this is rarely the case, in particular because of noise problems ($S$ contains examples whose observed class is not the true class), or quite simply because the problem is not linearly separable. An example of a nonlinearly separable classification problem is given in figure \ref{fig:nonlin}. To be able to use SVM on such data, it must therefore be made capable of accepting the misclassification of certain examples.

\begin{figure}[t!]
\centering
\includegraphics[width=.6\textwidth]{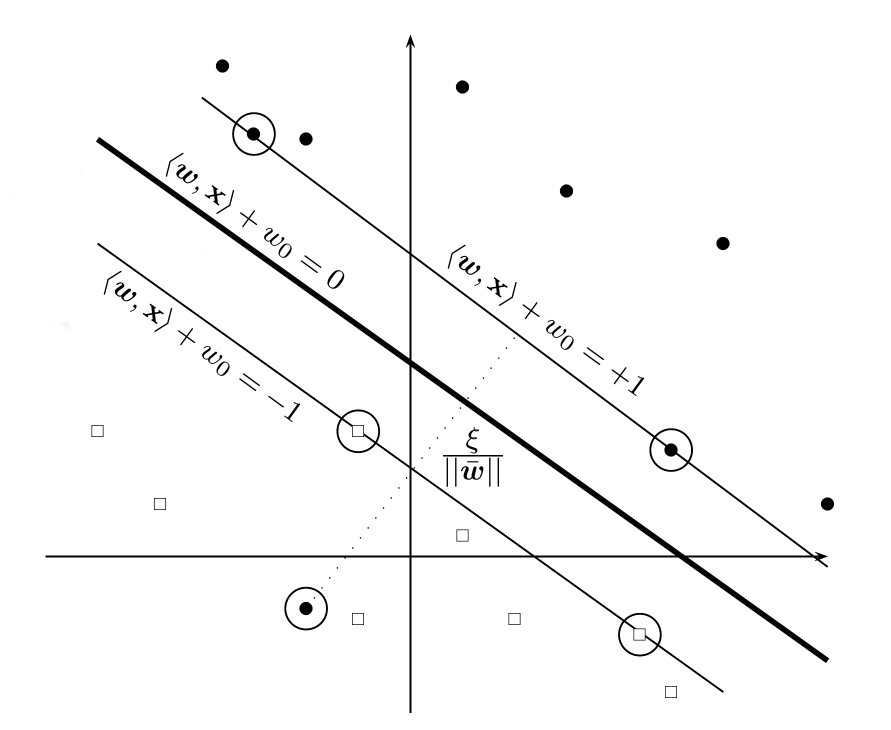}
\caption[Non-linearly separable problems]{Illustration of a non-linearly separable classification problem. Support vectors are circled.}
\label{fig:nonlin}
\end{figure}

A simple way to allow classification errors is to relax the constraints on the margin by introducing slack variables. To an example of $S$ is associated a slack variable, which allows us to associate a cost each time the corresponding example violates the constraint on the margin. The new SVM thus defined is then said to \textit{soft margin}. The new objective of the soft-margin SVM is therefore twofold: to maximize the margin and to minimize the number of examples violating the constraint on the margin. In other words, we will minimize the norm of $\boldsymbol{w}$ and the sum of the costs associated with the spring variables. The new optimization problem is written:
\begin{align*}
\min_{ \boldsymbol{w} \in \mathbb R^d, w_0 \in \mathbb R} & \frac{1}{2} \| \boldsymbol{w} \| ^2 + C \sum_{i=1}^m \xi_i \\
\textrm{u.c.} ~ &  \forall i, y_i (  \boldsymbol{w}^\top x_i + w_0 ) \geq 1 - \xi_i \\
~ & \forall i, \xi_i \geq 0 \\
\end{align*}
where the $ \xi_i $ are the slack variables, $C > 0$ a real to choose from. When we have $ \xi_i > 0$, then the corresponding constraint is violated. The cost associated with this violation is worth $ C \xi_i $, which we can compensate by decreasing the norm of $\boldsymbol{w}$. If $ C$ is large, then the slightest constraint violation will have a large cost, and the solution will therefore favor hyperplanes with a small margin but with few margin constraint violations. Conversely, a low $C$ will allow more classification errors and will favor large-margin hyperplanes. We thus see that $C$ allows to parameterize the compromise between the maximization of the margin and the violations of the constraints on the margin. In practice, the $C$ coefficient will be chosen by standard model selection methods such as cross-validation.

\paragraph*{Hinge loss and regularization} We will now interpret soft-margin SVMs from the perspective of regularization. An optimization problem is said to be regularized when the optimized function is the sum of two errors: the cost function that really interests us (the classification error for example), and a regularization term. This regularization term is used either to stabilize the solution (i.e. to ensure that it does not vary too much), or to incorporate \textit{a priori} knowledge of the problem (i.e. to introduce a bias). Many learning algorithms can be interpreted as regularized problems. We will see that this is particularly the case for soft-margin SVMs.

For this we define the loss function $ \ell( f(x), y ) = \max( 0, 1- y h(x) )$, called \textit{Hinge} loss. This function is shown in figure \ref{img_hinge_loss}.
\begin{figure}[h!]
   \begin{center}
      \includegraphics[width=0.6\textwidth]{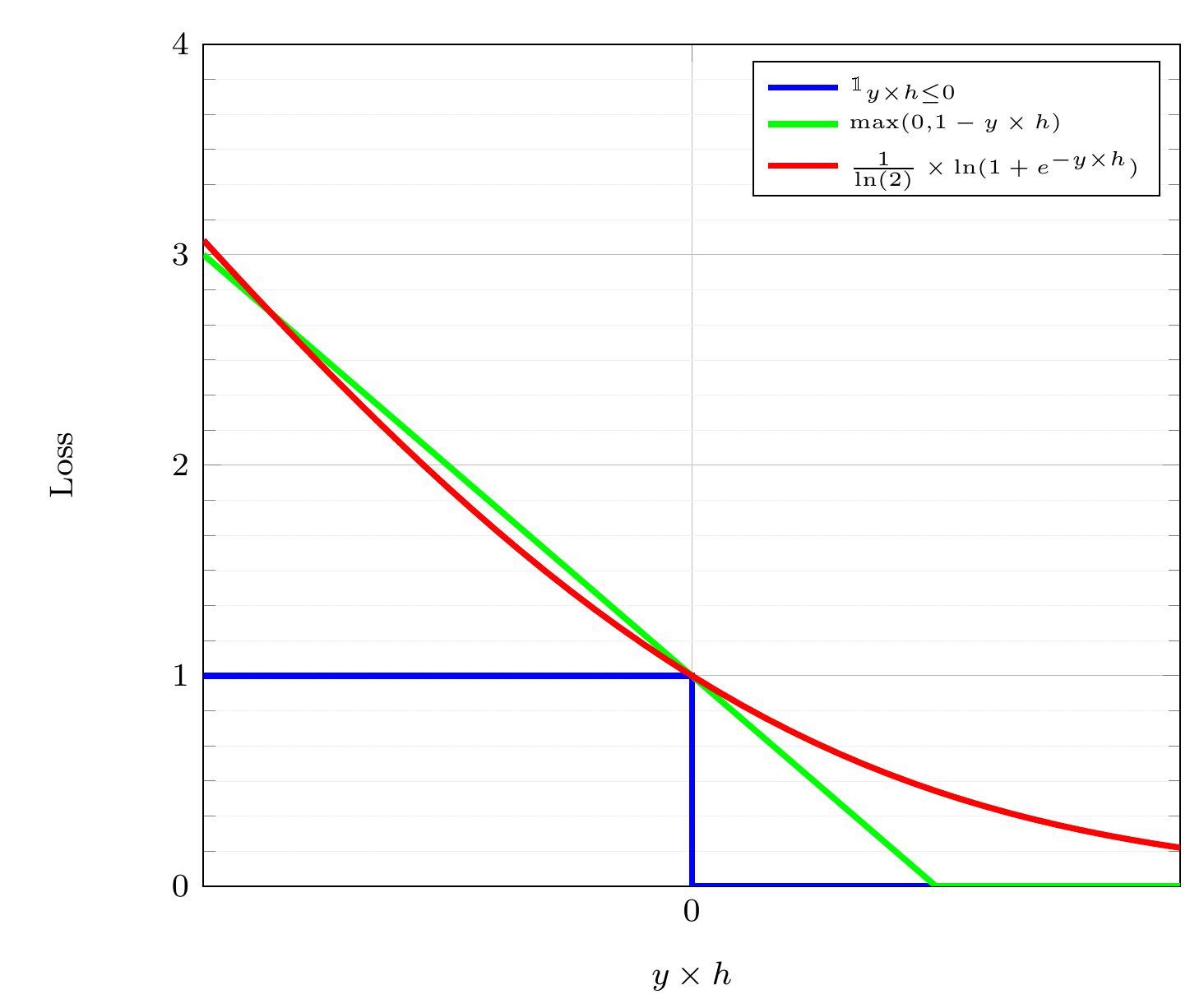}
   \end{center}
   \caption[The Hinge loss]{0/1 loss (blue), Hinge loss (green), and logistic loss (red)}
\label{img_hinge_loss}
\end{figure}

We can now rewrite the previous optimization problem without the constraints:
\begin{align*}
\min_{ \boldsymbol{w} \in \mathbb R^d, w_0 \in \mathbb R} \frac{1}{2} \|\boldsymbol{w}\| ^2 + C \sum_{i=1}^m\max \left( 0, 1 - y_i( \boldsymbol{w}^\top x_i + w_0 ) \right )
\end{align*}

If we divide the loss function by the constant $C$, we recognize a regularization term and the empirical error:
\begin{align*}
\hat{\mathcal{L}}(\boldsymbol{w},w_0)=\underbrace{ \sum_{i=1}^m \max \left( 0, 1 - y_i( \boldsymbol{w}^\top x_i + w_0 ) \right ) }_{ \textrm{empirical error} } + \underbrace{ \frac{1}{2C} \| \boldsymbol{w}\| ^2 }_{ \textrm{regularization term} } 
\end{align*}

Thus the soft-margin SVM can be seen as a regularized learning problem, where the regularization function introduces a bias towards large-margin hyperplanes. Note that with this regularized learning interpretation, the empirical error is not the 0/1 error that initially interested us but the surrogate Hinge loss. In \cite{BartlettM02} it is shown that the minimizer of any surrogate loss of the 0/1 loss, in which the associated instantaneous loss $\ell:(y,h(x))\mapsto \ell(y,h(x))$ is continuous and passes through 1 when $yh(x)=0$, is likewise the minimizer of the 0/1 classification error.

\subsection{Neural network approaches}
\label{sec:nn_supervise}

Artificial neural networks, are perhaps the most popular learning systems nowadays  whose design were originally schematically inspired by the functioning of biological neurons discovered by \cite{Cajal}.

Dendrites are the connections through which the neuron receives impulses, whereas an axon is the link through which the neuron transmits the impulse. Each neuron has one axon. Dendrites and axons have a complicated branching structure. A synapse is a connection between the axon and the dendrite. A neuron's primary purpose is to carry information from the dendrites to the axon. Distinct dendritic signals, on the other hand, can have different effects on the signal in the axon. If the overall impulse surpasses a specific threshold, the neuron will send out a signal. The neuron will not respond to the impulse if this happens, and no signal will be sent to the axon. 

From this discovery, the concept of artificial neural networks was proposed in \cite{Mccullogh43}, where the two researchers presented their theory that the activation of neurons is the basic unit of brain activity and described the formal neuron which mimics the functioning of a biological neuron as shown in Figure \ref{fig:biol_art_neuron}).

\begin{figure}[t!]
\centering
\includegraphics[width=.8\textwidth]{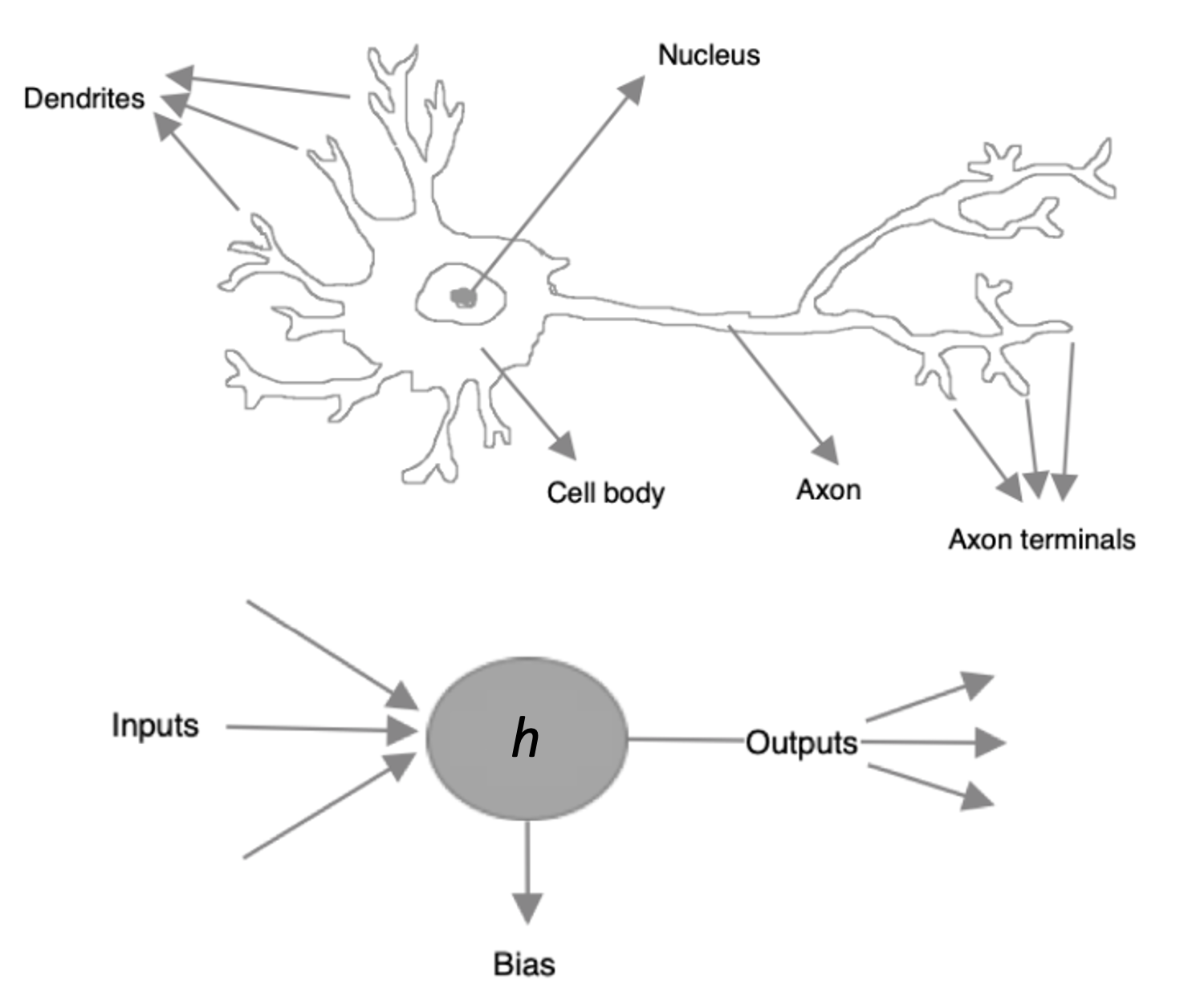}
\caption{Structure of a biological (top) and a formal (down) neuron.}  
\label{fig:biol_art_neuron}
\end{figure}

For a given input $x=(x_1,\ldots,x_d)^\top\in\mathbb{R}^d$, each characteristic $x_j$ is supposed to be a real valued signal which arrives (from a dentrite) to a computing unit (i.e. the nucleus). This unit estimates a weighted sum of the all characteristics: $\sum_{j=1}^d w_jx_j$ and compares the sum to a bias $w_0$. The output of the neuron is then computed using an activation function (see below) over a linear combination of the input: $a:x\mapsto a(\boldsymbol{w}^\top x+w_0)$

\cite{Rosenblatt57} invented Perceptron which is the oldest machine learning algorithm, designed to perform complex pattern recognition tasks. It is this algorithm that will later allow machines to learn to recognize objects in images. The activation function of Perceptron is the identity function and the weights of the model are learned per example at each time that the model makes an error on the class of an example in input. The weights are updated using a stochastic version of the gradient algorithm by minimising the distance of the misclassified example to the current hyperplan. 

At that time, neural networks were limited by technical resources. For example, computers were not powerful enough to process the data needed to run neural networks. This is the reason why research in the field of Neural Networks has remained dormant for many years. It took until the early 2010s, with the rise of Big Data and massively parallel processing, for Data Scientists to have the data and computing power needed to run complex neural networks. In 2012, during a competition organized by ImageNet, a Neural Network managed for the first time to surpass a human in image recognition. This is why this technology is again at the heart of the concerns of scientists. Nowadays, artificial neural networks are constantly improving and evolving day by day.

\paragraph*{Activation functions} The term \textit{activation function} comes from the biological equivalent "activation potential", the stimulation threshold which, once reached, triggers a neuron response. The activation function is often a nonlinear function. An example of an activation function is the Heaviside function, which always returns 1 if the input signal is positive, or 0 if it is negative.

The main activation functions are the following.

\begin{itemize}
    \item \textbf{ReLU}: ReLU has the following formula $a(z) = max(0, z)$ and implements a simple threshold transition at zero point. The use of ReLU significantly increases the speed of training, but ReLU has one significant drawback - neurons could "die" during the training. It means that they could come into a state where the output will always be 0.
    \item \textbf{Leaky ReLU}:Leaky ReLU is one of the attempts to solve the problem of dead neurons in ReLU described above. The usual ReLU gives a zero on the interval $z < 0$, while Leaky ReLU (LReLU) has a small negative value on this interval. That is, the function for LReLU has the form $a(z) = \beta z$ for $z < 0$ and $a(z) = z$ for $z \geq 0$, where $\beta$ is a small constant.
    \item \textbf{Randomized ReLU}: For a randomized ReLU (RReLU), the angular coefficient on the negative interval is randomly generated from the specified interval during training, and remains constant during testing. It also is noticed \cite{DBLP:journals/corr/XuWCL15} that RReLU allowed to reduce the overfitting due to their element of randomness.
\end{itemize}

These activation functions are shown below.
\begin{figure}[H]
\centering
\vspace*{-10pt}
\includegraphics[width=1.0\textwidth]{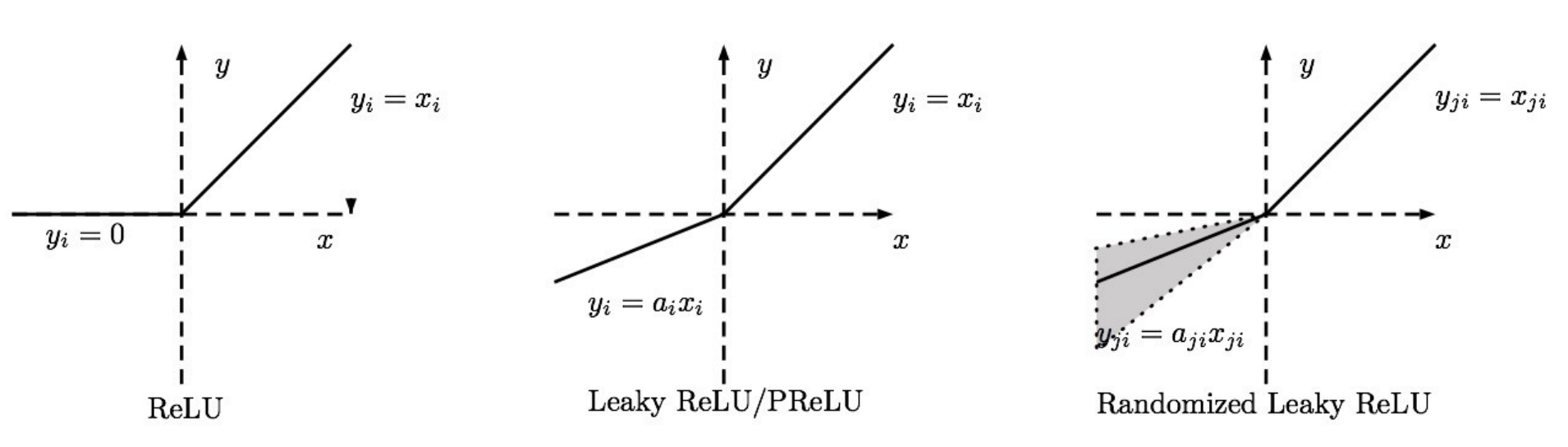}
\caption{ReLU based activation functions}  
\label{fig:relu}
\end{figure}

Other classicical activation function with smooth derivatives are:

\begin{itemize}
\item \textbf{Sigmoid}: The sigmoid is expressed by the following formula $\sigma(z) = \frac{1}{(1 + e^{-z})}$. This function takes an arbitrary real number at the input, and gives a real number in the range from 0 to 1 at the output. In particular, large negative numbers turn into zero, and large positive ones turn into one.
\item \textbf{Softmax}: This is a generalization of the sigmoid function for the multidimensional case. The function $\sigma(z_i) = \frac{e^{z_i}}{\sum_{k=1}^{K}e^{z_k}}$ converts a real number $z_i$ (element of the vector $\boldsymbol{z}$ with dimension $K$) into a real number from the interval $[0,1]$ and the resulting sum of the coordinates of the vector $\boldsymbol{z}$ is 1.
\item \textbf{Hyperbolic Tangent}: The hyperbolic tangent $tanh(z)$ takes an arbitrary real number at the input, and gives a real number in the range from -1 to 1 at the output.
\end{itemize}

\paragraph*{Multi-Layer Perceptron} The multilayer perceptron (MLP) is a sort of artificial neural network composed of many layers in which information flows directly from the input layer to the output layer. Each layer has a different amount of neurons, with the neurons in the last layer (known as "output") serving as the entire system's outputs. Layers between the input and the output layers are called hidden layers.

The two types of neural networks most studied in the literature are recurrent networks, where there are loops between the different hidden layers and also between units of these layers, and forward propagation networks (or \textit{feed forward}) without a loop, which we consider in the following. They are usually organized into layers of neural units each similar to the one described previously.

In feed forward neural network, the information is propagated as follows:
Given an input example $x$ and its desired output $y$.
The signal is propagated forward in the layers of the neural network:

\begin{itemize}
    \item  $x_{k}^{(n-1)}\mapsto x_{j}^{(n)}$, where $n$ is the layer number.
    \item Forward propagation is calculated using the activation function $a$, the aggregation function $h$ (often a dot product between the weights and the inputs of the neuron) and synaptic weights $w_{jk}$ between neuron $x_{k}^{(n-1)}$ and the neuron $x_{j}^{(n)}$. 
    \[
    x_{j}^{(n)}=a^{(n)}(h_{j}^{(n)})=a^{(n)}\left(\sum_{k}w_{jk}^{(n)}x_{k}^{(n-1)}+w_{j0}\right)
    \]
\end{itemize}

When forward propagation is complete, the output is $\hat{y}$.
We then calculate the error between the output given by the network $\hat{y}$ and the desired output vector $y$. :

\paragraph*{Loss functions}
The loss function can be calculated in a variety of ways, depending on the task formulation. For example, if the askt is regression, the most relevant losses are Mean squared error (MSE) or Mean absolute error (MAE):

\begin{align*}
\ell_{MSE}(y,\hat{y}) &= \sum_{k=1}^{K}(y_k-\hat{y}_{k})^{2}\\
\ell_{MAE}(y,\hat{y}) &= \sum_{k=1}^{K}|y_k-\hat{y}_{k}|
\end{align*}

If the task is a classification task, the most popular loss is the cross-entropy error
\[
\ell_{CE}(y,\hat{y}) = -\sum_{k=1}^K y_k\log(\hat{y}_{k})
\]
where $y_k\in\{0,1\}$ and $\hat{y}_{k}=\frac{1}{1+e^{-h_{k}^{(N)}}}$ is the predicted output using a sigmoid activation. 

\paragraph*{Back-propagation of the gradient}

The weights of a neural network are updated by minimizing the loss layer per layer, from the output to the input layer, throughout training. This sequential update of the weights is called the back-propagation of the gradient.

The gradient descent algorithm is in general used in this update.
\[
w_{jk}^{(n)}=w_{jk}^{(n)}-\eta\frac{\partial \ell(y,\hat{y})}{\partial w_{jk}^{(n)}}
\]

The chain rule is used to calculate the derivative of a loss with regard to a weight:

\[
\frac{\partial \ell(y,\hat{y})}{\partial w_{jk}^{(n)}}=\frac{\partial \ell(y,\hat{y})}{\partial h_j^{(n)}}\frac{\partial h_j^{(n)}}{\partial w_{jk}^{(n)}}
\]

\section{Ranking}
\label{sec:ranking}

The task of classification described above is arguably one of the most studied task in the literature. However, there are numerous situations where ranking the observations rather than assigning them to a class is more appropriate. The most typical example in information retrieval is search engines such, which give the user with a list of documents sorted by relevance, rather than a collection of papers all deemed relevant and presented in no particular order. Another example is recommendation systems that we will present in depth in the next chapter. The goal of these systems is to propose items that are likely to interest a user. Sorting items in order of relevance seems more acceptable from a recommendation standpoint than predicting a score or class for each item. 

The task of ordering a set of objects with respect to a fixed information request is called \textit{instance ranking} (or ranking in short). More precisely, a ranking problem is defined by an ordering relation on the space of instances $ \mathcal X $, allowing to order $ x_1 $ and $ x_2 $ for any pair of instances $ ( x_1, x_2 ) $ in $ \mathcal X $.

A simpler and more natural way to model the order relation  is to use a function $ f : \mathcal X \rightarrow \mathbb R $ which assigns an actual score to any example $ x \in \mathcal X$. The order between the instances is then deduced from the comparison of the values of $f$. So $f(x_1) > f(x_2)$ means that $x_1$ is ranked above $x_2$.

By modeling the order relation in this way, we formulate the ranking of examples as the learning of a score function, as in classification. On the other hand, let us underline an important difference: in classification, the learned functions directly give the expected predictions. In ranking, learned functions return scores whose absolute values are not important per se. Indeed, these values are only used to compare the examples with each other. It is therefore the relative values of the scores that are important. 

\subsection{Ordering induced by scores}

Given a set of examples $S = ( x_1, ..., x_m ) $, we assume that the desired ordering is induced by scores $ Y = ( y_1, ..., y_m ) $. These scores induce a strict partial order on the set of inputs $S$: $x_i$ is ordered above $x_j$ if $ y_i > y_j $. This is also the case in collaborative filtering (CF), where each user can attribute to each item a value expressing his or her preferences on a rating scale: $ y=5 $ if (s)he liked it a lot, and, $ y=1 $ the reverse. In this case, items with different scores can be ordered relative to each other. 

We assume that a part of the examples $(x_i, y_i )$ is known and available for training. The purpose of instance ordering is to learn a score function which must retrieve a desired order from the training examples. We therefore find a task similar to classification, where the goal is to learn a function from a few examples in order to find the outputs for new observations. On the other hand, our goal is no longer to predict the scores of the unobserved examples, but to predict the order between the instances. To take this difference into account, we must adapt the notions of learning error and error in generalization to the framework of ranking.

\subsection{Ranking error on crucial pairs}

A ranking loss is a function of the form $ \mathcal{L}_r: \mathbb R^m \times \mathbb R^m \rightarrow \mathbb R^+$. It takes two vectors of scores as input: the vector of desired scores, and the vector of predicted scores. It returns a positive real which measures the error between the vector of desired scores and the vector of predictions. Note the difference with the error functions used in classification, which compares the value of a desired output with its prediction. In ranking, the value of an output does not matter, it is its relative value in relation to the other outputs that matters. This is why the error functions consider score vectors and not individual scores.\\

We call \textit{crucial pairs} the pairs of observations $(x_i, x_j )$ such that $ y_i > y_j $. We expect a prediction function $f$ to make few prediction errors on crucial pairs, ie it satisfies $ f(x_i) > f(x_j) $. This leads us to define the \textit{ranking loss on crucial pairs}:
\begin{equation}
\label{eq:RL}
\mathcal{L}_r( f( s ), Y ) = \frac{1}{ \sum_{i,j} \mathbbm{1}_{y_i > y_j}} \sum_{ y_i > y_j } \mathbbm{1}_{f( x_i ) \leq f ( x_j )}
\end{equation}

where $ f(s) = \left ( f(x_1) , ..., f(x_m ) \right )$ and $ Y = \left ( y_1 , ..., y_m \right )$. The denominator is simply the number of crucial pairs that can be generated from the score vector $Y$. At the numerator, we recognize the number of crucial pairs on which the order predicted by $f$ is not the desired order. The ranking loss on the crucial pairs, which serves as our empirical error, is therefore simply the proportion of crucial pairs incorrectly predicted by $f$.

\subsection{Other ranking approaches}
There are two other Learning-to-Rank approaches which are pointwise and listwise ranking techniques \cite{Liu:2009}.

In Pointwise approaches, ranking is  formulated as a regression problem, in which the rank value of each example is estimated as an absolute quantity to be found. In the case
where relevance judgments are given as pairwise preferences
(rather than relevance degrees), it is usually not straightforward to apply these algorithms for learning. Moreover, pointwise techniques do not consider the inter dependency among examples, so that the position of examples in the final ranked list is missing in the regression-like loss functions used for learning. 

On the other hand, listwise approaches take the entire ranked
list of examples as a training instance. As
a direct consequence, these approaches are able to  consider the  position of examples in the output ranked list at the training stage. Listwise techniques aim to directly optimize a ranking
measure, so they generally face a complex optimization problem dealing with non-convex, non-differentiable and discontinuous functions.

In terms of models, perhaps the first ranking based model is RankProp, originally proposed by \cite{Caruana:1995}. RankProp is a pointwise approach that alternates between two phases of learning the desired real outputs by minimizing a Mean Squared Error (MSE) objective, and a modification of the desired values themselves to reflect the current ranking given by the net. Later on \cite{Burges:2005} proposed RankNet, a pairwise approach, that learns a preference function by minimizing a cross entropy cost over the pairs of relevant and irrelevant examples. SortNet proposed in \cite{DBLP:journals/tnn/RigutiniPMS11} also learns  a preference function by minimizing a ranking loss over the pairs of examples that are selected iteratively with the overall aim of maximizing the quality of the ranking. A complete survey on the complexity of the Google PageRank problem~\cite{brin1998anatomy}, a core of many modern algorithms, is given in~\cite{anikin2020efficient}.

\section{Conclusion}
In this chapter we have provided a brief overview of supervised learning by focusing on classification and ranking tasks. The two main other frameworks that are not covered are unsupservised learning \cite{Hinton99} and semi-supervised learning. In unsupervised learning, the aim is to find similar groups from a set of examples for which we do not have desired outputs. Unsupservised learning approaches  exploit the structure of data to find these clusters and have been applied in many applications, such as Information Retrieval \cite{Pessiot:10}, or image segmentation \cite{Xia17}. Semi-supervised learning, on the other hand, tries to take use of both label information in a small collection of labeled data and data structure in a large quantity of unlabeled data for learning \cite{CSZ2006}.

The presentation given in this chapter aims to  pave the way for the introduction of recommender systems that are the main application task that we considered in this thesis.

\chapter{Recommender Systems}
\label{ch:1}
\section{Introduction}\label{Introduction}

With the development of e-commerce, Internet users are offered a growing choice of products and services online. To guide them, most sites use recommendation systems. Their goal is to generate personalized recommendations, ie to determine for each user the products or articles most likely to interest him. To achieve this, the most effective implementations to date use the preferences of other users to generate these recommendations: this is the principle of collaborative filtering. Collaborative filtering is particularly suitable for recommending cultural products such as films, books or music, and is used successfully by commercial online recommendation systems such as \url{Amazon.com} or \url{CDnow.com}.\\

Collaborative filtering techniques have in particular given rise to a large number of recommendation systems on the Internet, for example for films (MovieLens\footnote{\url{http://www.movielens.org}}, ymdb.com, ...), or for web pages (Del.icio.us\footnote{\url{http://del.icio.us/}} ) through bookmark pooling \cite{Pessiot:07}. They are also the basis of the personalized proposals for items to buy that are made on commercial sites such as Amazon.com or CDNow.com. The development of high-performance collaborative filtering systems therefore presents significant economic challenges. Different approaches have been proposed for collaborative filtering; the most popular ones rely on matrix factorization which intends to factorize the sparse matrix of users and items where each cell of the matrix is either a note or a binary value corresponding for example to a click, into the multiplication of two matrices each corresponding to a latent representation of respectively users and items. The main challenges of matrix factorization approaches for recommender systems are how to tackle the great sparsity of the original matrix and how to make the models scalable?

To address these points, ranking models for recommendation have attracted many interest in both the industry and the academic research community in recent years. Given a system (set of users, customers etc.), the goal here is to provide a ranking of objects (items, products, adverts etc.), based on the information about the interaction of these objects with the system and their individual characteristics. Common examples of applications include the recommendation of movies (Netflix, Amazon Prime Video), music (Pandora), videos (Youtube), news content (Outbrain) or advertisements (Google).

Feedback provided by the system and exploited to learn ranking scores can be {\it explicit}, presented mostly by ratings; or {\it implicit} that include clicks, browsing over an item or listening to a song. Such implicit feedback is readily available in abundance but is more challenging to take into account as it does not clearly translate the preference of a user for an item. The idea here is that even a clicked item does not necessarily express the preference of a user for that item, it has much more value than a set of unclicked items for which no action has been made. In most of these approaches, the objective is to rank the clicked item higher than the unclicked ones by finding a suitable representation of users and items in a way that for each user the ordering of the clicked items over unclicked ones is respected by dot product in the joint learned space. One common characteristic of publicly available collections for recommendation systems is the huge unbalance between positive (click) and negative feedback (no-click) in the set of items displayed to the users, making the design of an efficient online RS extremely challenging. Some works propose to weight the impact of positive and negative feedback directly in the objective function \cite{Pan:2008} to improve the quality. Another approach is to sample the data over a predefined buffer before learning \cite{Liu2016}, but these approaches do not model the shift over the distribution of positive and negative items, and the system's performance on new test data may be affected.  

In this chapter, we will review main approaches proposed for recommender systems by focusing on  learning-to-rank setting for this task that have been developed for the off-line case. This problem differs from the problem of personalized recommendation where the goal is to perform the recommendation online. The process of the latter is as follows: a user comes into the systems and is displayed some ads based on his or her previous interactions with the systems. At this time, we assume that the user is starting a new session. Then, he or she starts to interact with the displayed ads. A traditional off-line approach will have to wait for the end of the session in order to learn potential new recommendations for the next visit of this user. However, in the on-line setting, the aim is to develop a model which can adapt the recommendation within the same session. Therefore, the parameters of the model will be updated online. Providing high-quality online ranking is a challenging task for several reasons: 
\begin{itemize}
  \item It is strongly time-dependent: the set of relevant items for each user changes over time and the relevance of the items depends on the preferences of a particular user at a specific instant of time.
  \item The number of positive feedback, for instance clicks, are very rare (i.e. the data are sparse).
  \item It is difficult to provide recommendations for new users and/or new items (cold-start).
\end{itemize}

We will come over these points on the next chapters. In the reminder, we will first present in Section \ref{sec:techniques} matrix factorization and ranking-based models for recommender systems, then in Section \ref{sec:metrics} we will describe classical measures that have been used to evaluate these systems.

\section{Different approaches}
\label{sec:techniques}

Two main approaches have been proposed to solve the problem of ranking. The first one, referred to as Content-Based recommendation techniques (CBF)\cite{Pazzani2007} make use of existing contextual information about 
the items (e.g. textual description, meta-data) for recommendation. The second approach, referred to as collaborative filtering (CF) and undoubtedly the most popular one \cite{Su:2009}, relies on the past interactions and recommends items to users based on the feedback provided by similar other users.

\subsection{Matrix Factorization}
CF methods operate with a huge matrix of users-objects (each row corresponds to a user, and each column is an item).
To solve this problem, many methods use a matrix decomposition \cite{Koren:2009}, which shows good results. Let's give some mathematics key-points regarding this approach.
\\
Suppose $\userS$ - the set of all users, $\itemS$ - the set of items and $R$ is a matrix of size $|\itemS|\times|\userS|$ which contains all ratings that users have given to the items. Assume, that we consider $K$ latent variables, then the goal is to find two matrices $P$ and $Q$ such that their product approximates the matrix $R$:
\begin{equation}
R\approx \tilde{R} =  P\times Q^{T},
\end{equation}
\\
where $P$ is a $|\userS|\times|K|$ matrix, and $Q$ is a $|K|\times|\itemS|$ matrix. As a result, the factorization gives us a low dimensional numerical representation of users and items. In the case of the stated problem, we want to build the regression model $\tilde{r}_{u,i}$, that predicts the missing values in the matrix $R$ for any arbitrary pair $(u,i)\in |\userS|\times|\itemS|$ :
\begin{equation}
\tilde{r}_{u,i} = p_{u}^{T}q_{i} = \sum_{k=1}^{K}p_{u,k}q_{k,i},
\end{equation}
where $p_{u}^{T}$ is the $u-th$ row of $P$ and $q_{i}$ is the $i-th$ column of $Q$.

The error function between the estimated and real ratings can be calculated as follows:
\begin{equation}\label{eq:x00}
\Loss(P,Q,R) = \sum_{(u,i)\in {\Theta}}(r_{u,i}-\tilde{r}_{u,i})^{2}+\lambda\left(\sum_{u\in \userS}\|p_u\|^{2}+\sum_{i\in \itemS}\|q_i\|^{2}\right),
\end{equation}
where $\Theta$ is the set of all user-item pairs, that have marks in matrix $R$, $\userS$ is the full set of users and $\itemS$ is the full set of items. The second term of the equation is the regularization part which allow to avoid overfitting. We also notice, that a non-convex regulrization in Eq.~\ref{eq:x00} may improve the quality of factorization~\cite{pogodin2017efficient,krechetov2018entropy}; however, computations often become harder.

The regression approach rely on the idea of providing the predictions that will as close as possible from the true score, so the problem leads to the prediction of ratings for each single pair user-item.  
It is important to note that the prediction of ratings is only an intermediate step towards recommendation, and that other directions are possible. In particular, given the typical use of recommender systems where the system presents each user with the top $N$ items without showing the associated ratings, we believe that ordering the items correctly is more important than correctly predicting their ratings. Although these two objectives are close, they are not equivalent from the point of view of the recommendation. Indeed, any method that correctly predicts all ratings will also correctly order all items. On the other hand, with equal performances in terms of rating prediction, two methods can have different performances in terms of order prediction. This phenomenon is illustrated in Figure \ref{fig_reg_vs_ord}.

\begin{figure}
	\centering
		\includegraphics{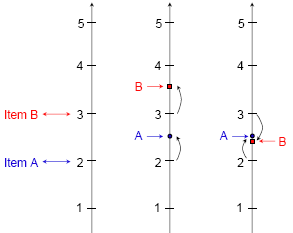}
		\caption[Item ranking illustration]{Let $[ 2, 3 ]$ be the ratings of two items A and B, $r_1 = [ 2.5, 3.6 ]$ and $r_2 = [ 2.5, 2.4 ]$ two vectors of predictions obtained by two different methods. Although $r_1$ and $r_2$ are equivalent in terms of squared error (the two are equal to $0.5^2 + 0.6^2$), only $r_1$ predicts the order correctly, since the score it assigns to B is greater than that of A.}
		\label{fig_reg_vs_ord}
\end{figure}

\subsection{Ranking based approaches for recommendation}
In the followings, we are interested in predicting if a user will prefer an item over another, rather than predicting a real-value  (such as a rating for instance). This task is tackled by ranking based approaches where the goal is to learn a list of items ordered according to their degree of relevance for a given user.

\subsection{Neural Language Models}

Ranking-based approaches for RS, like matrix factorization techniques, rely on the learning of latent representations for users and items. The main difference is that these techniques primarily use neural networks for representation learning and follow  the basic idea of traditional Natural Language Processing (NLP) approaches, which tackled the difficult task of finding the best representations of words to reflect their similarities and differences. 

Using the skip-gram training method (SG, implemented in the word2vec software package\footnote{\url{https://radimrehurek.com/gensim/models/word2vec.html}})  encouraging results were obtained by encoding words as embedding vectors \cite{word_emb,mikolov_13}. Similarly, \cite{levy_14} proposed new opportunities to extend the word representation learning \cite{word_emb,mikolov_13, shazeer2016swivel} to characterise more complicated piece of information. Indeed, the authors of this paper showed the equivalence of the SG model with negative sampling and implicit factorization of a point-wise mutual information (PMI) matrix. Furthermore, they demonstrated that word embedding may be used to a variety of data kinds (not only words) if a suitable context matrix can be created. 

Since then, this idea has been successfully applied to recommendation systems, where different approaches attempted to learn representations of items and users. In \cite{liang_16}, the authors proposed a model that relies on the intuitive idea that pairs of items scored the same way by different users are similar. The approach  reduces to finding both the latent representations of users and items, with the traditional Matrix Factorization (MF) approach, and simultaneously learning item embeddings using a co-occurrence Shifted Positive Pointwise Mutual Information (SPPMI) matrix defined by items and their context.

In \cite{GrbovicRDBSBS15}  the authors proposed Prod2Vec, which embeds items using the word2vec technique, by modelling user purchases as a sequence. This model was then extended in \cite{vasile_16} who, by defining appropriate context matrices, proposed to learn embedding for meta information available in the system. In addition, they demonstrated that the improvement obtained was mainly the result of the ability of their approach to deal with item cold-start. Inspired by the concept of sequence of words; the approach proposed by \cite{guardia_15} defined the consumption of items by users as trajectories. Then, the embedding of items is learned using the Skip-Gram (SG) model and the users embedding is further inferred as to predict the next item in the trajectory. \\

\subsection{Deep Neural Networks architectures for recommmendation}

Different topologies of Deep Neural Networks could be used to handle the challenge of employing context information while taking into consideration the time over interactions in the system. The advantage of neural network design over traditional ranking models is its extensibility. It is easy to start with a simple model and then demonstrate the approach's effectiveness; as it is possible to define what will be effective in a specific task during the experiments. In the following, we will present the most popular neural network based approaches for recommender systems.

\bigskip

\begin{itemize}
    \item \underline{Autoencoders}
    
The network first creates a low-dimensional representation of the user from the data, removing all except the most important information, then decoding the data in its original dimension. As a consequence, a noise-free, averaged representation is created, from which any item's preference may be estimated.
    
    \begin{figure}[H]
    \centering
    \vspace*{-10pt}
    \includegraphics[width=0.8\textwidth]{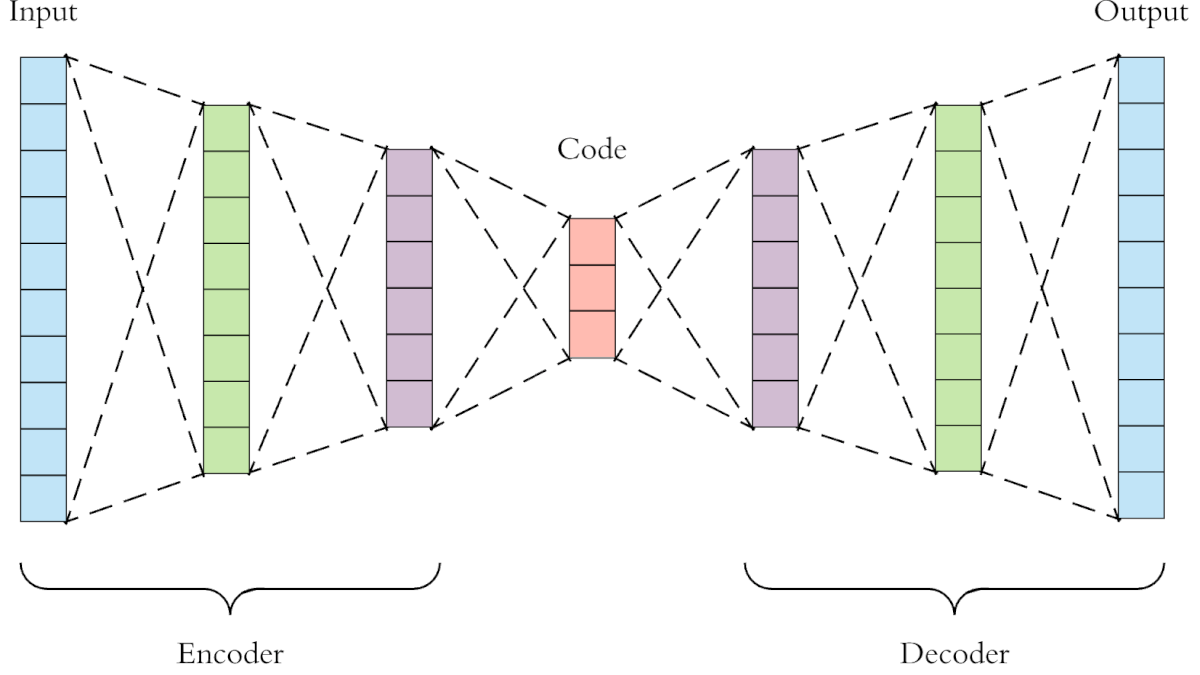}
    \caption[Autoencoder architecture]{Autoencoder architecture; figure is taken from \url{https://towardsdatascience.com}}
    \label{fig:ch1_autoenc}
    \end{figure}
    
    Deep AutoEncoder \cite{kuchaiev2017training}, might be used as an example of an  autoencoder model in a ranking problem (figure \ref{fig:ch1_recautoenc}). During training, sparse vector of rates is taken as the input of the model because there are no users in reality that can estimate all set of items. The model's output is dense, which indicates that the network predicts all of the user's future ratings.
    
    \begin{figure}[H]
    \centering
    \vspace*{-10pt}
    \includegraphics[width=0.8\textwidth]{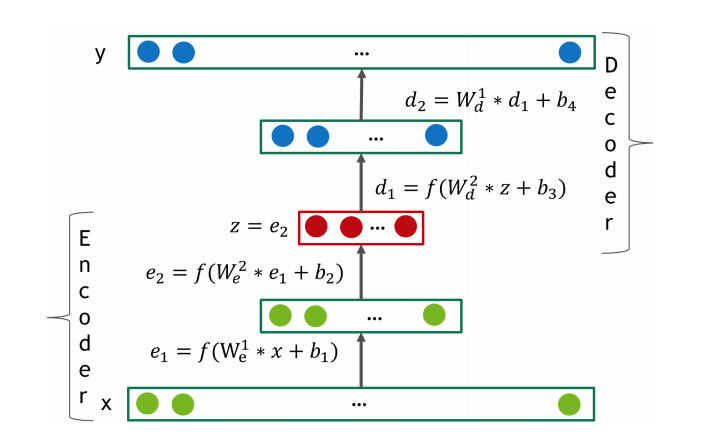}
    \caption{Deep AutoEncoder for Collaborative Filtering}  
    \label{fig:ch1_recautoenc}
    \end{figure}
    
    \item \underline{Convolutional Neural Networks (CNN)}
    
The principle behind convolutional neural networks is that convolutional layers are alternated with non-linear and fully-connected layers. Initially, CNN was used to do effective image recognition, but it currently now performs well in other areas such as ranking problems.
    
    Convolutional neural networks work on the basis of filters (see figure \ref{fig:ch1_convolution}) that are engaged in recognizing certain image characteristics (for example, straight lines). A filter is an ordinary matrix of weights, which are trained. The filter moves along the image and determines whether some desired characteristic is present in a specific part of it, by applying convolution operation, which is the sum of the products of the filter elements and the input signal matrix.
    
    \begin{figure}[H]
    \centering
    \vspace*{-10pt}
    \includegraphics[width=1.0\textwidth]{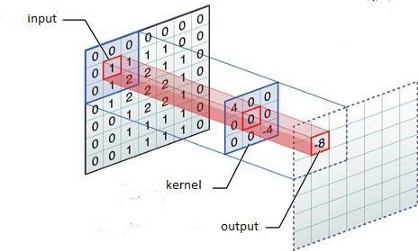}
    \caption{The principle of work convolve operation}  
    \label{fig:ch1_convolution}
    \end{figure}
    
    For applying CNN's in ranking models proposed in chapters \ref{ch:2} - \ref{ch:3} \caser \cite{tang2018caser} as the baseline presented below (figure \ref{fig:ch1_caser}): 
    
    \begin{figure}[H]
    \centering
    \vspace*{-10pt}
    \includegraphics[width=1.0\textwidth]{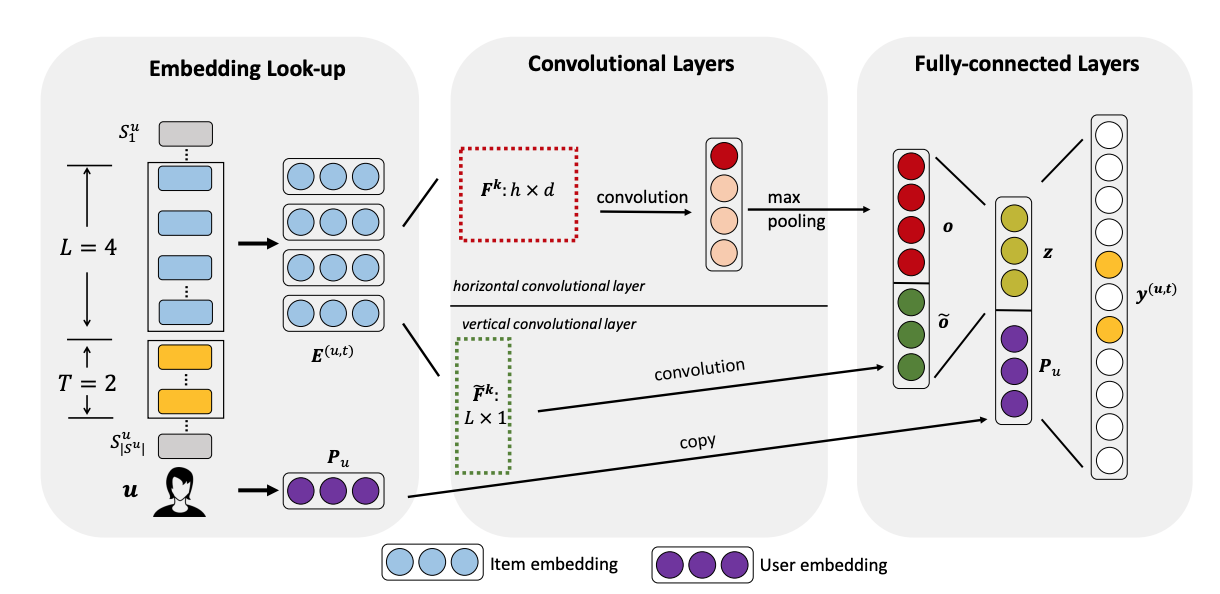}
    \caption{The network architecture of Caser}  
    \label{fig:ch1_caser}
    \end{figure}
    
    The main idea of the model comes from the image recognition field, considering embedding matrix for items as the image and passing the horizontal and vertical convolution filters for searching the local features.
    As the input the model takes $L$ clicked items for each user $u$ we and their next $T$ ($T$ is the parameter) items as the targets.
    To make the recommendations for a user $u$ at time step $t$ using trained \caser, latent user embedding and matrix of trained embeddings for the user's last  $L$ interactions are taken as the input to predict $N$ next items. 
    
    \item \underline{Recurrent Neural Networks (RNN)}
    
    Inside the RNN the basic recurrent cell is located. The model takes input data and pass it through RNN, which has a hidden internal state. This state is updated each time when new data is received in the RNN. Often the task requires that RNN generates some output at each time interval, therefore, after reading the input data and updating the hidden state, the RNN will create the output data.
    
    \begin{figure}[H]
    \centering
    \vspace*{-10pt}
    \includegraphics[width=0.3\textwidth]{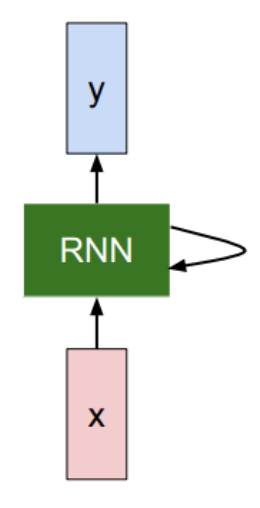}
    \caption{The principle of RNN work. The information is spread from the input to the output with some recursion on the connections between the nodes.}  
    \label{fig:ch1_rnn}
    \end{figure}
    
    So, what is going on in the green part of the figure \ref{fig:ch1_rnn}? Inside the green box, the recurrence relation is calculated:
    
    \begin{equation}
        h_t = f_W(h_{t-1},x_t)
    \end{equation}
    
    To find the new state $h_t$, the previous hidden state $h_{t-1}$ and the current input $x_t$ are taken. When the next input data come to the model, the received hidden state $h_t$ is passed through the same function $f_W$, and the whole process is repeated. If there are sufficiently long input sequences, the network could face with the problem of forgetting the information about remote input objects. But in some cases, there is a necessity for the network to "remember" information about the objects located at the beginning of the sequence. To solve this problem, the modifications of RNN, such as GRU and LSTM were proposed.
    
    The authors of the \GRU{} \cite{hidasi2018recurrent} adapt the idea of using recurrent neural networks in recommender systems. The basical structure of architecture presented in the figure \ref{fig:ch1_gru4rec}, where the main part is a stack of GRU's.
    
    \begin{figure}[H]
    \centering
    \vspace*{-10pt}
    \includegraphics[width=0.8\textwidth]{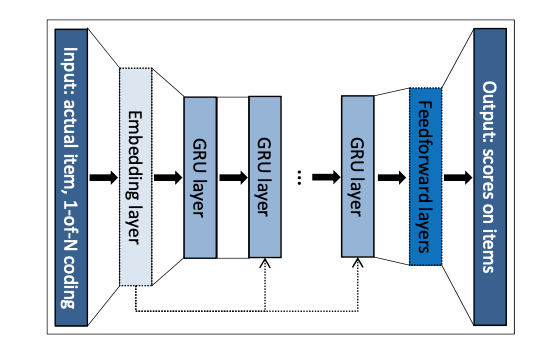}
    \caption{\GRU architecture}  
    \label{fig:ch1_gru4rec}
    \end{figure}
    
    Because of the different session size for each user and the concept of strong time dependency between the interactions inside the session, classical sliding window over the sentences for building mini-batches seems not relevant for recommender systems. That is why authors suggest the strategy from the figure \ref{fig:ch1_gru4rec_sessions}. First mini-batch is represented by the first event for the first $X$ sessions from the full set of time-ordered sessions. For the second mini-batch the second events are used and so on. When any of the session ends, the next available session is taken. 
    
    \begin{figure}[H]
    \centering
    \vspace*{-10pt}
    \includegraphics[width=0.8\textwidth]{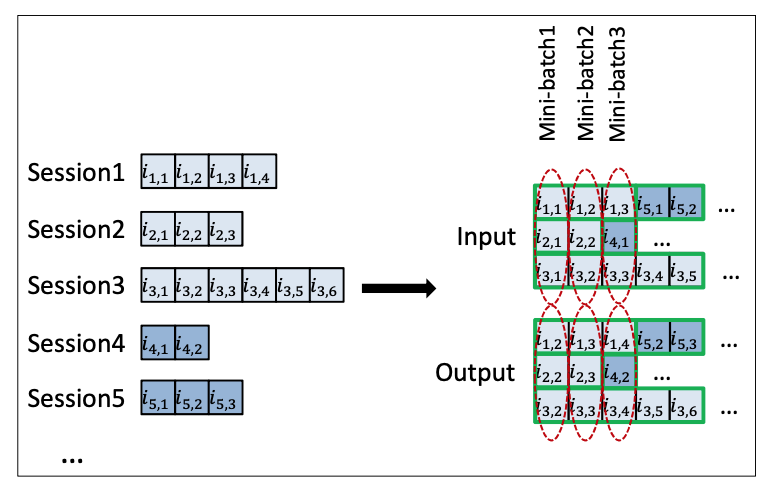}
    \caption{Mini batches for \GRU}  
    \label{fig:ch1_gru4rec_sessions}
    \end{figure}

\bigskip

    \item \underline{Graph Neural Networks (GNN)}
    
The idea lying behind Graph Neural Networks (GNN) is to learn a mapping that represents nodes or entire (sub)graphs as points in a low-dimensional vector space. The goal is to optimize this mapping so that the geometric relations in this studied space display the structure of the original graph.
    
    \begin{figure}[H]
    \centering
    \vspace*{-10pt}
    \includegraphics[width=0.8\textwidth]{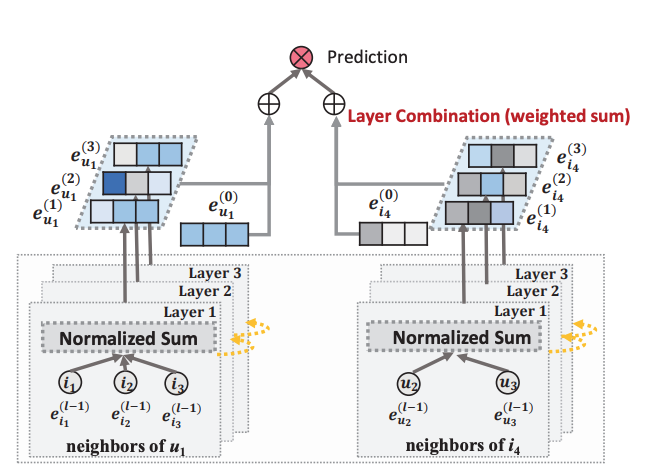}
    \caption{\GCN architecture}  
    \label{fig:ch1_gcn}
    \end{figure}

    The normalized sum of neighbor embeddings of \GCN{} is taken towards next layer. 
    Proposed by \cite{He2020} \GCN{} for recommendations consists of the graph convolution with the discarded feature transformation and nonlinear activation operations (see figure \ref{fig:ch1_gcn}). In layer combination part, the final embedding of the node is constructed as the weighted sum of its embeddings on all layers. These models have demonstrated their ability to handle complicated contextual information, such as item summaries generated by extrative summarization approaches \cite{Amini07}.
    
    \item \underline{Transformers}
    
    The architecture of Transformers also was designed to process sequences as RNN's. But unlike RNN's, transformers do not require processing sequences consecutively. For example, if the input data is text, then the transformer does not need to process the end of the text after processing its beginning. Due to this, transformers are parallelized more easily than RNN's and can be trained faster
    
    \begin{figure}[H]
    \centering
    \vspace*{-10pt}
    \includegraphics[width=0.9\textwidth]{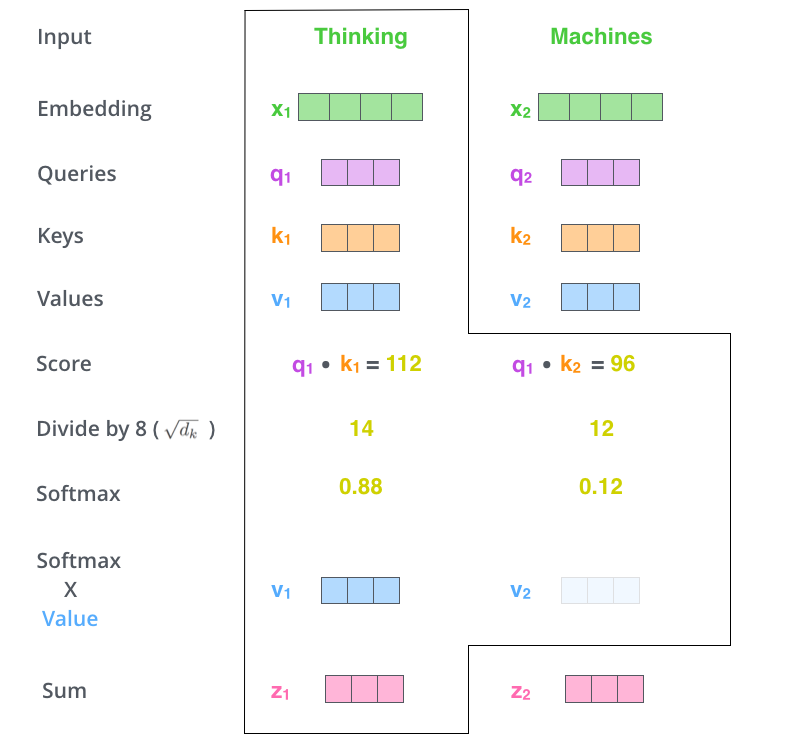}
    \caption{The transformer architecture}  
    \label{fig:ch1_transformer}
    \end{figure}
    
    The main distinctive feature of transformers is in calculating attention, that consists in transforming the embedding vector into three vectors: query, key and value vectors. These vectors are created by multiplying embedding into three matrices that are trained during the training process. As a result, $z$ vector is calculated (see figure \ref{fig:ch1_transformer}) and then it is transmitted further through forward path of the neural network.
    
    \begin{figure}[H]
    \centering
    \vspace*{-10pt}
    \includegraphics[width=0.9\textwidth]{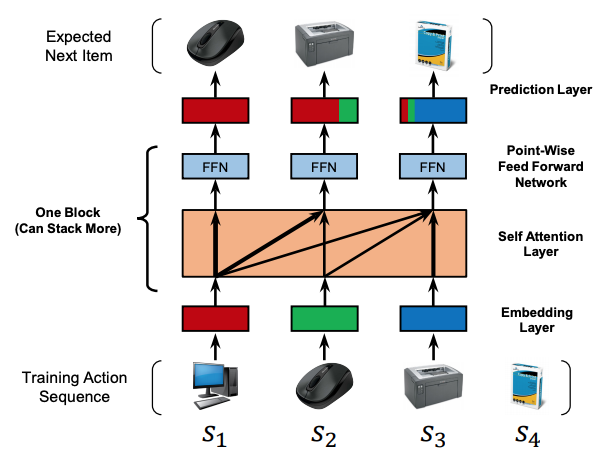}
    \caption{\SASR training structure}  
    \label{fig:ch1_sasrec}
    \end{figure}
    
    In figure \ref{fig:ch1_sasrec} training process for adapted transformer \SASR{} proposed by \cite{Kang18} is presented. At each time step, the model considers all previous items, and apply attention mechanism to focus on the items relevant to the next action.

\end{itemize}

\section{Evaluation metrics}
\label{sec:metrics}

In order to choose the best model from the whole variety of algorithms and approaches, it is necessary to be able to assess their quality quantitatively. In this section, we will present the most common ranking metrics used to evaluate recommender systems.

Consider $N$ users $U=\{u_i\}_{i=1}^{N}$ and $M$ items $E = \{e_j\}_{j=1}^{K}$. The result of the ranking algorithm is the mapping, which assigns to each item $e\in E$ the weight $r(e)$, which characterizes the degree of relevance of this item to the particular user $u\in U$ (the greater the weight, the more relevant the object). That is why the set of weights determines the permutation $\pi$ on the set of items based on their sorting in descending order.

To estimate the quality of the ranking, it is necessary to have some ground true with which the results of the algorithm can be compared. Suggest $r^{true}\in [0,1]$ is the reference relevance characterizes the real relevance of items for a given user ($r^{true}=1$ item is ideal, $r^{true}=0$ - completely irrelevant), $\pi^{true}$ is the corresponding permutation of $r^{true}(e)$.

It is worth noting that when $r^{true}$ takes only extreme values: 0 and 1, the permutation $\pi^{true}$ is usually not considered and only the set of relevant items for which $r^{true}=1$ is taken into account. So, the purpose of the metric is to determine how well the relevance obtained by the algorithm and the corresponding permutation  $\pi$ to the true relevance values $r^{true}$. In the next subsections the main metrics would be considered.

\subsection{Mean Average Precision}

Mean average precision at K ($\text{MAP}@K$) is one of the most frequently used ranking metrics. Precision measures are used in binary problem, where relevance accepted two values: 0 and 1. 
\[
\text{MAP}@K=\frac{1}{N}\sum_{u=1}^{N}\text{AP}@K.
\]
Here the Average Precision at rank $K$, AP$@K$, is defined as :
\[
\text{AP}@K=\frac{1}{K}\sum_{k=1}^K r_{k} Pr(k),
\]
where, $Pr(k)$ is the precision at rank $k$ of the relevant items and $r_k=1$ if the item at rank $k$ is preferred or clicked, and $0$ otherwise.

The idea of $\text{MAP}@K$ is to calculate $\text{AP}@K$ for each user and then take the average. The idea is quite reasonable, assuming that all users are equally important.

\subsection{Normalized Discounted Cumulative Gain}

If the users are not equally important, simple averaging is not enough to express the quality of ranking. In this case, another popular metric $\text{NDCG}@K$ could be applied. To compute $\text{NDCG}@K$, the term called $\text{DCG}@K$ is calculated taking into account the order of the items in the list by multiplying the relevance of the item by a weight equal to the inverse logarithm of the position number.

\[
\text{DCG}@K = \sum_{k=1}^{K}\frac{2^{r_{k}}-1}{\log_{2}(1+k)}.
\]

In contrast to $\text{MAP}@K$, $r_k$ here can also be used in the case of non-binary values of the reference relevance.
The use of the logarithm as a discount function can be explained by the following intuitive considerations: the positions at the beginning of the ranking differ much more than the positions at the end of it. It means that for a user it is more important to have accurate ranking at the first positions, like 1 to 10, and almost not important how correctly items will be distributed between 50 and 60 positions. Normalized version of $\text{DCG}@K$ is called $\text{NDCG}@K$ and computed below:

\[
\text{NDCG}@K = \frac{1}{N}\sum_{u=1}^{N}\frac{\text{DCG}@K}{\text{IDCG}@K},
\]

where $\text{IDCG}@K$ is $\text{DCG}@K$ with an ideal ordering equals to $\sum_{k=1}^{K}\frac{1}{\log_{2}(1+k)}$.

\subsection{Mean reciprocal rank}

Another popular metric $\text{MRR}@K$ define at which position of ranking customer find the first useful recommendation.

\[
\text{MRR}@K = \frac{1}{N}\sum_{u=1}^{N}\text{RR}@K,
\]

$\text{RR}@K$ the value equal to the inverse rank of the first correctly predicted item:

\[
\text{RR}@K = \frac{1}{\text{min}\{k\in {1,..,K}: r_{k}=1\}}.
\]

\subsection{Rank Correlation based metrics}

The rank correlation coefficient takes into account not the values of element's relevances , but only their rank. Below the two most common rank correlation coefficients, the Spearman and Kendall, are presented.

\begin{itemize}
    \item Kendall correlation coefficient:
    
    Consider $\{(x_{1},y_{1}),...,(x_{k},y_{k})\}$ be a set of observations of the joint random variables $X$ and $Y$. 
    Based on the calculation of concordant (and discordant) pairs — pairs of elements to which the permutations have assigned the same (different) order.
    
    \[
    \tau = \frac{|\text{concordant pairs}|-|\text{discordant pairs}|}{K(K-1)/2},
    \]
    
    where concordant pairs are satisfy the condition: both $x_{i}>x_{j}$ and $y_{i}>y_{j}$ hold or both $x_{i}<x_{j}$ and $y_{i}<y_{j}$ hold; otherwise the pairs are discordant.
    
    \item Spearman correlation coefficient:
    
    The Spearman correlation coefficient is defined as the Pearson correlation coefficient between the rank variables.
    
    \[
    r_S = \rho(\pi,\pi^{true}) = \frac{cov(\pi,\pi^{true})}{\sigma_{\pi},\sigma_{\pi^{true}}},
    \]
    
    $\rho$ denotes the usual Pearson correlation coefficient, but applied to the rank variables.
    
\end{itemize}

The quality metrics defined by rank correlation coefficient do not take into account the position of elements and the correlation is calculated for all elements simultaneously, not just for the top-K elements with the highest rank. Therefore, in practice these metrics are applied extremely rarely.

\section{Conclusion}

In this chapter we provided an overview of work on Recommender Systems (RS) that provide personalized recommendations to users by adapting to their taste. The study of RS has become an active area of research these past years, especially since the  Netflix Prize \cite{Bennett07}. One characteristic of online recommendation is the huge unbalance between the available number of products and those shown to the users. Another aspect is the existence of bots that interact with the system by providing too many feedback over some targeted items; or many users that do not interact with the system over the items that are shown to them. In this context, the main challenges concern the design of a scalable and an efficient online RS in the presence of noise and unbalanced data. These challenges have evolved in time with the continuous development of data collections released for competitions or issued from e-commerce\footnote{\scriptsize \url{https://www.kaggle.com/c/outbrain-click-prediction}}.  New approaches for RS now primarily consider {\it implicit} feedback, mostly in the form of clicks, that are easier to collect than {\it explicit} feedback which is in the form of scores. Implicit feedback is more challenging to deal with as they do not depict the preference of a user over items, i.e., (no)click does not necessarily mean (dis)like \cite{hu2008collaborative}. For this case, most of the developed approaches are based on the Learning-to-rank paradigm and focus on how to leverage the click information over the unclick one without considering the sequence of users' interactions. 

~\\~\\~\\~\\~\\~\\~\\~\\~\\~\\~\\~\\~\\
\addcontentsline{toc}{chapter}{Part II: Contribution}
\begin{center}{\Huge \textsc{\underline{Part II}}\\~\\ \textsc{Contribution}}\end{center}\normalsize
\chapter{Sequential Learning over Implicit Feedback for Robust Large-Scale Recommender Systems}
\label{ch:2}
\section{Introduction}

In this chapter, we propose our first contribution which is a new SequentiAl RecOmmender System for implicit feedback (called \SO), that updates the model parameters user per user over blocks of items constituted by a sequence of unclicked items followed by a clicked one. The parameter updates are discarded for users who interact very little or a lot with the system. For other users, the update is done by minimizing the average ranking loss of the current model that scores the clicked item below the unclicked ones in a corresponding block. Recently, many other approaches that model the sequences of users feedback have been proposed, but they all suffer from a lack of theoretical analysis formalizing the overall learning strategy. 
In this work, we analyze the convergence property of the proposed approach and show that in the case where the global ranking loss estimated over all users and items is convex; then the minimizer found by the proposed sequential approach converges to the minimizer of the global ranking loss. 
Experimental results conducted on five large publicly available datasets show that our approach is highly competitive compared to the state-of-the-art models and, it is significantly faster than both the batch and the online versions of the algorithm. 
The results of this chapter were presented at the European Conference in Machine Learning \& Principles ans Practices in Knowledge Discovery (ECML-PKDD) in 2019 \citepers{Burashnikova19} and \textit{Conférence sur l'Apprentissage Automatique} (CAp) in 2021 \citepers{Burashnikova21b}.

The rest of this chapter is organized as follows. Section \ref{sec:soa} relates our work to previously proposed approaches. Section \ref{sec:ch3:Frame} introduces the general ranking learning problem that we address. Then, in Section~\ref{sec:ch3:TA}, we present the \SO{} algorithm and provide an analysis of its convergence. Section \ref{sec:ch3:Exps} presents the experimental results that support this approach. Finally, in Section \ref{sec:ch3:Conclusion}, we discuss the outcomes of this study and give some pointers to further research.

\section{Sequential learning for recommender systems}\label{sec:soa}

Many new approaches tackle the sequential learning problem for RS by taking into account the temporal aspect of interactions directly in the design of a dedicated model and are mainly based on Markov Models (MM), Reinforcement Learning (RL) and Recurrent Neural Networks (RNN) \cite{Donkers:2017}. Recommender systems based on Markov Models, consider the sequential interaction of users as a stochastic process over discrete random variables related to predefined user behavior. These approaches suffer from some limitations mainly due to the sparsity of the data leading to a poor estimation of the transition matrix \cite{GuyShani}. 
Various strategies have been proposed to leverage the impact of sparse data, for example by considering only the last frequent sequences of items and using finite mixture models \cite{GuyShani}, or by combining similarity-based methods with high-order Markov Chains \cite{He}. Although it has been shown that in some cases the proposed approaches can capture the temporal aspect of user interactions but these models suffer from high complexity and generally they do not pass the scale.  Some other methods consider RS as a Markov decision process (MDP) problem and solve it using reinforcement learning (RL) \cite{Moling,Tavakol}. The size of discrete actions bringing the RL solver to a larger class of problems is also a bottleneck for these approaches. Very recently Recurrent neural networks such as GRU or LSTM, have been proposed for personalized recommendations \cite{hidasi2018recurrent,tang2018caser,Kang18},  where the input of the network is generally the current state of the session, and the output is the predicted preference over items (probabilities for each item to be clicked next).

Our proposed strategy differs from other sequential based approaches in the way that the model parameters are updated, at each time a block of unclicked items followed by a clicked one is constituted; and by controlling the number of blocks per user interaction. If for a given user, this number is below or above two predefined thresholds found over the distribution of the number of blocks,  parameter updates for that particular user are discarded. Ultimately, we provide a proof of convergence of the proposed approach.

\section{Framework and Problem Setting}
\label{sec:ch3:Frame}
Throughout, we use the following notation. For any positive integer $n$, $[n]$ denotes the set $[n]\doteq \{1,\ldots,n\}$. We suppose that $\itemS\doteq [M]$ and $\userS\doteq [N]$ are two sets of indexes defined over items and users. Further, we assume that each pair constituted by a user $u$ and an item $i$ is identically and independently distributed according to a fixed yet unknown distribution ${\cal D}_{\cal U, \cal I}$. 

At the end of his or her session, a user $u\in\userS$ has reviewed a subset of items $\itemS_u\subseteq \itemS$ that can be decomposed into two sets: the set of preferred and non-preferred items denoted by $\posI_u$ and $\negI_u$, respectively. Hence, for each pair of items $(i,i')\in\posI_u\times \negI_u$, the user $u$ prefers item $i$ over item $i'$; symbolized by the relation $i\!\prefu\! i'$. From this preference relation a desired output $y_{u,i,i'}\in\{-1,+1\}$ is defined over the pairs $(u,i)\in\userS\times\itemS$ and $(u,i')\in\userS\times\itemS$, such that $y_{u,i,i'}=+1$ if and only if   $i\!\prefu\! i'$.  We suppose that the indexes of users as well as those of items in the set $\itemS_u$, shown to the active user $u\in\userS$, are ordered by time. 

Finally, for each user $u$, parameter updates are performed over blocks of consecutive items where a  block $\bloc_u^t=\text{N}_u^{t}\sqcup\Pi_u^{t}$, corresponds to a time-ordered sequence (w.r.t. the time when the interaction is done) of no-preferred items, $\text{N}_u^{t}$, and at least one preferred one, $\Pi_u^{t}$. Hence, $\posI_u=\bigcup_t \Pi_u^{t}$ and $\negI_u=\bigcup_t \text{N}_u^{t}; \forall u\in\userS$. Notations are summarized in Table~\ref{table:notation}.

 \begin{table}[t!]
\centering
 \begin{tabular}{c|l} 
 $\calI= [M]$ & The set of item indexes\\ \hline
 $\calS= [N]$ & The set of user indexes\\ \hline
 ${\cal D}$ & joint distribution over users and items \\ \hline
 ${\cal D}_{u}$ & conditional distribution of items  for a fixed user $u$\\ \hline
 $\text{N}_u^{t}$ & Negative items in block $t$ for user $u$\\\hline
 $\Pi_u^{t}$ & Positive items in block $t$ for user $u$\\\hline
 $\calB_u^t=\text{N}_u^{t}\sqcup\Pi_u^{t}$ & Negative and positive items in block $t$ for user $u$\\\hline
 $\posI_u$ & The set of all positive items for user $u$\\ \hline
 $\negI_u$ & The set of all negative items for user $u$\\ \hline 
 $\ell_{u, i, i'}(\omega)$ & Instantaneous loss for user $u$ and a pair of items $(i, i')$\\ \hline 
 $\hat{\calL}_u(\omega)$ & Empirical ranking loss with respect to user $u$\\
 & \qquad $\hat{\calL}_u(\omega) = \frac{1}{|\posI_u||\negI_u|}\!\sum_{i\in \posI_u}\!\sum_{i'\in \negI_u} \!\ell_{u,i,i'} (\omega)$\\
 \hline 
 ${\hat {\calL}}_{\calB^t_u}(\omega)$ &  Empirical ranking loss with respect to a block of items \\ 
 & \qquad ${\hat {\calL}}_{\calB^t_u}(\omega) = \frac{1}{|\Pi_u^{t}||\text{N}_u^{t}|}\sum_{i \in \Pi_u^{t}} \sum_{i'\in \text{N}_u^{t}} \ell_{u, i, i'} (\omega)$ \\
 \hline
 ${\calL}(\omega)$ & Expected ranking loss  ${\calL}(\omega) = \mathbb{E}_{{\cal D}_u}\hat{\calL}_u(\omega)$
 \end{tabular}
 \caption{Notation for the proposed \SO{} algorithm and its variants.}
 \label{table:notation}
\end{table}

\section{Proposed Approach}
\label{sec:ch3:TA}
Our objective here is to minimize an expected error penalizing the misordering of all pairs of interacted items $i$ and $i'$ for a user $u$. Commonly, this objective is given under the Empirical Risk Minimization (ERM) principle, by minimizing the empirical ranking loss estimated over the items and the final set of users who interacted with the system~:
\begin{equation}
\label{eq:RLx}
\hat{\Loss}_u(\w)\!=\!\frac{1}{|\posI_u||\negI_u|}\!\sum_{i\in \posI_u}\!\sum_{i'\in \negI_u} \!\ell_{u,i,i'} (\w),
\end{equation}
and $\Loss(\w) = {\mathbb E}_{u} \left[\hat{\cal L}_u(\w)\right]$, where ${\mathbb E}_{u}$ is the expectation with respect to users chosen randomly according to the uniform distribution, and  $\hat{\Loss}_u(\w)$ is the pairwise ranking loss with respect to user $u$'s interactions. As in other studies, we represent each user $u$ and each item $i$ respectively by vectors $\vecU_u\in\R^k$ and $\vecI_i\in\R^k$ in the same latent space of dimension $k$ \cite{Koren:2009}. The set of weights to be found $\w=(\vecU,\vecI)$, are then matrices formed by the vector representations  of users $\vecU=(\vecU_u)_{u\in [N]}\in\R^{N\times k}$ and items $\vecI=(\vecI_i)_{i\in[M]}\in\R^{M\times k}$. 
The minimization of the ranking loss above in the batch mode with the goal of finding user and item embeddings, such that the dot product between these representations in the latent space reflects the best the preference of users over items, is a common approach. Other strategies have been proposed for the minimization of the empirical loss \eqref{eq:RLx}, among which the most popular one is perhaps the Bayesian Personalized Ranking (\BPR) model \cite{rendle_09}. In this approach, the instantaneous loss, $\ell_{u,i,i'}$, is the surrogate regularized logistic  loss for some hyperparameter $\mu \ge 0$:
\begin{equation}
\hspace{-1mm}\ell_{u,i,i'}(\w) =  \log\left(1+e^{-y_{i,u,i'}\vecU_u^\top(\vecI_{i}-\vecI_{i'})}\right) + \mu (\|\vecU_u\|_2^2+\|\vecI_{i}\|_2^2 + \|\vecI_{i'}\|_2^2) \label{eq:instloss}
\end{equation}

The {\BPR} algorithm proceeds by first randomly choosing a user $u$, and then repeatedly selecting two pairs $(i,i')\in \itemS_u\times \itemS_u$. 

In the case where one of the chosen items is preferred over the other one (i.e. $y_{u,i,i'}\in\{-1,+1\}$), the algorithm then updates the weights using the stochastic gradient descent method over the instantaneous loss \eqref{eq:instloss}.  In this case, the expected number of rejected pairs is proportional to  $O(|\itemS_u|^2)$ \cite{Sculley09largescale} which may be time-consuming in general. Another drawback is that user preference over items depend mostly on the context where these items are shown to the user. A user may prefer (or not) two items independently one from another, but within a given set of shown items, he or she may completely have a different preference over these items. By sampling items over the whole set of shown items, this effect of local preference is unclear. 
\subsection{Algorithm \SO{}}
\label{sec:Algo}
Another particularity of online recommendation which is not explicitly taken into account 
by existing approaches is the bot attacks in the form of excessive clicks over some target items. They are made to force the RS to adapt its recommendations toward these target items, or a very few interactions which in both cases introduce biased data for the learning of an efficient RS. In order to tackle these points, our approach 
updates the parameters whenever the number of constituted blocks per user is lower and upper-bounded (Figure \ref{fig:SILICOM}). 

\begin{figure}[t!]
    \centering
    \includegraphics[width=0.95\textwidth]{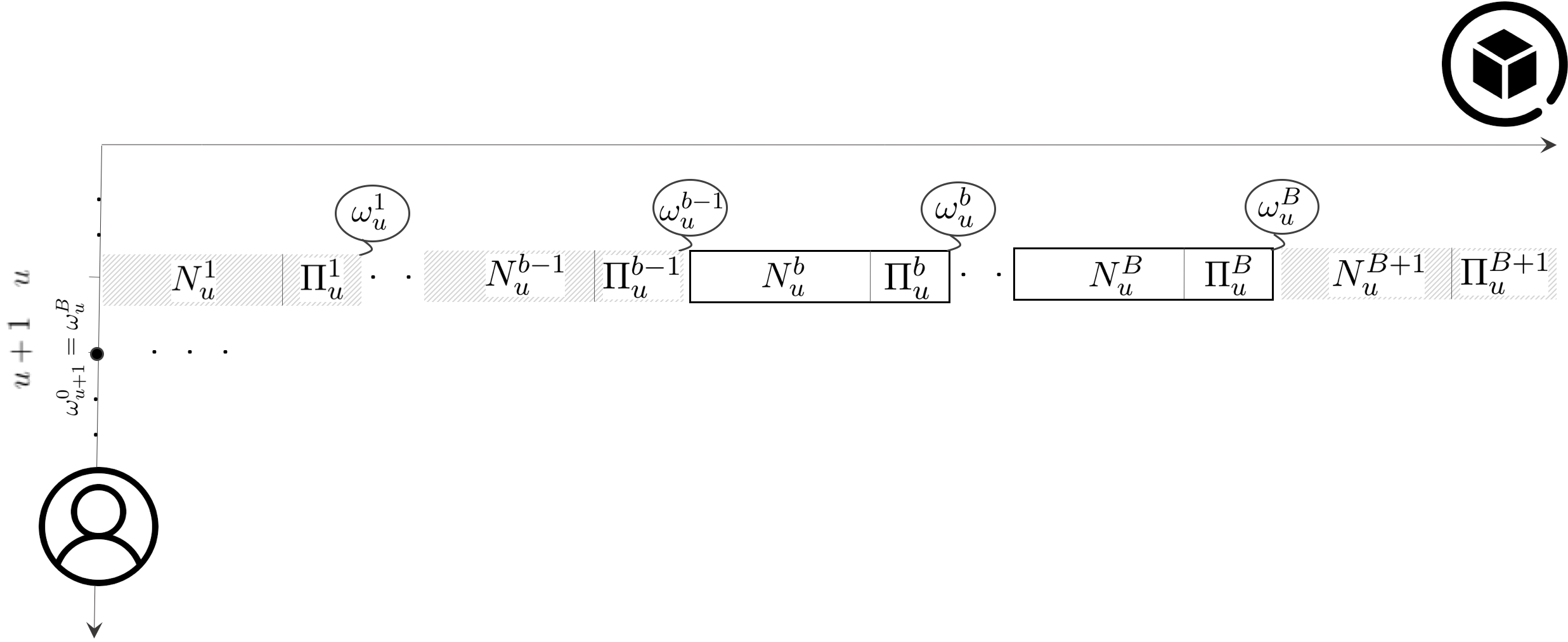}
    \caption[Sequential update strategy]{A pictorial depiction of the sequential updates of weights $(\omega_u^{t})_{1\leq t\leq B}$ for a user $u\in\userS$. The horizontal axis represents the sequence of interactions over items ordered by time. Each update of weights $\omega_u^{t}; t\in\{b,\ldots,B\}$ occurs whenever the corresponding sets of negative interactions, $\text{N}^t_u$, and positive ones, $\Pi_u^t$, exist, and that these number of interactions is lower and upper-bounded. For a new user $u+1$, the initial weights $\omega_{u+1}^0=\omega_u^B$ are the ones obtained from the last update of the previous user's interactions.}  
    \label{fig:SILICOM}
\end{figure}

In this case, at each time  a block $\bloc_u^t=\text{N}_u^{t}\sqcup\Pi_u^{t}$ is formed; weights are updated by miniminzing the ranking loss corresponding to this block~:
\begin{equation}
\label{eq:ch4:CLoss}
{\hat {\cal L}}_{\bloc_u^t}(\w_u^{t}) = \frac{1}{|\Pi_u^{t}||\text{N}_u^{t}|}\sum_{i \in \Pi_u^{t}} \sum_{i'\in \text{N}_u^{t}} \ell_{u, i, i'} ({\w}_u^{t}).
\end{equation}


The pseudo-code of {\SO} is shown in the following. Starting from initial weights $\w_1^0$ chosen randomly for the first user. For each current user $u$, having been shown $I_u$ items, the sequential update rule consists in updating the weights, block by block where after $t$ updates; where the $(t+1)^{th}$ update over the current block $\bloc_u^t=\text{N}_u^{t}\sqcup\Pi_u^{t}$ corresponds to one gradient descent step over the ranking loss estimated on these sets and which with the current weights $\w_u^{t}$ writes,

\begin{equation}
    \label{eq:updaterule}
    \w_u^{t+1} \leftarrow \w_u^t - \eta \nabla {\hat {\cal L}}_{\bloc_u^t}(\w_u^{t})
\end{equation}


\begin{algorithm}[t!]
    \label{alg:CC-conv}
   \caption*{{\bf Algorithm} {\SO}: SequentiAl RecOmmender System}
\begin{algorithmic}
   \State {\bfseries Input:} A time-ordered sequence (user and items)  $\{(u,(i_1, \dots, i_{|I_u|})\}_{u=1}^N$ drawn i.i.d. from~${\cal D}_{{\cal U}, {\cal I}}$
   \State {\bfseries Input:} maximal $B$ and minimal $b$ number of blocks allowed per user $u$
   \State {\bfseries Input:} number of epochs $E$
   \State {\bfseries Input:} initial parameters $\omega_1^0$, and (possibly non-convex) surrogate loss function $\ell(\omega)$
   \For{$e \in E$} 
       \For{$u \in\userS$} 
            \State {Let $\text{N}_u^{t} = \varnothing, \, \Pi_u^t = \varnothing$ be the sets of positive and negative items, counter $t = 0$}
            \For{$i_k\in \itemS_u$} \Comment{\color{gray}Consider all items displayed to user $u$\color{black}}
                \While{$t \leq B$}
                    \If{ $u$ \text{provides a negative feedback on item} $i_k$}
                        \State $\text{N}_u^t \leftarrow \text{N}_u^t \cup \{i_k\}$
                    \Else
                    \State $\Pi_u^t \leftarrow \Pi_u^t \cup \{i_k\}$
                    \EndIf
                    \If{ $\text{N}_u^{t} \neq \varnothing$ and $\Pi_u^{t} \neq \varnothing$ and $t
                    \leq B$}
                        \State {\small $\w_u^{t+1} \leftarrow \w_u^t - \frac{\eta}{|{\text{N}_u^{t}}||\Pi_u^{t}|}  \displaystyle{\sum_{i\in\Pi_u^{t}}\sum_{i'\in \text{N}_u^{t}}} \nabla \ell_{u,i,i'} (\w_{u}^t)$}
                        \State $t = t+1, \text{N}_u^t = \varnothing$, $\Pi_u^t = \varnothing$
                    \EndIf
                \EndWhile
            \EndFor
            \If{$t \geq b$}
                \State ${\omega}_{u+1}^0 = {\omega}_{u}^t$
            \Else
                \State ${\omega}_{u+1}^0 = {\omega}_{u}^0$
            \EndIf
       \EndFor
    \EndFor
    \State {\bfseries Return:} 
    ${\bar\omega}_N = \sum_{u\in\userS} \omega_u^0$ 
\end{algorithmic}
\end{algorithm}


To prevent from a very few interactions or from bot attacks, two thresholds $b$ and $B$ are fixed over the parameter updates. For a new user $u+1$, the parameters are initialized as the last updated weights from the previous user's interactions in the case where the corresponding number of updates $t$ was in the interval $[b,B]$; i.e. $\omega^0_{u+1}=\omega^t_u$. On the contrary case, they are set to the same previous initial parameters; i.e., $\omega^0_{u+1}=\omega^0_u$.

\subsection{Convergence analysis}
\label{sec:TA}

We provide proofs of convergence for the {\SO} algorithm under the typical hypothesis that the system is not instantaneously affected by the sequential learning of the weights. This hypothesis stipulates that the generation of items shown to users is independently and identically distributed with respect to some stationary in time underlying distribution~${\cal D}_{\cal I}$, and constitutes the main hypothesis of almost all the existing studies. Furthermore, we make  the following technical assumption.

\begin{assumption}\label{asmp:smooth}
Let the loss functions $\ell_{u, i, i'}(\omega)$ and ${\cal L}(\omega)$, $\omega\in\mathbb{R}^d$ be such that for some absolute constants $\gamma\ge \beta > 0$ and $\sigma>0$~:
\begin{enumerate}
    \item $\ell_{u, i, i'}(\omega)$ is non-negative for any user $u$ and a pair of items $(i, i')$;
    \item $\ell_{u, i, i'}(\omega)$ is twice continuously differentiable, with a continuous Lipschitz gradient for both instantaneous loss and the ranking loss (Chapter \ref{ch:0} definition \ref{def:Lipschitz}). That is for any user $u$ and a pair of items $(i,i')$ we have
        $ \|\nabla \ell_{u, i, i'}(\omega) - \nabla \ell_{u, i, i'}(\omega')\|_2\leq \gamma \|\omega - \omega'\|_2  , 
        $
        as well as $
         \|\nabla {\cal L}(\omega) -  \nabla
        {\cal L}(\omega')\|_2 \leq \beta \|\omega - \omega'\|_2, 
    $.
    \item Variance of the empirical loss is bounded
    $
        \mathbb{E}_{\cal D}\left\|\nabla \hat{\cal L}_u(\omega) - \nabla {\cal L}(\omega)\right\|_2^2 \le \sigma^2. 
    $
\end{enumerate}
Moreover, there exist some positive lower and upper bounds $b$ and $B$, such that the number of updates for any $u$ is within the interval $[b, B]$ almost surely.
\end{assumption}

Our main result in this chapter is the following theorem which provides a bound over the deviation of the  ranking loss  with respect to the sequence of weights found by the {\SO} algorithm and its minimum in the case where the latter is convex.

\begin{theorem}[\cite{Burashnikova19}]\label{thm:10}
Let $\ell_{u, i, i'}(\omega)$ and ${\cal L}(\omega)$ satisfy Assumption~\ref{asmp:smooth}.
Then for any constant step size $\eta$, verifying $0< \eta \le \min(\frac{1}{\beta B}, 1/\sqrt{UB(\sigma^2 + 3\gamma^2/b)})$, and any set of users $\userS\doteq [U]$; algorithm {\SO} iteratively generates a sequence $\{\omega_u^0\}_{u\in \userS}$ such that 
\begin{gather*}
    \frac{1}{\beta} \mathbb{E}\|\nabla {\cal L}({\omega}_u^0)\|_2^2
    \le
    \frac{\beta B \Delta_{\cal L}^2}{u} + 2\Delta_{\cal L}\sqrt{\frac{B\sigma^2 + 3B\gamma^2/b}{u}}, 
\end{gather*}
where $\Delta_{\cal L}^2 = {\frac{2}{\beta}({\cal L}(\omega_0) - {\cal L}(\omega^*))}$, and the expectation is taken with respect to users chosen randomly according to the uniform distribution $p_u=\frac{1}{N}$.

Furthermore, if the ranking loss ${\cal L}(\omega)$ is convex, then for the sequence $\{\omega_u^0\}_{u\in \userS}$ generated by algorithm {\SO} and $\bar\omega_u = \sum_{j\le u} \omega_j^0$ we have
\begin{gather*}
    {\cal L}({\bar \omega}_u) - {\cal L}({\omega_*}) \le \frac{\beta B \Delta_\omega^2}{u} + 2\Delta_{\omega}\sqrt{\frac{B\sigma^2 + 3B\gamma^2/b}{u}},
\end{gather*}
where $\Delta_{\omega} = \|\omega_0 - \omega_*\|_2^2$, and $\omega_*=\mathop{\argmin}_\omega {\cal L}({\omega})$.
\end{theorem}

\begin{Proof}
{\it Sketch.} Expectation of the empirical loss taken over a random block ${\cal B}_u^t$ for a user $u$, equals to the expected loss for this user. Then by the law of total expectation 
one has $
  \mathbb{E}_{{\cal D}_u} \left[
        k^{-1}\sum_{l=1}^k \nabla {\hat {\cal L}}_{\bloc^l_u}(\w)\right]
     = \nabla \hat{\cal L}_u(\omega)
$, where ${\cal D}_{u}$ is the conditional distribution of items for a fixed user $u$. The variance of the gradient estimation over $k$ blocks is bounded by $3\gamma^2/k$, as for any block after the next to ${\cal B}_u^t$ and before the previous to ${\cal B}_u^t$ are conditionally independent for any fixed ${\cal B}_u^t$. 

Let $g_u^t$ be a gradient of the loss function taken for user $u$ over block ${\cal B}^t_u$:
\[
    g_u^t = \frac{1}{|N_u^t||\Pi_u^t|} \sum_{i \in N_u^t, i' \in \Pi_u^t} \nabla \ell_{u, i, i'} (\omega_u^{t-1}),
\]
According to the notation of Algorithm {\SO} let $\delta_u^t = g_u^t - \nabla {\cal L}(\omega_u^{0})$ and 
$\omega_u^{t+1} = \omega_u^t - \eta g_u^t$,
        $\omega_{u+1}^{0} = \omega_u^{|{\cal B}_u|}$, and 
    $\omega_{u+1}^{0} - \omega_u^{0} = \eta \sum_{t\in {\cal B}_u} g_u^t$,
where ${\cal B}_u$ is the set of all interacted blocks corresponding to user $u$. Using the smoothness of the loss function implied by Assumption~\ref{asmp:smooth}, it comes :
\begin{align}\label{eq:proof01}
    {\cal L}(\omega_{u+1}^0)
    &
        = {\cal L}(\omega_{u}^0) - \left(\hat\eta_u - \frac{\beta}{2} {\hat\eta}^2_u \right) \|\nabla {\cal L}(\omega_{u}^0)\|_2^2  \nonumber\\
    &
         - (\hat\eta_u - \beta {\hat\eta}^2_u) \sum_{t \in {\cal B}_u}\left\langle \nabla {\cal L}(\omega_{u}^0), \frac{\delta_u^t}{|{\cal B}_u|}\right\rangle + \frac{\beta}{2}{\hat\eta}^2_u \sum_{t\in {\cal B}_u} \left\|\frac{\delta_u^t}{|{\cal B}_u|}\right\|_2^2 
\end{align}
where $\hat\eta_u = |{\cal B}_u|\eta$. Then by re-arranging and summing up, we get
\begin{align*}
    \sum_{u=1}^N & \left(\hat\eta_u -  \frac{\beta}{2}{\hat\eta}^2_u\right) \|\nabla {\cal L}(\omega_u)\|_2^2 \le {\cal L}(\bar \omega_u) - {\cal L}(\omega^*) \\
    &  - \sum_{u=1}^{N} (\hat\eta_u - \beta\hat\eta^2_u)\left\langle\nabla{\cal L}(\omega_u), \sum_{t\in {\cal B}_u} \frac{\delta_u^t}{|{\cal B}_u|}\right\rangle + \frac{\beta}{2}\sum_{u=1}^N \hat\eta_u^2 \left\|\sum_{t\in {\cal B}_u}\frac{\delta_u^t}{|{\cal B}_u|}\right\|_2^2
\end{align*}

As the stochastic gradient taken with respect to a block of items gives an unbiased estimate of the gradient, thus
\begin{align}\label{eq:thm01-proof02}
    \mathbb{E}_{{\cal D}_u}\biggl[\biggl\langle\nabla{\cal L}(\omega_u), \sum_{t\in {\cal B}_u} \frac{\delta_u^t}{|{\cal B}_u|}\biggr\rangle \bigg| \xi_u\biggr] = 0,
\end{align}
where $\xi_u$ is a set of users preceding $u$. As in the conditions of the theorem $b\le |{\cal B}_u|$ almost surely, by  the law of total variation, $\Var \psi = \mathbb{E}[\Var(\psi|\eta)] + \Var[\mathbb{E}[\psi|\eta]]$:
\begin{gather}\label{eq:thm01-proof03}
    \mathbb{E}_{{\cal D}_u}\,\left\|\sum_{t\in {\cal B}_u}\frac{\delta_u^t}{|{\cal B}_u|}\right\|_2^2 \le \sigma^2 + \frac{3\gamma^2}{b} 
\end{gather}
where the first term on the right-hand side of Eq.~\eqref{eq:thm01-proof03} comes from Assumption~\ref{asmp:smooth}, and the second term is due to the variance estimate. Condition $\beta \eta B \le 1$ implies $\hat\eta_u - {\beta}{\hat\eta}^2_u/2 \ge \hat\eta_u/2$, thus
\begin{align*}
    \frac{1}{\beta}\mathbb{E}_{\cal D}\,\|\nabla {\cal L}(\omega)\|_2^2 \le
    \frac{1}{\sum_{u=1}^N\hat\eta_u}\left[
    \frac{2({\cal L}(\omega_0) - {\cal L}(\omega_*))}{\beta} + \left(\sigma^2 + 3\frac{\gamma^2}{b}\right) \sum_{u=1}^N\hat\eta^2_u
    \right]
\end{align*}

The rest of the proof of the theorem comes along the same lines according to the randomized stochastic gradient descent analysis~\cite{ghadimi2013stochastic}.
\end{Proof}

The full proof is provided in the Supplementary. This result implies that the loss over a sequence of weights $(\omega_u^0)_{u\in \userS}$ generated by the algorithm converges to the true minimizer of the ranking loss ${\cal L}(\omega)$ with a rate proportional to $O({1}/{\sqrt{u}})$. The stochastic gradient descent strategy implemented in the Bayesian Personalized Ranking model (\BPR) \cite{rendle_09} also converges to the minimizer of the ranking loss ${\cal L}(\omega)$ with the same rate. However, the main difference between \BPR{} and \SO{} is their computation time. As stated previously, the expected number of rejected random pairs sampled by algorithm \BPR{} before making one update is $O(|\itemS_u|^2)$ while with \SO{}, blocks are created sequentially as and when users interact with the system.  For each user $u$, weights are updated whenever a block is created, with the overall complexity of   $O(\max_t (|\Pi_u^t|\times |\text{N}_u^t|))$, with $\max_t (|\Pi_u^t|\times |\text{N}_u^t|)\ll|\itemS_u|^2$.
\section{Experimental Setup and Results}
\label{sec:ch3:Exps}

In this section, we provide an empirical evaluation of our optimization strategy on some popular benchmarks proposed for evaluating RS. All subsequently discussed components were implemented in Python3 using the TensorFlow library \footnote{\url{https://www.tensorflow.org/}.} and computed on Skoltech CDISE HPC cluster \textquote{Zhores} \cite{DBLP:journals/corr/abs-1902-07490}. We first proceed with a presentation of the general experimental set-up, including a description of the datasets and the baseline models.

\paragraph{Datasets. } We report results obtained on five publicly available data\-sets, for the task of personalized Top-N recommendation on the following collections~:
\begin{itemize}
\item \ML-1M \cite{Harper:2015:MDH:2866565.2827872} and {\NetF} \cite{Bennett07} 
consist of user-movie ratings, on a scale of one to five, collected from a movie recommendation service and the Netflix company. The latter was released to support the Netflix Prize competition \cite{Bennett07}. 
For both datasets, we consider ratings greater or equal to $4$ as positive feedback, and negative feedback otherwise.
\item  We extracted a subset out of the {\Out} dataset from of the Kaggle challenge\footnote{\url{https://www.kaggle.com/c/outbrain-click-prediction}} that consisted in the recommendation of news content to users based on the 1,597,426 implicit feedback collected from multiple publisher sites in the United States.
\vspace{1mm}\item  {\kasandr}\footnote{\url{https://archive.ics.uci.edu/ml/datasets/KASANDR}} dataset \cite{sidana17}  
contains 15,844,717 interactions of 2,158,859 users in Germany using Kelkoo's (\url{http://www.kelkoo.fr/}) online advertising platform.
\item {\PANDOR}\footnote{\url{https://archive.ics.uci.edu/ml/datasets/PANDOR}} is another publicly available dataset for online recommendation \cite{sidana18}  provided by Purch (\url{http://www.purch.com/}). The dataset records 2,073,379 clicks generated by 177,366 users of one of the Purch's high-tech website over 9,077 ads they have been shown during one month. 
\end{itemize}
\noindent Table \ref{tab:ch4:datasets} presents some detailed statistics about each collection. Among these, we report the average number of positive (click, like) feedback and the average number of negative feedback. As we see from the table,  {\Out}, {\kasandr}, and {\PANDOR} datasets are the most unbalanced ones in regards to the number of preferred and non-preferred items.
\begin{table}[th]
    \centering
    \begin{tabular}{lccccc}
    \hline
    Data&$|\mathcal{U}|$&
    $|\mathcal{I}|$ & Sparsity&~~~~Avg. \# of $+$~~~~ & ~Avg. \# of $-$~~~~\\
    \hline
    {\ML}-1M&6,040&3,706&.9553&95.2767& 70.4690\\
   \Out&49,615&105,176&.9997&6.1587& 26.0377 \\
   \PANDOR&177,366&9,077&.9987&1.3266& 10.3632\\
     \NetF&90,137&3,560&.9914&26.1872& 20.2765\\
   \kasandr&2,158,859&291,485&.9999&2.4202& 51.9384\\
        \hline
    \end{tabular}
    \caption[Statistics on {\ML}-1M, {\NetF}, {\Out}, {\kasandr} and {\PANDOR} datasets]{Statistics on the \# of users and items; as well as the sparsity and the average number of $+$ (preferred) and $-$ (non-preferred) items on {\ML}-1M, {\NetF}, {\Out}, {\kasandr} and {\PANDOR} collections after preprocessing.}
    \label{tab:ch4:datasets}
\end{table}
\begin{table}[th]
    \centering
    \begin{tabular}{lcccc}
    \hline
    Dataset&$|S_{train}|$~~~~&~~~$|S_{test}|$~~&~~$pos_{train}$~~&~~$pos_{test}$\\
    \hline
    {\ML}-1M&797,758&202,451&58.82&52.39\\
     \Out&1,261,373&336,053&17.64&24.73\\
    \PANDOR&1,579,716&493,663&11.04&12.33\\
    \NetF&3,314,621&873,477&56.27&56.70\\
    \RecS&5,048,653&1,281,909&17.07&13.81\\
    \kasandr&12,509,509&3,335,208&3.36&8.56\\
    \hline
    \end{tabular}
    \caption[Interaction feedback statistics]{Number of interactions used for train and test on each dataset, and the percentage of positive feedback among these interactions.}
    \label{tab:detail_setting}
\end{table}

To construct the training and the test sets, we discarded users who did not interact over the shown items and sorted all interactions according to time-based on the existing time-stamps related to each dataset. Furthermore, we considered $80\%$ of each user's first interactions (both positive and negative) for training, and the remaining for the test. Table \ref{tab:detail_setting} presents the size of the training and the test sets as well as the percentage of positive feedback (preferred items) for all collections ordered by increasing training size. The percentage of positive feedback is inversely proportional to the size of the training sets, attaining $3\%$ for the largest, \kasandr{} collection. 

We also analyzed the distributions of the number of blocks and their size for different collections. Figure \ref{fig:boxplots} (left) shows boxplots representing the logarithm of the number of blocks through their quartiles for all collections. From these plots, it comes out that the distribution of the number of blocks on {\PANDOR}, {\NetF} and {\kasandr} are heavy-tailed with more than the half of the users interacting no more than 10 times with the system. Furthermore, we note that on {\PANDOR} the average number of blocks is much smaller than on the two other collections; and that on all three collections the maximum numbers of blocks are $10$ times more than the average. These plots suggest that a very small number of users (perhaps bots) have an abnormal interaction with the system generating a huge amount of blocks on these three collections. To have a better understanding, Figure \ref{fig:boxplots} (right) depicts the number of blocks concerning their size on {\kasandr}. The distribution of the number of blocks follows a power law distribution and it is the same for the other collections that we did not report for the sake of space. In all collections, the number of blocks having more than $5$ items drops drastically. As the {\SO} does not sample positive and negative items for updating the weights, these updates are performed on blocks of small size, and are made very often. 

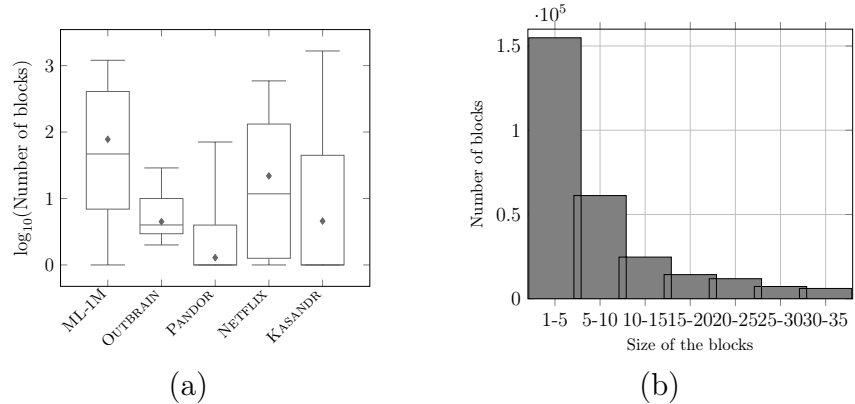
\begin{figure}[!ht]
    \begin{center}
    \begin{tabular}{ccc}
    \begin{tikzpicture}[scale=0.6]
  \begin{axis}
    [ 
    ylabel = {$\log_{10}(\text{Number of blocks})$},
    boxplot/draw direction=y,
    xtick={1,2,3,4,5},
    xticklabels={\ML-1M, \Out,\PANDOR,\NetF,\kasandr},
    x tick label style={font=\footnotesize, rotate=45,anchor=east}
    ]
    \addplot+[mark = *, mark options = {black!30},
    boxplot prepared={
      lower whisker=0,
      lower quartile=0.84,
      median=1.67,
      average = 1.89,
      upper quartile=2.61,
      upper whisker=3.08
    }, color = black!60
    ]coordinates{};
    \addplot+[mark = *, mark options = {black!30},
    boxplot prepared={
      lower whisker=0.301,
      lower quartile=0.47,
      median=0.602,
      average = 0.651,
      upper quartile=1,
      upper whisker=1.46
    }, color = black!60
    ]coordinates{};
       \addplot+[mark = *, mark options = {black!30},
    boxplot prepared={
      lower whisker=0,
      lower quartile=0,
      median=0,
      average = 0.11,
      upper quartile=0.6,
      upper whisker=1.85
    }, color = black!60
    ]coordinates{};
        \addplot+[mark = *, mark options = {black!30},
    boxplot prepared={
      lower whisker=0,
      lower quartile=0.1,
      median=1.07,
      average = 1.34,
      upper quartile=2.12,
      upper whisker=2.77
    }, color = black!60
    ]coordinates{};
        \addplot+[mark = *, mark options = {black!30},
    boxplot prepared={
      lower whisker=0,
      lower quartile=0,
      median=0,
      average = 0.66,
      upper quartile=1.65,
      upper whisker=3.22
    }, color = black!60
    ]coordinates{};
    \end{axis}
\end{tikzpicture}
& ~~~~& 
\begin{tikzpicture}[scale=0.63]
\vspace{-8mm}\begin{axis}
[ 
    ybar,
 xmajorgrids, 
 bar width = {2.675em},
 yminorticks=true, 
 ymajorgrids, 
 yminorgrids,
 ylabel={Number of blocks},
 xlabel={Size of the blocks},
 ymin = 0.0,
 ymax= 160000.0,
 symbolic x coords={1-5, 5-10, 10-15, 15-20, 20-25, 25-30, 30-35, 35-40, 40-45},
 xtick=data,
label style={font=\footnotesize},
 ];
 \addplot [fill=gray] coordinates {
      (1-5, 154873)
      (5-10, 61243) 
      (10-15, 24733)
      (15-20, 14330)
      (20-25, 11907)
      (25-30, 7281)
      (30-35, 6120)
    };
\end{axis}
\end{tikzpicture}\\
(a) &~& (b) 
\end{tabular}
\vspace{-10pt}
        \end{center}
    \caption[Statistics of {\kasandr} dataset]{(a) Boxplots depicting the logarithm of the number of  blocks through their quartiles for all collections. The median (resp. mean) is represented by the band (resp. diamond) inside the box. The ends of the whiskers represent the minimum and the maximum of the values. (b) Distributions of negative feedback over the blocks in the training set on  {\kasandr}.}
    \label{fig:boxplots}
    \vspace*{-20pt}
        \end{figure}
        

\paragraph{Compared approaches. } To validate the sequential approach described earlier, we compared the proposed \SO{} algorithm\footnote{The code is available on \url{https://github.com/SashaBurashnikova/SAROS}.} with the following methods:  
\begin{itemize}
    \item {\MostPop} is a non-learning based approach which consists in recommending the same set of popular items to all users.
    \item Matrix Factorization ({\MF}) \cite{Koren08}, is a factor model which decomposes the matrix of user-item interactions into a set of low dimensional vectors in the same latent space, by minimizing a regularized least square error between the actual value of the scores and the dot product over the user and item representations. 
    \item {\BPR} \cite{rendle_09} corresponds to the model described in the problem statement above, a stochastic  gradient-descent  algorithm, based  on  bootstrap  sampling  of  training  triplets, and {\batch} the batch version of the model which consists in finding  the model parameters $\w=(\vecU,\vecI)$ by minimizing the global ranking loss over all the set of triplets simultaneously (Eq.~\ref{eq:RLx}). 
   \item {\ProdVec} \cite{GrbovicRDBSBS15},  learns the representation of items using a Neural Networks based model, called word2vec \cite{word_emb}, and performs next-items recommendation using the similarity between the representations of items.
   
    \item {\GRU}$+$ \cite{hidasi2018recurrent} is an extended version of {\GRU} \cite{hidasi} adopted to different loss functions, that applies recurrent neural network with a GRU architecture for session-based recommendation. The approach considers the session as the sequence of clicks of the user that depends on all the previous ones for learning the model parameters by optimizing a regularized approximation of the relative rank of the relevant item which favors the preferred items to be ranked at the top of the list.
    
    \item {\caser} \cite{tang2018caser} is a CNN based model that embeds a sequence of interactions into a temporal image and latent spaces and find local characteristics of the temporal image using convolution filters. 
    
    \item {\SASR} \cite{Kang18} uses  an attention mechanism to capture long-term semantics and then predicts the next item to present based on a user's action history.
\end{itemize}

Hyper-parameters of different models and the dimension of the embedded space for the representation of users and items; as well as the regularisation parameter over the norms of the embeddings for {\BPR}, {\batch}, {\MF}, {\caser} and {\SO} approaches were found by cross-validation. We fixed $b$ and $B$, used in {\SO}, to respectively the minimum and the average number of blocks found on the training set of each corresponding collection. With the average number of blocks being greater than the median on all collections, the motivation here is to consider the maximum number of blocks by preserving the model from the bias brought by the too many interactions of the very few number of users. For more details regarding the exact values for the parameters, see the Table \ref{parameters}.

\begin{table}[!ht]
    \centering
    \begin{tabular}{c|ccccc}
    \hline
    Parameter&{\ML}&{\Out}&{\PANDOR}&{\NetF}&{\kasandr}\\
    \hline
    $B$&78&5&2&22&5\\
    $b$&1&2&1&1&1\\
    Learning rate&.05&.05&.05&.05&.4\\
    \hline
    \end{tabular}
    \caption{{\SO} parameters values.}
    \label{parameters}
\end{table}

\paragraph{Evaluation setting and results. }

We begin our comparisons by testing {\batch}, \BPR{} and \SO{} approaches over the logistic ranking loss (Eq. \ref{eq:instloss}) which is used to train them. Results on the test, after training the models 30 minutes and at convergence are shown in Table \ref{test_loss}. {\batch} (resp. {\SO}) techniques have the worse (resp. best) test loss on all collections, and the difference between their performance is larger for bigger size datasets. 

\begin{table}[ht]
\centering
{\begin{tabular}{cc|ccccc|ccccc}
\hline
\multirow{3}{*}{Dataset}&\multicolumn{8}{c}{Test Loss, Eq.~\eqref{eq:RLx}}\\
\cline{3-12}
&~~&\multicolumn{5}{c|}{30 min}&\multicolumn{5}{c}{At convergence}\\
\cline{3-12}
&~~&\batch&~~&{\BPR}&~~&\SO&\batch&~~&{\BPR}&~~&\SO\\
\hline
\ML-1M &~~&0.751&~~&0.678&~~&\bf{0.623}&0.744&~~&0.645&~~&\bf{0.608}\\
\Out &~~&0.753&~~&0.650&~~&\bf{0.646}& 0.747&~~&0.638&~~&\bf{0.635}\\
\PANDOR &~~&0.715&~~&0.671&~~&\bf{0.658}&0.694&~~&0.661&~~&\bf{0.651}\\
\NetF &~~&0.713&~~&0.668&~~&\bf{0.622}&0.694&~~&0.651&~~&\bf{0.614}\\
\kasandr &~~&0.663&~~&0.444&~~&\bf{0.224}&0.631&~~&0.393&~~&\bf{0.212}\\
\hline
\end{tabular}
}
\caption[Empirical comparison of \BPR, \batch{} and \SO{}]{Comparison between \BPR, \batch{} and \SO{} approaches in terms on test loss after $30$ minutes of training and at convergence.}
\label{test_loss}
\end{table}

These results suggest that the local ranking between preferred and no-preferred items present in the blocks of the training set reflects better  the preference of users than the ranking of random pairs of items or their global ranking without this contextual information. Furthermore, as in {\SO} updates occur after the creation of a block, and that the most of the blocks contain very few items (Figure \ref{fig:boxplots} - right),  weights are updated more often than in {\BPR} or {\batch}. This is depicted in Figure \ref{fig:losses} which shows the evolution of the training error over time for {\batch}, \BPR{} and \SO{} on all collections. As we can see, the training error decreases in all cases, and theoretically, the three approaches converge to the same minimizer of the ranking loss (Eq. \ref{eq:RLx}). However, the speed of convergence is much faster with {\SO}.

\begin{flushleft}
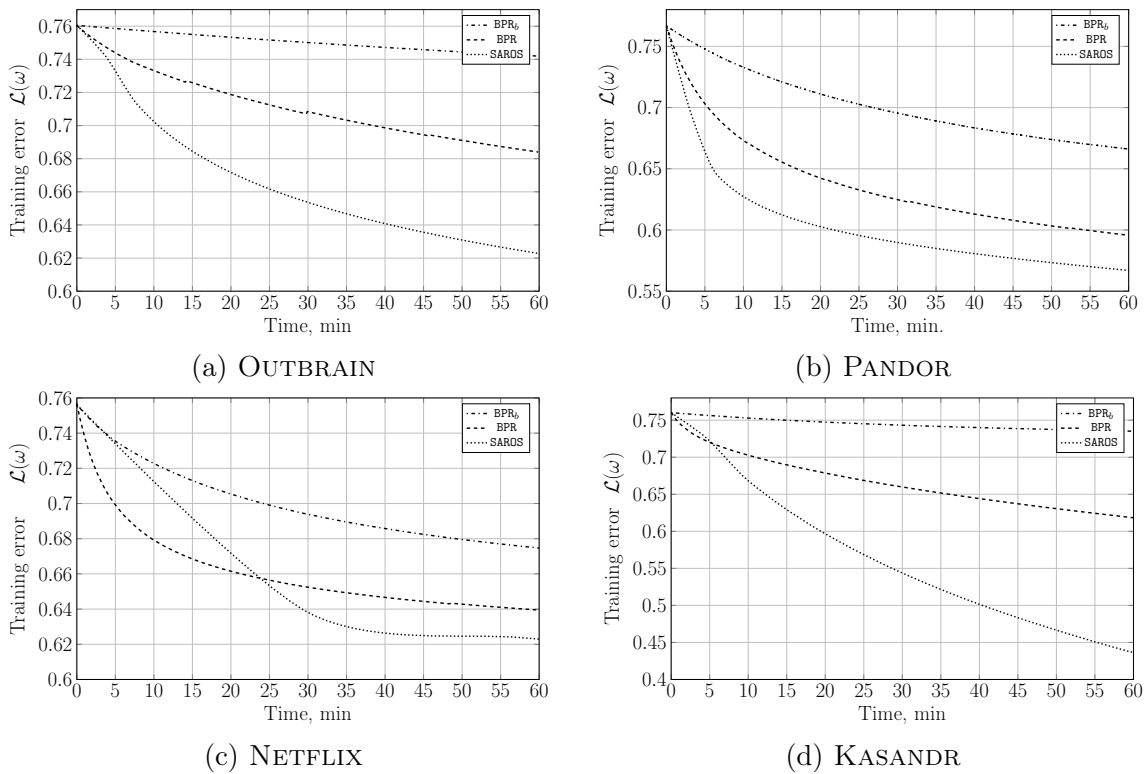
\begin{figure}[!th]
\vspace*{-10pt}
\small
    \centering
    \begin{tabular}{cc}
\begin{tikzpicture}[scale=0.45]
\begin{axis}[ 
 width=\columnwidth, 
 height=0.65\columnwidth, 
 xmajorgrids, 
 yminorticks=true, 
 ymajorgrids, 
 yminorgrids,
 ylabel={Training error ~$\mathcal{L}(\omega)$},
 xlabel = {Time, min},
 ymin = 0.60,
 ymax=0.77,
 xmin = 0,
 xmax = 60,
label style={font=\Large} ,
tick label style={font=\Large}
 ];

  \addplot  [color=black,
                dash pattern=on 1pt off 3pt on 3pt off 3pt,
                mark=none,
                mark options={solid},
                smooth,
                line width=1.2pt]  file {chapter2-out_batch.txt }; 
 \addlegendentry{ \batch };

  \addplot  [color=black,
                dashed,
                mark=none,
                mark options={solid},
                smooth,
                line width=1.2pt]  file {chapter2-out_sgd.txt }; 
 \addlegendentry{ \BPR };
  \addplot  [color=black,
                dotted,
                mark=none,
                mark options={solid},
                smooth,
                line width=1.2pt]  file {chapter2-out_online.txt}; 
 \addlegendentry{ \SO };
\end{axis}
\end{tikzpicture} & \begin{tikzpicture}[scale=0.45]
\begin{axis}[ 
 width=1.0\columnwidth, 
 height=0.65\columnwidth, 
 xmajorgrids, 
 yminorticks=true, 
 ymajorgrids, 
 yminorgrids,
 ylabel={Training error ~$\mathcal{L}(\omega)$},
 xlabel = {Time, min.},
 ymin = 0.55,
 ymax= 0.78,
 xmin = 0,
 xmax = 60,
label style={font=\Large} ,
tick label style={font=\Large}
 ]
 
 \addplot  [color=black,
                dash pattern=on 1pt off 3pt on 3pt off 3pt,
                mark=none,
                mark options={solid},
                smooth,
                line width=1.2pt]  file { chapter2-pandor_batch.txt }; 
 \addlegendentry{ \batch }; 
 
   \addplot  [color=black,
                dashed,
                mark=none,
                mark options={solid},
                smooth,
                line width=1.2pt]  file { chapter2-pandor_sgd.txt }; 
 \addlegendentry{ \BPR };

  \addplot  [color=black,
                dotted,
                mark=none,
                mark options={solid},
                smooth,
                line width=1.2pt]  file { chapter2-pandor_online.txt }; 
 \addlegendentry{ \SO };
\end{axis}
\end{tikzpicture} \\
(a) \Out & (b) \PANDOR\\
\begin{tikzpicture}[scale=0.45]
\begin{axis}[ 
 width=\columnwidth, 
 height=0.65\columnwidth, 
 xmajorgrids, 
 yminorticks=true, 
 ymajorgrids, 
 yminorgrids,
 ylabel={Training error ~~~~$\mathcal{L}(\omega)$},
 xlabel = {Time, min},
 ymin = 0.6,
 ymax=0.76,
 xmin = 0,
 xmax = 60,
label style={font=\Large} ,
tick label style={font=\Large}
 ];
 
  \addplot  [color=black,
                dash pattern=on 1pt off 3pt on 3pt off 3pt,
                mark=none,
                mark options={solid},
                smooth,
                line width=1.2pt]  file { chapter2-netflix_batch.txt }; 
 \addlegendentry{ \batch };

  \addplot  [color=black,
                dashed,
                mark=none,
                mark options={solid},
                smooth,
                line width=1.2pt]  file { chapter2-netflix_sgd.txt }; 
 \addlegendentry{ \BPR };
  \addplot  [color=black,
                dotted,
                mark=none,
                mark options={solid},
                smooth,
                line width=1.2pt]  file { chapter2-netflix_online.txt }; 
 \addlegendentry{ \SO };
 \end{axis}
\end{tikzpicture} & \begin{tikzpicture}[scale=0.45]
\begin{axis}[ 
 width=\columnwidth, 
 height=0.65\columnwidth, 
 xmajorgrids, 
 yminorticks=true, 
 ymajorgrids, 
 yminorgrids,
 ylabel={Training error ~$\mathcal{L}(\omega)$},
 xlabel = {Time, min},
 ymin = 0.4,
 ymax=0.78,
 xmin = 0,
 xmax = 60,
label style={font=\Large} ,
tick label style={font=\Large}
 ];

 \addplot  [color=black,
                dash pattern=on 1pt off 3pt on 3pt off 3pt,
                mark=none,
                mark options={solid},
                smooth,
                line width=1.2pt]  file { chapter2-kassandr_batch.txt }; 
 \addlegendentry{ \batch };

  \addplot  [color=black,
                dashed,
                mark=none,
                mark options={solid},
                smooth,
                line width=1.2pt]  file { chapter2-kassandr_sgd.txt }; 
 \addlegendentry{ \BPR };
 
  \addplot  [color=black,
                dotted,
                mark=none,
                mark options={solid},
                smooth,
                line width=1.2pt]  file { chapter2-kassandr_online.txt }; 
 \addlegendentry{ \SO };

\end{axis}
\end{tikzpicture}\\
(c) \NetF & (d) \kasandr\\
\end{tabular}
\caption[Loss {\batch}, {\BPR} and {\SO} as a function of time]{Evolution of the loss on training sets for both {\batch}, {\BPR} and {\SO} as a function of time in minutes for all collections.}
    \label{fig:losses}
\end{figure}
\end{flushleft}

\vspace{-8mm}We also compare the performance of all the approaches on the basis of the common ranking metrics, which are the Mean Average Precision at rank $K$ (\mapk) over all users  defined as $\mapk=\frac{1}{N}\sum_{u=1}^{N}\apk(u)$, where $\apk(u)$ is the average precision of  preferred items of user $u$ in the top $K$ ranked ones; and the Normalized Discounted Cumulative Gain at rank $K$ (\ndcgk) that computes the ratio of the obtained ranking to the ideal case and allow to consider not only binary relevance as in Mean Average Precision, $\ndcgk = \frac{1}{N}\sum_{u=1}^{N}\frac{\dcgk(u)}{\idcgk(u)}$, where $\dcgk(u) = \sum_{i=1}^{K}\frac{2^{rel_{i}}-1}{\log_{2}(1+i)}$, $rel_{i}$ is the graded relevance of the item at position $i$; and $\idcgk(u)$ is $\dcgk(u)$ with an ideal ordering equals to $\sum_{i=1}^{K}\frac{1}{\log_{2}(1+i)}$ for $rel_{i}\in[0,1]$  \cite{Manning:2008}.

Table \ref{tab:online_vs_all_ndcg_1h} presents {\mapfive} and {\mapten} (top), and {\ndcgfive} and {\ndcgten} (down)  of  all approaches over the test sets of the different collections.
\begin{table*}[h!]
    \centering
     \resizebox{1.0\textwidth}{!}{\begin{tabular}{c|ccccc|ccccc}
    \hline
     \multirow{2}{*}{}&\multicolumn{5}{c|}{\mapfive}&\multicolumn{5}{c}{\mapten}\\
     \cline{2-11}
     &{\ML-1M}&\Out&\PANDOR&{\NetF}&{\kasandr}&{\ML-1M}&\Out&\PANDOR&{\NetF}&{\kasandr}\\
     \hline
     \MostPop  & .074&.007 &.003 &.039 & .002& .083&.009 &.004 &.051 &.3e-5 \\
     \ProdVec  & .793& .228& .063& .669& .012& .772& .228&.063 &.690 &.012 \\
     \MF       & .733&.531 &.266 & .793& .170& .718& .522& .267& .778& .176\\
    \batch     & .713& .477& .685& .764& .473& .688& .477& .690& .748& .488\\
     \BPR      & \underline{.826}& \underline{.573}& \underline{.734}& \underline{.855}& .507&\underline{.797}& \underline{.563}& \underline{.760} & \underline{.835}& .521\\
     {\GRU}$+$ & .777& .513 & .673& .774& \underline{.719}& .750& .509&.677 &.757 &\underline{.720} \\
      {\caser} & .718& .471& .522& .749& .186& .694& .473& .527& .733& .197\\
     {\SASR}   & .776 & .542& .682& .819& .480& .751& .534& .687& .799& .495\\
     {\SO}     & \bf{.837}& \bf{.619}& \bf{.750}& \bf{.866}& \bf{.732}& \bf{.808}& \bf{.607}& \bf{.753}& \bf{.846}& \bf{.747}\\

     \hline
    \end{tabular}
    }
~\\
~\\
     \resizebox{1.0\textwidth}{!}{\begin{tabular}{c|ccccc|ccccc}
    \hline
     \multirow{2}{*}{}&\multicolumn{5}{c|}{\ndcgfive}&\multicolumn{5}{c}{\ndcgten}\\
     \cline{2-11}
     &{\ML-1M}&\Out&\PANDOR&{\NetF}&{\kasandr}&{\ML-1M}&\Out&\PANDOR&{\NetF}&{\kasandr}\\
     \hline
     \MostPop  & .090&.011 &.005 &.056 & .002& .130&.014 &.008 &.096 &.002 \\
     \ProdVec  & .758& .232& .078& .712& .012& .842& .232&.080 &.770 &.012 \\
     \MF       & .684&.612 &.300 & .795& .197& .805& .684& .303& .834& .219\\
    \batch     & .652& .583& .874& .770& .567& .784& .658& .890& .849& .616\\
     \BPR      & \underline{.776}& \underline{.671}& \underline{.889}& \underline{.854}& .603 &\underline{.863}& \underline{.724}& \underline{.905}& \underline{.903}& .650\\
     {\GRU}$+$ & .721& .633 & .843& .777& \underline{.760}& .833& .680&.862 &.854 &\underline{.782} \\
      {\caser} & .665& .585& .647& .750& .241& .787& .658& .666& .834& .276\\
     {\SASR}   & .721 & .645& .852& .819& .569& .832& .704& .873& .883& .625\\
     {\SO}     & \bf{.788}& \bf{.710}& \bf{.903}& \bf{.865}& \bf{.791}& \bf{.874}& \bf{.755}& \bf{.913}& \bf{.914}& \bf{.815}\\

     \hline
    \end{tabular}
    }
    \caption[Empirical comparison of \MostPop, \ProdVec, \MF, \batch, \BPR{}, {\GRU}$+$, \SASR, \caser \ and \SO{}]{Comparison between \MostPop, \ProdVec, \MF, \batch, \BPR{}, {\GRU}$+$, \SASR, \caser \ and \SO{} approaches in terms of \mapfive{} and \mapten (top), and \ndcgfive{} and \ndcgten (down). Best performance is in bold and the second best is underlined.}
    \label{tab:online_vs_all_ndcg_1h}
\end{table*}
The non-machine learning method, {\MostPop},  gives results of an order of magnitude lower than the learning based approaches. Moreover,  the factorization model {\MF} which predicts clicks by matrix completion is less effective when dealing with implicit feedback than ranking based models especially on large datasets where there are fewer interactions. We also found that embeddings found by ranking based models, in the way that the user preference over the pairs of items is preserved in the embedded space by the dot product, are more robust than the ones found by {\ProdVec}. When comparing {\GRU}$+$ with {\BPR} that also minimizes the same surrogate ranking loss, the former outperforms it in case of {\kasandr} with a huge imbalance between positive and negative interactions. This is mainly because  {\GRU}$+$ optimizes an approximation of the relative rank that favors interacted items to be in the top of the ranked list while the logistic ranking loss, which is mostly related to the Area under the ROC curve \cite{Usunier:1121}, pushes up clicked items for having good ranks in average. However, the minimization of the logistic ranking loss over blocks of very small size pushes the clicked item to be ranked higher than the no-clicked ones in several lists of small size and it has the effect of favoring the clicked item to be at the top of the whole merged lists of items. Moreover, it comes out that {\SO} is the most competitive approach, performing better than other approaches over all collections even such as last published {\caser} and {\SASR}.

\section{Conclusion}\label{sec:ch3:Conclusion}

In this chapter, we first proposed {\SO},  a novel learning framework for large-scale Recommender Systems that sequentially updates the weights of a ranking function user by user over blocks of items ordered by time where each block is a sequence of negative items followed by a last positive one. The main hypothesis of the approach is that the preferred and no-preferred items within a local sequence of user interactions express better the user preference than when considering the whole set of preferred and no-preferred items independently one from another. The approach updates the model parameters user per user over blocks of items constituted by a sequence of unclicked items followed by a clicked one. The parameter updates are discarded for users who interact very little or a lot with the system.  The second contribution is a theoretical analysis of the proposed approach which bounds the deviation of the ranking loss concerning the sequence of weights found by the algorithm and its minimum in the case where the loss is convex.  Empirical results conducted on five real-life implicit feedback datasets support our founding and show that the proposed approach is significantly faster than the common batch and online optimization strategies that consist in updating the parameters over the whole set of users at each epoch, or after sampling random pairs of preferred and no-preferred items. The approach is also shown to be highly competitive concerning state of the art approaches on \texttt{MAP} and \texttt{NDCG} measures.

\chapter{Learning over No-Preferred and Preferred Sequence of Items for Robust Recommendation}
\label{ch:3}
\section{Introduction}\label{Introduction3}

In this chapter, we consider  two variants of the $\SO_{}$ strategy presented in chapter  \ref{ch:2}. The first variant, referred to as $\SO_m$, updates the model parameters at each time a block of unclicked items followed by a clicked one is formed after a user's interaction. Parameters' updates are carried out by minimizing the average ranking loss of the current model that scores the clicked item below the unclicked ones using a momentum method \cite{Polyak63,nesterov1983method,nesterov2018lectures}. The second strategy, which we refer to as $\SO_b$, is the same approach described in chapter \ref{ch:2}, where model parameters are updated by minimizing a ranking loss over the same blocks of unclicked items followed by a clicked one using a gradient descent approach; with the difference that parameter updates are discarded for users who interact very little or a lot with the system. The results of this chapter were published at the Journal of Artificial Intelligence Research (JAIR) in 2021, \cite{DBLP:journals/jair/BurashnikovaMCL21} and the European Conference in Information Retrieval (ECIR) in 2022 \citepers{DBLP:journals/corr/abs-2112-02242}. 

Our main contributions here are, 
\begin{itemize}
    \item We propose a unified framework in which we study the convergence properties of both  versions of \SO{} in the general case of non-convex ranking losses. This is an extension of our earlier results~\cite{Burashnikova19}, where only the convergence of $\SO_b$ was studied in the case of convex ranking losses.
    \item Furthermore, we provide empirical evaluation over six large publicly available datasets showing that both versions of $\SO_{}$ are highly competitive compared to the state-of-the-art models in terms of quality metrics and, that are significantly faster than both the batch and the online versions of the algorithm.
    \item Finally, we show the impact of homogeneous user/items interactions for prediction, after removal of non-stationarities
\end{itemize}

\bigskip

The rest of this chapter is organized as follows. Section \ref{sec:Frame} introduces the general ranking learning problem that we address in this study. Then, in Section~\ref{sec:TA3}, we present both versions of the $\SO_{}$ algorithm, $\SO_b$ and $\SO_m$, and provide an analysis of their convergence. Section \ref{sec:Exps} presents experimental results that support our approach. In Section \ref{sec:ch5:mo} we introduce a strategy to filter the dataset with respect to homogeneity
of the behavior in the users when interacting with the system, based on the concept
of memory. Finally, in Section \ref{sec:Conclusion}, we discuss the outcomes of this study and give some pointers to further research.

\section{Framework and Problem Setting}
\label{sec:Frame}

 A key point in recommendation is that user preferences for items are largely determined by the context in which they are presented to the user. A user may prefer (or not) two items independently of one another, but he or she may have a totally different preference for these items within a given set of shown items. This effect of local preference is not taken into account by randomly sampling triplets formed by a user and corresponding clicked and unclicked items over the entire set of shown items to the user. Furthermore, triplets corresponding to different users are non uniformly distributed, as interactions vary from one user to another one, and for parameter updates; triplets corresponding to low interactions have a small chance to be chosen. In order to tackle these points; in chapter \ref{ch:2} $\SO_{}$ was suggested to update the parameters sequentially. 

\subsection{Two strategies of \SO{}}
\label{sec:Algox}





Note that this is different from session-based recommendations \cite{Wang2020} in which each session is also made up of a series of user-item interactions that take place over a period of time. However, session-based recommendations approaches capture both  user's short-term preference from recent sessions and the preference dynamics representing the change of preferences from one session to the next by using each session as the basic input unit, which is not the case in our study.

We propose two strategies for the minimization of (Eq. \ref{eq:ch4:CLoss}, ch. \ref{ch:2}) and the update of weights. In the first one, referred to as {\SO$_m$}, the aim is to carry out an effective minimization of the ranking loss \eqref{eq:ch4:CLoss} by lessening the oscillations of the updates through the minimum. This is done by defining the updates as the linear combination of the gradient of the loss of (Eq. \ref{eq:ch4:CLoss}), $\nabla  \widehat{\calL}_{{\mathcal B}^t_u}(w_u^{t})$, and the previous update as in the momentum technique at each iteration $t$~:

\begin{align}
\label{thm11:eq1}
    v_u^{t+1}&=\mu\cdot v_u^{t}+(1-\mu)\nabla  \widehat{\calL}_{{\mathcal B}^t_u}(w_u^{t})\\
    w_u^{t+1}&= w_u^{t}-\alpha v_u^{t+1}
\end{align}

\noindent 
where $\alpha$ and $\mu$ are hyperparameters of the linear combination. In order to explicitly take into account bot attacks -- in the form of excessive clicks over some target items -- we propose a second variant of this strategy, referred to as {$\SO_b$}. This variant consists in fixing two thresholds $b$ and $B$  over the parameter updates. For a new user $u$, model parameters are updated if and only if the number of  blocks of items constituted for this user is within the interval $[b,B]$.

The pseudo-code of {$\SO_b$} is shown in Algorithm \SO{} of chapter \ref{ch:2}. The sequential update rule, for each current user $u$ consists in updating the weights by making one step towards the opposite direction of the gradient of the ranking loss estimated on the current block,  $\calB_u^t=\text{N}_u^{t}\sqcup\Pi_u^{t}$~:

\begin{equation}
\weight_u^{t+1} = \weight_u^t - \frac{\eta}{|{\text{N}_u^{t}}||\Pi_u^{t}|}  \displaystyle{\sum_{i\in\Pi_u^{t}}\sum_{i'\in \text{N}_u^{t}}} \nabla \ell_{u,i,i'} (\weight_{u}^t)
\end{equation}

For a given user $u$, parameter updates are discarded if the number of blocks $(\calB_u^t)_t$ for the current user falls outside the interval $[b,B]$.

\subsection{Convergence Analysis}
\label{sec:TA3}

The proofs of algorithms' convergence are given under a common hypothesis that the sample distribution is not instantaneously affected by learning of the weights, i.e. the samples can be considered as i.i.d. More precisely, we assume the following hypothesis.

\begin{assumption}\label{ass:1}
    For an i.i.d. sequence of user and any $u, t \ge 1$, we have 
    \begin{enumerate}
        \item 
        $\mathbb{E}_{(u,{\cal B}_u^t)} \|\nabla {\calL}(\omega_u^t) - \nabla \hat{\calL}_{{\cal B}_u^t}(\omega_u^t)\|_2^2 \le \sigma^2$,
        \item  For any $u$,  $\left|\mathbb{E}_{ {{\cal B}_u^t}|u} \langle\nabla {\calL}(\omega_u^t), \nabla {\calL}(\omega_u^t) -  \nabla \hat{\calL}_{{\cal B}_u^t}(\omega_u^t) \rangle \right| \le a^2 \|\nabla {\calL}(\omega_u^t)\|_2^2$
    \end{enumerate}
    for some parameters $\sigma>0$ and $a\in [0,1/2)$ independent of $u$ and $t$. 
\end{assumption}

The first assumption is common in stochastic optimization and it implies consistency of the sample average approximation of the gradient. However, this assumption is not sufficient to prove the convergence because of interdependency of different blocks of items for the same user. 

The second assumption implies that in the neighborhood of the optimal point, we have $\nabla {\calL}(\omega_u^t)^\top \nabla \hat{\calL}_{{\cal B}_u^t}(\omega_u^t) \approx \|\nabla {\calL}(\omega_u^t)\|_2^2$, which greatly helps to establish consistency and convergence rates for both variants of the methods. In particular, if an empirical estimate of the loss over a block is unbiased, e.g. $\mathbb{E}_{\calB_u^t} \nabla {\hat \calL}_{\calB_u^t}(\omega) = \nabla \calL (\omega)$, the second assumption is satisfied with $a = 0$.

The following theorem establishes the convergence rate for the \SO{}$_b$ algorithm. 

\begin{theorem}\label{thrm:new-01}
Let $\ell$ be a (possibly non-convex) $\beta$-smooth loss function. Assume, moreover,  that the number of interactions per user belongs to an interval $[b, B]$ almost surely and assumption~\ref{ass:1} is satisfied with some constants $\sigma^2$ and $a$, $0 < a < 1/2$. 
Then, for a step-size policy $\eta_u^t \equiv \eta_u$ with $\eta_u\leq 1/(B\beta)$ for any user $u$, one has
\begin{align}
\min_{u:\, 1\le u \le N}\mathbb{E}    \|\nabla {\calL}(\omega_u^0)\|_2^2 \le  \frac{2({\calL}(\omega_1^0) - {\calL}(\omega_u^0)) + \beta \sigma^2 \sum_{u=1}^N \sum_{t=1}^{|{\cal B}_u|}(\eta_u^t)^2}{\sum_{u=1}^N \sum_{t=1}^{|{\cal B}_u|} \eta_u^t(1 - a^2 - \beta \eta_u^t(1/2 - a^2))}.
\end{align}
In particular, for a constant step-size policy $\eta_u^t = \eta = c/\sqrt{N}$ satisfies $\eta \beta \le 1$, one has 
\begin{align*}
    \min_{t, u} \|\nabla {\calL}(\omega_{u}^t)\|_2^2 \le 
    \frac{2}{b(1-4a^2)} \frac{2 ({\calL}(\omega_1^0) - {\calL}(\omega_*))/c + \beta c \sigma^2 B}{\sqrt{N}}. 
\end{align*}
\end{theorem}

\begin{Proof}
Since $\ell$ is a $\beta$ smooth function, we have for any $u$ and $t$:
\begin{align*}
    {\calL}(\omega_{u}^{t+1}) & \le {\calL}(\omega_{u}^t) + \langle\nabla {\calL}(\omega_{u}^t), \omega_u^{t+1} - \omega_u^{t} \rangle + \frac{\beta}{2}(\eta_u^t)^2 \|\nabla \hat{\calL}_{{\cal B}_u^t}(\omega_u^t)\|_2^2
    \\
    & = {\calL}(\omega_{u}^t) - \eta_u^t \langle \nabla {\calL}(\omega_{u}^t), \nabla \hat{\calL}_{{\cal B}_u^t}(\omega_u^t) \rangle +   \frac{\beta}{2}(\eta_u^t)^2 \|\nabla \hat{\calL}_{{\cal B}_u^t}(\omega_u^t)\|_2^2
\end{align*}    
Following~\cite{lan2020first}; by denoting $\delta_u^t = \nabla \hat{\calL}_{{\cal B}_u^t}(\omega_u^t) - \nabla {\calL}(\omega_u^t)$, we have: 
 \begin{align}\label{eq:01}
    {\calL}(\omega_{u}^{t+1})   & \le {\calL}(\omega_{u}^t) - \eta_u^t \langle \nabla {\calL}(\omega_{u}^t), \nabla {\calL}(\omega_{u}^t) + \delta_u^t \rangle +   \frac{\beta}{2}(\eta_u^t)^2 \|\nabla {\calL}(\omega_u^t) + \delta_u^t\|_2^2\nonumber\\
    & = 
    {\calL}(\omega_{u}^t) 
    +  \frac{\beta(\eta_u^t)^2}{2}\|\delta_u^t\|_2^2 
    - \!\!\left(\eta_u^i - \frac{\beta(\eta_u^t)^2}{2}\right)\!\!\|\nabla {\calL}(\omega_{u}^t)\|_2^2 \nonumber\\ 
    & \hspace{60mm} - \left(\eta_u^t - \beta (\eta_u^t)^2\right) \langle\nabla {\calL}(\omega_u^t), \delta_u^t\rangle 
\end{align}

\noindent Our next step is to take the expectation on  both sides of inequality~\eqref{eq:01}. According to  Assumption~\ref{ass:1}, one has for some $a\in [0, 1/2)$:
\begin{align*}
    \left(\eta_u^t - \beta (\eta_u^t)^2\right)\left|\mathbb{E} \langle\nabla {\calL}(\omega_u^t), \delta_u^t\rangle\right| \le \left(\eta_u^t - \beta (\eta_u^t)^2\right) a^2 \|\nabla \calL (\omega_u^t)\|_2^2, 
\end{align*}
where the expectation is taken over the set of blocks and users seen so far. 

Finally, taking the same expectation on both sides of inequality~\eqref{eq:01}, it comes:
\begin{align}\label{eq:02}
    {\calL}(\omega_{u}^{t+1}) &\le  {\calL}(\omega_{u}^{t}) + \frac{\beta}{2}(\eta_u^t)^2\mathbb{E}\|\delta_u^t\|_2^2 - 
    \eta_u^t(1 - \beta\eta_u^t/2  - a^2|1 - \beta\eta_u^t|) \|\nabla {\calL}(\omega_{u}^t)\|_2^2 \nonumber\\
    & \le  {\calL}(\omega_{u}^{t}) + \frac{\beta}{2}(\eta_u^t)^2\|\delta_u^t\|_2^2 - 
    \eta_u^t\underbrace{(1 - a^2 
    -    \beta \eta_u^t(1/2 - a^2))}_{:= z_u^t} \|\nabla {\calL}(\omega_{u}^t)\|_2^2 \nonumber\\
    & = {\calL}(\omega_{u}^{t}) + \frac{\beta}{2}(\eta_u^t)^2\|\delta_u^t\|_2^2 - 
    \eta_u^t z_u^t \|\nabla {\calL}(\omega_{u}^t)\|_2^2 \nonumber\\
    & = {\calL}(\omega_{u}^{t}) + \frac{\beta}{2}(\eta_u^t)^2\sigma^2 - 
    \eta_u^t z_u^t \|\nabla {\calL}(\omega_{u}^t)\|_2^2, 
\end{align}
where the second inequality is due to  $|\eta_u^t\beta|\leq 1$. Also, as $|\eta_u^t\beta|\leq 1$ and $a^2\in [0,1/2)$ one has $z_u^t>0$ for any $u,t$. Rearranging the terms, one has
\begin{align*}
    \sum_{u=1}^N\sum_{t=1}^{|{\cal B}_u|} \eta_u^t z_u^t \|\nabla {\calL}(\omega_{u}^t)\|_2^2 \le {\calL}(\omega_1^0) - {\calL}(\omega_*) + \frac{\beta \sigma^2}{2}  \sum_{u=1}^N\sum_{t=1}^{|{\cal B}_u|}(\eta_u^t)^2.
\end{align*}
and
\begin{align*}
    \min_{t, u} \|\nabla {\calL}(\omega_{u}^t)\|_2^2
    & 
    \le 
    \frac{{\calL}(\omega_1^0) - {\calL}(\omega_*) + \frac{\beta}{2}  \sum_{u=1}^N\sum_{t=1}^{|{\cal B}_u|}(\eta_u^t)^2 \sigma^2 }{\sum_{u=1}^N\sum_{t=1}^{|{\cal B}_u|} \eta_u^t z_u^t} \\
    & 
    \le \frac{{\calL}(\omega_1^0) - {\calL}(\omega_*) + \frac{\beta}{2}  \sum_{u=1}^N\sum_{t=1}^{|{\cal B}_u|}(\eta_u^t)^2 \sigma^2 }{\sum_{u=1}^N\sum_{t=1}^{|{\cal B}_u|} \eta_u^t (1 - a^2 - \beta \eta_u^t(1/2 - a^2))} 
\end{align*}
Where, $\omega_*$ is the optimal point. Then, using a constant step-size policy, $\eta_u^i = \eta$, and the bounds on a block size, $b\leq |{\cal B}_u|\leq B$, we get:
\begin{align*}
    \min_{t, u} \|\nabla {\calL}(\omega_{u}^t)\|_2^2 & \le \frac{{\calL}(\omega_1^0) - {\calL}(\omega_*) + \frac{\beta\sigma^2}{2}  N\sum_{u=1}^N\eta_u^2  }{b\sum_{u=1}^N\eta_u (1 - a^2 - \beta \eta_u (1/2 - a^2))} \\
    &
    \le  \frac{4{\calL}(\omega_1^0) - 4{\calL}(\omega_*) + 2\beta \sigma^2 B \sum_{u=1}^N\eta^2}{b(1 - 4a^2)\sum_{u=1}^N\eta} \\
    &
    \le 
    \frac{2}{b(1-4a^2)} \left\{\frac{2{\calL}(\omega_1^0) - 2{\calL}(\omega_*)}{N \eta} + \beta\sigma^2 B \eta\right\}. 
\end{align*}

Taking $\eta = c/\sqrt{N}$ so that $0 < \eta \le 1/\beta$, one has 
\begin{align*}
    \min_{t, u} \|\nabla {\calL}(\omega_{u}^t)\|_2^2 \le 
    \frac{2}{b(1-4a^2)} \frac{2 ({\calL}(\omega_1^0) - {\calL}(\omega_*))/c + \beta c \sigma^2 B}{\sqrt{N}}. 
\end{align*}
If $b = B = 1$, this rate matches up to a constant factor to the standard $O(1/\sqrt{N})$ rate of the stochastic gradient descent. 
\end{Proof}

Note that the stochastic gradient descent strategy implemented in the Bayesian Personalized Ranking model (\BPR) \cite{rendle_09} also converges to the minimizer of the ranking loss ${\calL}(\omega)$ (Eq. \ref{eq:RL}) with the same rate.

The analysis of momentum algorithm \SO{}$_m$ is slightly more involved. We say that a function $f(x)$ satisfies the Polyak-\L{}ojsievich condition \cite{Polyak63,Lojas63,karimi2016linear} if the following inequality holds for some $\mu > 0$: 
\[
    \frac{1}{2} \|\nabla f(x)\|_2^2 \ge \mu (f(x) - f^*),
\]
where $f^*$ is a global minimum of $f(x)$.

From this definition, we can derive an analysis on the convergence of \SO{}$_m$ stated below.

\begin{theorem}\label{thrm:new-02}
Let ${\calL}(\omega)$ be a (possibly non-convex) $\beta$-smooth function which satisfies the Polyak-Lojasievich condition with a constant $\mu > 0$. 
Moreover, assume the number of interactions per user belongs to an interval $[b, B]$ almost surely for some positive $b$ and $B$, and Assumption \ref{ass:1} is satisfied with some $\sigma^2$ and $a$.
Then, for $N = \sum_{u = 1}^N |{\cal B}_u|$ 
and a constant step-size policy $\eta_u^t = \eta $ with $\eta \beta \le 1$, one has
\begin{align*}
    {\calL}(\omega_u^{t+1}) - {\calL}(\omega_*) 
    \le \exp(- \mu \eta N) ({\calL}(\omega_1^{0}) - {\calL}(\omega_*)) + \frac{\beta\sigma^2 \eta^2}{2(1 - \mu/\beta)}, \quad \eta\beta \le 1.
\end{align*}
where the estimation is uniform for all $a$, $0 \le a < 1/2$. 
 
In particular, if $\eta = c/\sqrt{N}$, under the same conditions one has
\begin{align*}
     {\calL}(\omega_u^t) - {\calL}(\omega_*) \le \exp(-\mu c \sqrt{N}) ({\calL}(\omega_1^0) - {\calL}(\omega_*)) + \frac{\beta\sigma^2c^2}{2(1 - \mu/\beta)N}. 
 \end{align*}
\end{theorem}

\begin{Proof}
Similarly to the Theorem~\ref{thrm:new-01}, From Ineq.~\eqref{eq:02} we have: 
\begin{align*}
    {\calL}(\omega_u^{t+1}) \le {\calL}(\omega_u^{t}) + \frac{\beta}{2} (\eta_u^t)^2 \sigma^2 - \eta_u^t z_u^t \|\nabla {\calL}(\omega_u^{t})\|_2^2
\end{align*}
for $z_u^t = 1 - a^2 - \beta \eta_u^t(1/2 - a^2)>0$. Further, using the Polyak-Lojasievich condition, it comes: 
\begin{align*}
    - \eta_u^t z_u^t \|\nabla {\calL}(\omega_u^{t})\|_2^2 \le - 2\mu \eta_u^t z_u^t (\calL(\omega_u^t) - \calL(\omega_*)), 
\end{align*}
and
\begin{align*}
    {\calL}(\omega_u^{t+1}) - {\calL}(\omega_*) & \le {\calL}(\omega_u^{t}) - {\calL}(\omega_*) + \frac{\beta}{2} (\eta_u^t)^2 \sigma^2 - 2\mu \eta_u^t z_u^t({\calL}(\omega_u^{t}) - {\calL}(\omega_*))\\
    & \le ({\calL}(\omega_u^{t}) - {\calL}(\omega_*)) (1 - 2\mu\eta_u^t z_u^t) + \frac{\beta}{2} (\eta_u^t)^2 \sigma^2\\
    & \le \prod_{u}\prod_{t} (1 - 2\mu\eta_u^t z_u^t) ({\calL}(\omega_1^{0}) - {\calL}(\omega_*)) + \frac{\beta\sigma^2}{2}\sum_{v\le u} (\eta_v^t)^2\prod_{v}\prod_{t} (1 - 2\mu\eta_v^t z_v^t)
\end{align*}

Finally, for a constant step-size policy, $\eta_u^t = \eta$, one has $z_u^t = z = 1 - a^2 - \beta\eta(1/2-a^2)$ and
\begin{align*}
    {\calL}(\omega_u^{t+1}) - {\calL}(\omega_*) \le (1 - 2\mu \eta z)^N ({\calL}(\omega_1^{0}) - {\calL}(\omega_*)) + \frac{\beta\sigma^2 \eta^2}{2(1 - 2\mu \eta z)},
\end{align*}
where the last term is given by summing the geometric progression. As $\beta\eta \le 1$ and $a< 1/2$ one has $z \ge 1/2$. Thus
\begin{align*}
    {\calL}(\omega_u^{t+1}) - {\calL}(\omega_*) & \le (1 - \mu \eta )^N ({\calL}(\omega_1^{0}) - {\calL}(\omega_*)) + \frac{\beta\sigma^2 \eta^2}{2(1 - \mu/\beta)}\\
    & \le \exp(- \mu \eta N) ({\calL}(\omega_1^{0}) - {\calL}(\omega_*)) + \frac{\beta\sigma^2 \eta^2}{2(1 - \mu/\beta)}, \quad \eta\beta \le 1.
\end{align*}
Taking $\eta = c/\sqrt{N}$ for some positive $c$ guarantees a rate of convergence $O(1/N)$. With a different choice of the step-size policy, rates almost up to $O(1/N^2)$ are possible; however, these rates imply $O(1/N)$ convergence for the norm of the gradient which matches the standard efforts of stochastic gradient descent under the Polyak-Lojasievich condition \cite{karimi2016linear}. 
\end{Proof}

\section{Recommender systems: when memory matters}
\label{sec:ch5:mo}
In this section, we  put in evidence $(a)$  the impact of homogeneous user/items interactions for prediction, after removal of non-stationarities and $(b)$ the need of designing specific strategies to remove non-stationarities due to a specificity of RS, namely the presence of memory in user/items interactions. Thereafter, we turn this preliminary study into a novel and successful strategy combining sequential learning per blocks of interactions and removing user with non--homogeneous behavior from the training.

In the following, we present the mathematical framework, used to model stationarity in RS data. Thereafter, we explain that in the case, where we have presence of long-memory in the data removing non-stationarites is specially tricky. We present our novel strategy combining the efficiency of sequential learning per block of interactions and the knowledge of the memory behavior of each user  to remove non-stationarities. We then illustrate that memory is intrinsically present in RS user/items interactions and that we have to take it into account to remove non-stationarities and improve generalization. We then prove through  experiments on different large-scale benchmarks the effectiveness of our approach.

\subsection{Stationarity}\label{s:memory}
Our claim is that all user/items interactions may not be equally relevant in the learning process. We prove in the sequel that we can improve the learning process, considering only the subset of users whose interactions with the system are {\it homogeneous in time}, meaning that the user feedback is statistically the same, whatever the time period is. Unfortunately, non-stationarities are not easy to detect, since we have to take into account another additional effect in RS, which is {\it long-range dependence}. Indeed, in RS the choice of a given user may be influenced not only by its near past but by the whole history of interactions.

\noindent We propose to model these two natural characteristics of user feedbacks, using two mathematical notions introduced for sequential data analysis : {\it stationarity} and {\it memory}. We recall that a time series $X=\{X_t,t\in\mathbb{Z}\}$, here the sequence of user's feedback, is said to be (wide-sense) stationary (see Section 2.4 in~\cite{brillinger2001time}) if its two first orders moments are homogeneous with time: 
\begin{equation}
\forall t,k,l\in\mathbb{Z},\,\mathbb{E}[X_t]=\mu, \mbox{ and }Cov(X_k,X_l)=Cov(X_{k+t},X_{l+t})
\end{equation} 
Under such assumptions the autocovariance of a stationary process only depends on the difference between the terms of the series $h=k-l$. We set $\gamma(h)=Cov(X_0,X_h)$.\\

Our other concept of interest, memory arouses in time series analysis to model memory that can be inherently present in sequential data.  It provides a quantitative measure of the persistence of information related to the history of the time series in the long-run and it can be related to presence of non-stationarities in the data. 
Its definition is classically done in the Fourier domain and is based on the so-called spectral density. The spectral density is the discrete Fourier transform of the autocovariance function : \begin{equation}
    f(\lambda)=\frac{1}{2\pi}\sum\limits_{h=-\infty}^{+\infty}\gamma(h)e^{-ih\lambda}, \qquad \lambda\in(-\pi,\pi].
\end{equation}
and reflects the energy contains at each frequency $\lambda$ if the times series. A time series $X$ admits memory parameter $d\in \mathbb{R}$ iff its spectral density satisfies : \begin{equation}
    f(\lambda)\sim \lambda^{-2d}\mbox{ as } \lambda\to 0\;.
\end{equation}

\subsection{Memory}
In the time domain, the memory parameter is related to the decay of the autocovariance function. The more it is large, the more the past of the time series has an impact on its next future. Interestingly, when the memory parameter is large, the time series tends to have a sample autocorrelation function with large spikes at several lags which is well known to be the signature of non-stationarity for many practitioners. It can then be used as a measure of non-stationarity.

In order to infer this memory parameter, we use one of the most classical estimators of the memory parameter, the GPH estimator introduced in~\cite{GPH}. It consists of a least square regression of the log-periodogram of $X$. One first defines a biased estimator of the spectral density function, the periodogram $I(\lambda)$ and evaluate it on the Fourier frequencies $\lambda_k=\frac{2\pi k}{N}$ where $N$ is the length of the sample : \begin{equation}
    I_N(\lambda_k)=\frac{1}{N}\left|\sum\limits_{t=1}^N X_t e^{it\lambda_k}\right|^2
\end{equation}

The estimator of the memory parameter is therefore as follows : \begin{equation}
    \hat{d}(m)=\frac{\sum_{k=1}^m(Y_k-\Bar{Y})\log(I(\lambda_k))}{\sum_{k=1}^m(Y_k-\Bar{Y})^2},
\end{equation}

 where $Y_k=-2\log|1-e^{i\lambda_k}|$, $\Bar{Y}=(\sum_{k=1}^m Y_k)/m$ and $m$ is the number of used frequencies. 
 
We then classify the time series as non-stationary if $d\geq 1/2$, and as stationary otherwise.

The inclusion of the Memory-Aware step of our algorithm, allowing to include stationarity in the pipeline (called \MOSAIC), can be carried out in two steps. In the first step we train \SO{} on the full dataset. Thereafter we remove non-stationary embeddings, using a preliminary estimation of the memory parameter of each time series. Finally we train once more this filtered dataset and return the last updated weights.
\section{Experimental Setup and Results}
\label{sec:Exps}

The group of the experiments was expanded to a new dataset \RecS compared to the one of the chapter \ref{ch:2}. The dataset represents a sample based on historic XING data provided 6,330,562 feedback given by 39,518 users on the job posting items and the items generated by XING's job recommender system.

\noindent Updated statistics for \RecS{} are represented in table \ref{tab:datasets_new}.
\begin{table}[h!]
    \centering
    \begin{tabular}{lccccc}
    \hline
    Data&$|\mathcal{U}|$&
    $|\mathcal{I}|$ & Sparsity&~~~~Avg. \# of $+$~~~~ & ~Avg. \# of $-$~~~~\\
    \hline
    {\ML}-1M&6,040&3,706&.9553&95.2767& 70.4690\\
   \Out&49,615&105,176&.9997&6.1587& 26.0377 \\
   \PANDOR&177,366&9,077&.9987&1.3266& 10.3632\\
     \NetF&90,137&3,560&.9914&26.1872& 20.2765\\
     \kasandr&2,158,859&291,485&.9999&2.4202& 51.9384\\
    \textcolor{blue}{\RecS}&\textcolor{blue}{39,518}&\textcolor{blue}{28,068}&\textcolor{blue}{.9943}&\textcolor{blue}{26.2876}&\textcolor{blue}{133.9068}\\
        \hline
    \end{tabular}
    \caption[Statistics on the users-items interaction]{Statistics on the number of users and items; as well as the sparsity and the average number of $+$ (preferred) and $-$ (non-preferred) items on {\ML}-1M, {\NetF}, {\Out}, {\kasandr}, {\PANDOR} and {\RecS} collections after preprocessing.}
    \label{tab:datasets_new}
\end{table}

\RecS{} also is included into the group of the most unbalanced datasets, such as \kasandr{}, \PANDOR{} and \Out{}. Table with the sizes of the train/test parts and percentages of positive/negative feedback in the benchmarks also was updated and displayed in the table \ref{tab:detail_setting_new}.

\begin{table}[h!]
    \centering
    \begin{tabular}{lcccc}
    \hline
    Dataset&$|S_{train}|$~~~~&~~~$|S_{test}|$~~&~~$pos_{train}$~~&~~$pos_{test}$\\
    \hline
    {\ML}-1M&797,758&202,451&58.82&52.39\\
     \Out&1,261,373&336,053&17.64&24.73\\
    \PANDOR&1,579,716&493,663&11.04&12.33\\
    \NetF&3,314,621&873,477&56.27&56.70\\
    \kasandr&12,509,509&3,335,208&3.36&8.56\\
    \textcolor{blue}{\RecS}&\textcolor{blue}{5,048,653}&\textcolor{blue}{1,281,909}&\textcolor{blue}{17.07}&\textcolor{blue}{13.81}\\
    \hline
    \end{tabular}
    \caption[Positive feedback statistics]{Number of interactions used for train and test on each dataset, and the percentage of positive feedback among these interactions.}
    \label{tab:detail_setting_new}
\end{table}

\paragraph{Compared Approaches. } To estimate both strategies of sequential learning approach $\SO_m$ and $\SO_b$, we compared them with the same state-of-the-art approaches suggested in chapter \ref{ch:2}, section \ref{sec:ch3:Exps}. The set of baselines was extended by the modern graph-convolution-based model \GCN{} proposed by \cite{He2020}. This graph convolution network learns user and item embedding by linearly propagating them on the user-item interaction graph. The final representations are the weighted sum of the embeddings learned at all layers.

Hyper-parameters of \GCN{} and the dimension of the embedded space for the representation of users and items; as well as the regularisation parameter over the norms of the embeddings for all approaches were found using grid search on the validation set as before for the remains benchmarks.  
Supplemented information with the adjusted number of blocks and learnig rate involving the new \RecS{} is summarized below in the table \ref{parameters3}.

\begin{table}[!ht]
    \centering
    \begin{tabular}{c|cccccc}
    \hline
    Parameter&{\ML}&{\Out}&{\PANDOR}&{\NetF}&{\kasandr}&\textcolor{blue}{\RecS}\\
    \hline
    $B$&78&5&2&22&5&\textcolor{blue}{22}\\
    $b$&1&2&1&1&1&\textcolor{blue}{1}\\
    Learning rate&.05&.05&.05&.05&.4&\textcolor{blue}{.3}\\
    \hline
    \end{tabular}
    \caption{Hyperparameter values of $\SO_b$.}
    \label{parameters3}
\end{table}

\paragraph{Evaluation Setting and Results. }
All the experimental steps suggested for the practical part of chapter \ref{ch:2} were repeated on the additional new benchmark \RecS, baseline \GCN{} and $\SO_{m}$. For the first results, the testing {\batch}, \BPR{} and \SO{} approaches over the logistic ranking loss after training the models till the convergence are shown in Table \ref{test_loss_new}. 

\begin{table}[!ht]
\centering
{
\begin{tabular}{cc|cccccc}
\hline
Dataset &~~&\multicolumn{6}{c}{Test loss at convergence, Eq.~\eqref{eq:RL}}\\
\cline{3-8}
&~~&\batch&~~&{\BPR}&~~&$\SO_b$&$\SO_m$\\
\hline
\ML-1M &~~&0.744&~~&0.645&~~&\bf{0.608}&0.637\\
\Out &~~& 0.747&~~&0.638&~~&0.635&\bf{0.634}\\
\PANDOR &~~&0.694&~~&0.661&~~&\bf{0.651}&0.666\\
\NetF &~~&0.694&~~&0.651&~~&\bf{0.614}&0.618\\
\kasandr &~~&0.631&~~&0.393&~~&\bf{0.212}&0.257\\
\RecS &~~&0.761&~~&0.644&~~&0.640&\bf{0.616}\\
\hline
\end{tabular}
}
\caption[Comparison between \BPR, \batch{} and \SO{}]{Comparison between \BPR, \batch{} and \SO{} approaches in terms of test loss at convergence.}
\label{test_loss_new}
\end{table}


Figure \ref{fig:losses_new} shows the evolution of the training error over time for {\batch}, \BPR{}, $\SO_m{}$ and $\SO_b{}$ on \kasandr, \PANDOR, \Out{} and \NetF.
As we can see, the training error decreases in all cases and the three approaches converge to the same minimizer of the ranking loss (Eq.~\ref{eq:RL}). This is an empirical evidence of the convergence of $\SO_b$ and $\SO_m$, showing that the sequence of weights found by the proposed algorithm allows to minimize the general ranking loss (Eq.~\ref{eq:RL}) as it is stated in Theorems 1 and 2.

\begin{figure}[!ht]
\small
    \centering
    \begin{tabular}{cc}
\begin{tikzpicture}[scale=0.42]
\begin{axis}[ 
 width=\columnwidth, 
 height=0.65\columnwidth, 
 xmajorgrids, 
 yminorticks=true, 
 ymajorgrids, 
 yminorgrids,
 ylabel={Training error ~$\mathcal{L}(\omega)$},
 xlabel = {Time, min},
 ymin = 0.58,
 ymax=0.77,
 xmin = 0,
 xmax = 60,
label style={font=\Large} ,
tick label style={font=\Large}
 ];

  \addplot  [color=black,
                dash pattern=on 1pt off 3pt on 3pt off 3pt,
                mark=none,
                mark options={solid},
                smooth,
                line width=1.2pt]  file {chapter3-out_batch.txt }; 
 \addlegendentry{ \batch };

  \addplot  [color=black,
                dashed,
                mark=none,
                mark options={solid},
                smooth,
                line width=1.2pt]  file {chapter3-out_sgd.txt }; 
 \addlegendentry{ \BPR };
  \addplot  [color=black,
                dotted,
                mark=none,
                mark options={solid},
                smooth,
                line width=1.2pt]  file {chapter3-out_online.txt }; 
 \addlegendentry{ $\SO{}_b$ };
 
 \addplot  [color=black,
                dotted,
                mark=*,
                mark options={scale = 1.5, fill = red},
                smooth,
                line width=1.2pt]  file {chapter3-out_momentum.txt }; 
 \addlegendentry{ $\SO{}_m$ };
\end{axis}
\end{tikzpicture} & \begin{tikzpicture}[scale=0.42]
\begin{axis}[ 
 width=1.0\columnwidth, 
 height=0.65\columnwidth, 
 xmajorgrids, 
 yminorticks=true, 
 ymajorgrids, 
 yminorgrids,
 ylabel={Training error ~$\mathcal{L}(\omega)$},
 xlabel = {Time, min.},
 ymin = 0.55,
 ymax= 0.78,
 xmin = 0,
 xmax = 60,
label style={font=\Large} ,
tick label style={font=\Large}
 ]
 
 \addplot  [color=black,
                dash pattern=on 1pt off 3pt on 3pt off 3pt,
                mark=none,
                mark options={solid},
                smooth,
                line width=1.2pt]  file {chapter3-pandor_batch.txt }; 
 \addlegendentry{ \batch }; 
 
   \addplot  [color=black,
                dashed,
                mark=none,
                mark options={solid},
                smooth,
                line width=1.2pt]  file {chapter3-pandor_sgd.txt }; 
 \addlegendentry{ \BPR };

  \addplot  [color=black,
                dotted,
                mark=none,
                mark options={solid},
                smooth,
                line width=1.2pt]  file {chapter3-pandor_online.txt }; 
 \addlegendentry{ $\SO{}_b$ };
 
   \addplot  [color=black,
                dotted,
                mark=*,
                mark options={scale = 1.5, fill = red},
                smooth,
                line width=1.2pt]  file {chapter3-pandor_momentum.txt }; 
 \addlegendentry{ $\SO{}_m$ };
\end{axis}
\end{tikzpicture} \\
(a) \Out & (b) \PANDOR\\
\begin{tikzpicture}[scale=0.4]
\begin{axis}[ 
 width=\columnwidth, 
 height=0.65\columnwidth, 
 xmajorgrids, 
 yminorticks=true, 
 ymajorgrids, 
 yminorgrids,
 ylabel={Training error ~~~~$\mathcal{L}(\omega)$},
 xlabel = {Time, min},
 ymin = 0.6,
 ymax=0.76,
 xmin = 0,
 xmax = 60,
label style={font=\Large} ,
tick label style={font=\Large}
 ];
 
  \addplot  [color=black,
                dash pattern=on 1pt off 3pt on 3pt off 3pt,
                mark=none,
                mark options={solid},
                smooth,
                line width=1.2pt]  file {chapter3-netflix_batch.txt }; 
 \addlegendentry{ \batch };

  \addplot  [color=black,
                dashed,
                mark=none,
                mark options={solid},
                smooth,
                line width=1.2pt]  file {chapter3-netflix_sgd.txt }; 
 \addlegendentry{ \BPR };
  \addplot  [color=black,
                dotted,
                mark=none,
                mark options={solid},
                smooth,
                line width=1.2pt]  file {chapter3-netflix_online.txt }; 
 \addlegendentry{ $\SO{}_b$ };
 \addplot  [color=black,
                dotted,
                mark=*,
                mark options={scale = 1.5, fill = red},
                smooth,
                line width=1.2pt]  file {chapter3-netflix_momentum.txt }; 
 \addlegendentry{ $\SO{}_m$ };
 \end{axis}
\end{tikzpicture} & \begin{tikzpicture}[scale=0.42]
\begin{axis}[ 
 width=\columnwidth, 
 height=0.65\columnwidth, 
 xmajorgrids, 
 yminorticks=true, 
 ymajorgrids, 
 yminorgrids,
 ylabel={Training error ~$\mathcal{L}(\omega)$},
 xlabel = {Time, min},
 ymin = 0.4,
 ymax=0.78,
 xmin = 0,
 xmax = 60,
label style={font=\Large} ,
tick label style={font=\Large}
 ];

 \addplot  [color=black,
                dash pattern=on 1pt off 3pt on 3pt off 3pt,
                mark=none,
                mark options={solid},
                smooth,
                line width=1.2pt]  file {chapter3-kassandr_batch.txt }; 
 \addlegendentry{ \batch };

  \addplot  [color=black,
                dashed,
                mark=none,
                mark options={solid},
                smooth,
                line width=1.2pt]  file {chapter3-kassandr_sgd.txt }; 
 \addlegendentry{ \BPR };
 
  \addplot  [color=black,
                dotted,
                mark=none,
                mark options={solid},
                smooth,
                line width=1.2pt]  file {chapter3-kassandr_online.txt }; 
 \addlegendentry{$\SO{}_b$};
    \addplot  [color=black,
                dotted,
                mark=*,
                mark options={scale = 1.5, fill = red},
                smooth,
                line width=1.2pt]  file {chapter3-kassandr_momentum.txt }; 
 \addlegendentry{ $\SO{}_m$ };

\end{axis}
\end{tikzpicture}\\
(c) \NetF & (d) \kasandr\\
\end{tabular}
\caption[Training loss statistics for {\batch}, {\BPR} and {\SO}]{Evolution of the loss on training sets for both {\batch}, {\BPR} and {\SO} as a function of time in minutes for all collections.}
    \label{fig:losses_new}
\end{figure}
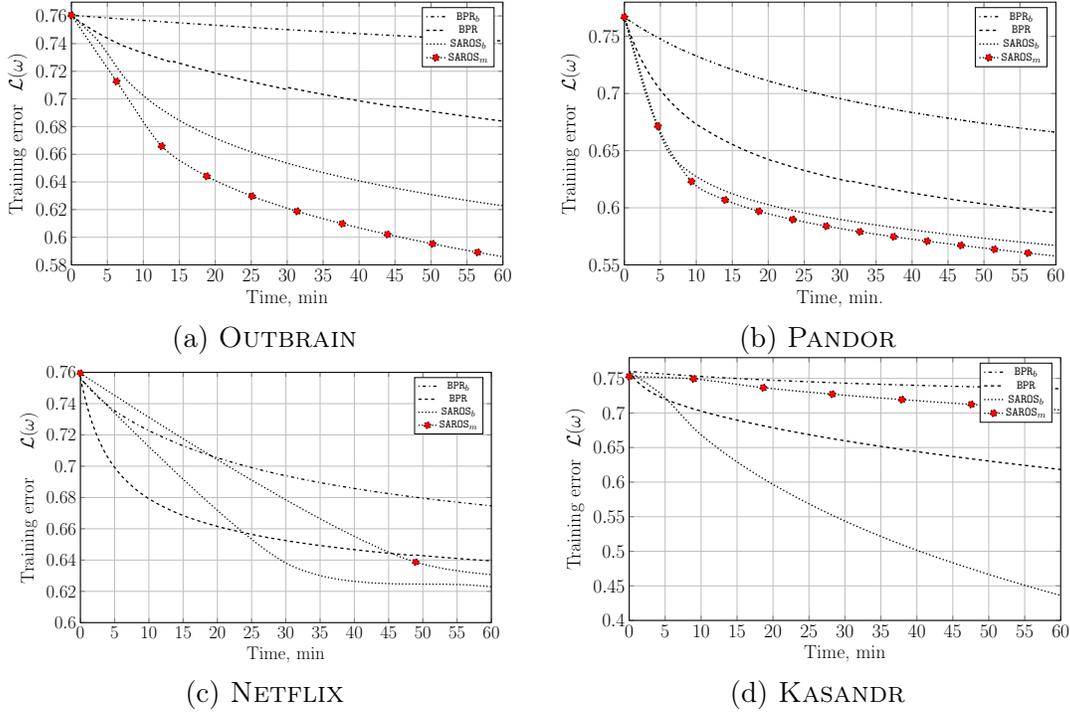


To estimate the importance of the maximum number of blocks ($B$) for $\SO_b$, we explore the dependency between quality metrics {\mapk} and {\ndcgk} on {\ML-1M} and {\PANDOR} collections (Figure \ref{fig:map5_blocks}). The latter records the clicks generated by users on one of Purch’s high-tech website and it was subject to bot attacks \cite{sidana18}. For this collection, large values of $B$ affects  {\mapk} while the measure reaches a plateau on {\ML-1M}. The choice of this hyperparameter may then have an impact on the results. As future work, we are investigating the modelling of bot attacks by studying the effect of long memory in the blocks of no-preferred and preferred items in small and large sessions with the aim of automatically fixing this threshold $B$.

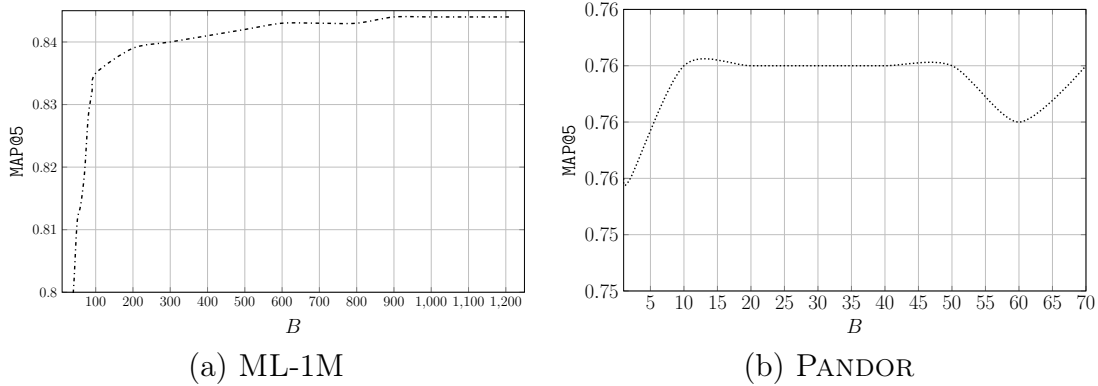
\begin{figure}[!ht]
    \centering
    \begin{tabular}{cc}

\begin{tikzpicture}[scale=0.45]
\begin{axis}[ 
 width=1.0\columnwidth, 
 height=0.65\columnwidth, 
 xmajorgrids, 
 yminorticks=true, 
 ymajorgrids, 
 yminorgrids,
 ylabel={\mapfive},
 xlabel = {$B$},
 ymin = 0.800,
 ymax= 0.845,
 xmin = 10,
 xmax = 1250,
label style={font=\Large} ,
 ];

 \addplot  [color=black,
                dash pattern=on 1pt off 3pt on 3pt off 3pt,
                mark=none,
                mark options={solid},
                smooth,
                line width=1.2pt]  file { chapter3-ml_blocks_tuning.txt};

\end{axis}
\end{tikzpicture}
&
\begin{tikzpicture}[scale=0.45]
\begin{axis}[ 
 width=\columnwidth, 
 height=0.65\columnwidth, 
 xmajorgrids, 
 yminorticks=true, 
 ymajorgrids, 
 yminorgrids,
 ylabel={{\mapfive}},
 xlabel = {$B$},
 ymin = 0.754,
 ymax=0.759,
 xmin = 1,
 xmax = 70,
legend style={font=\Large},
label style={font=\Large} ,
tick label style={font=\Large}
 ];

  \addplot  [color=black,
                dotted,
                mark=none,
                mark options={solid},
                smooth,
                line width=1.2pt]  file { chapter3-pandor_map5_blocks}; 
\end{axis}
\end{tikzpicture}\\
(a) \ML-1M & (b) \PANDOR\\
\end{tabular}
\caption[Loss dependence on the number of blocks]{Evolution of {\mapfive} with respect to largest number of allowed blocks, $B$.}
    \label{fig:map5_blocks}
\end{figure}
Table \ref{tab:online_vs_all_ndcg_1h_new} presents full set of results for {\ndcgfive} and {\ndcgten}, and {\mapfive} and {\mapten} of all approaches over the test sets including \RecS{} of the different collections.
\begin{table*}[!ht]
    \centering
     \resizebox{1.0\textwidth}{!}{\begin{tabular}{c|cccccc|cccccc}
    \hline
     &\multicolumn{6}{c|}{\ndcgfive}&\multicolumn{6}{c}{\ndcgten}\\
     \cline{2-13}
     &{\ML-1M}&\Out&\PANDOR&{\NetF}&{\kasandr}&\textcolor{blue}{\RecS}&{\ML-1M}&\Out&\PANDOR&{\NetF}&{\kasandr}&\textcolor{blue}{\RecS}\\
     \hline
     \MostPop  & .090&.011 &.005 &.056 & .002&\textcolor{blue}{.004}& .130&.014 &.008 &.096 &.002&\textcolor{blue}{.007}\\
     \ProdVec  & .758& .232& .078& .712& .012&\textcolor{blue}{.219}& .842& .232&.080 &.770 &.012&\textcolor{blue}{.307} \\
     \MF       & .684&.612 &.300 & .795& .197&\textcolor{blue}{.317}& .805& .684& .303& .834& .219&\textcolor{blue}{.396}\\
    \batch     & .652& .583& .874& .770& .567&\textcolor{blue}{.353}& .784& .658& .890& .849& .616&\textcolor{blue}{.468}\\
     \BPR      & .776& \underline{.671}& .889& \underline{.854}& .603 &\textcolor{blue}{\bf{.575}}&\underline{.863}& \underline{.724}& .905& .903& .650&\textcolor{blue}{\bf{.673}}\\
     {\GRU}$+$ & .721& .633 & .843& .777& .760&\textcolor{blue}{.507}& .833& .680&.862 &.854 &.782&\textcolor{blue}{.613}\\
      {\caser} & .665& .585& .647& .750& .241&\textcolor{blue}{.225}& .787& .658& .666& .834& .276&\textcolor{blue}{.225}\\
     {\SASR}   & .721 & .645& .852& .819& .569&\textcolor{blue}{.509}& .832& .704& .873& .883& .625&\textcolor{blue}{.605}\\
     {\GCN} & \underline{.784} & .652 & \underline{.901} & .836 & \bf{.947} &\textcolor{blue}{.428} & \bf{.874} & .710 & \underline{.915} & .895 & \bf{.954} & \textcolor{blue}{.535}\\
     {$\SO_m$}&.763&.674&.885&.857&.735&\textcolor{blue}{.492}&.858&.726&.899&\underline{.909}&.765&\textcolor{blue}{.603}\\
     {$\SO_b$}     & \bf{.788}& \bf{.710}& \bf{.904}& \bf{.866}& \underline{.791} &\textcolor{blue}{\underline{.563}}& \bf{.874}& \bf{.755}& \bf{.917}& \bf{.914}& \underline{.815}&\textcolor{blue}{\underline{.662}}\\

     \hline
    \end{tabular}
    }
    ~\\~\\     \resizebox{1.0\textwidth}{!}{\begin{tabular}{c|cccccc|cccccc}
    \hline
      &\multicolumn{6}{c|}{\mapfive}&\multicolumn{6}{c}{\mapten}\\
     \cline{2-13}
     &{\ML-1M}&\Out&\PANDOR&{\NetF}&{\kasandr}&\textcolor{blue}{\RecS}&{\ML-1M}&\Out&\PANDOR&{\NetF}&{\kasandr}&\textcolor{blue}{\RecS}\\
     \hline
     \MostPop  & .074&.007 &.003 &.039 & .002&\textcolor{blue}{.003}& .083&.009 &.004 &.051 &.3e-5&\textcolor{blue}{.004} \\
     \ProdVec  & .793& .228& .063& .669& .012&\textcolor{blue}{.210}& .772& .228&.063 &.690 &.012&\textcolor{blue}{.220} \\
     \MF       & .733&.531 &.266 & .793& .170&\textcolor{blue}{.312}& .718& .522& .267& .778& .176&\textcolor{blue}{.306}\\
    \batch     & .713& .477& .685& .764& .473&\textcolor{blue}{.343}& .688& .477& .690& .748& .488&\textcolor{blue}{.356}\\
     \BPR      & 
     .826& .573& .734& .855& .507&\textcolor{blue}{\bf{.578}}&.797& .563& \underline{.760} & \underline{.835}& .521&\textcolor{blue}{\bf{.571}}\\
     {\GRU}$+$ & .777& .513 & .673& .774& .719&\textcolor{blue}{.521}& .750& .509&.677 &.757 &\underline{.720}&\textcolor{blue}{.500}\\
      {\caser} & .718& .471& .522& .749& .186&\textcolor{blue}{.218}& .694& .473& .527& .733& .197&\textcolor{blue}{.218}\\
     {\SASR}   & .776 & .542& .682& .819& .480&\textcolor{blue}{.521}& .751& .534& .687& .799& .495&\textcolor{blue}{.511}\\
    {\GCN} & \bf{.836} & .502 & \bf{.793} & .835 & \bf{.939} & \textcolor{blue}{.428} & \underline{.806} & .507 & \bf{.796} & .817 & \bf{.939} & \textcolor{blue}{.434}\\
     $\SO_m$&.816&\underline{.577}&.720&\underline{.857}&.644&\textcolor{blue}{.495}&.787&\underline{.567}&.723&.837&.651&\textcolor{blue}{.494}\\
     $\SO_b$     & \underline{.832}& \bf{.619}& \underline{.756} & \bf{.866}& \underline{.732}&\textcolor{blue}{\underline{.570}}& \bf{.808}& \bf{.607}& .759& \bf{.846}& .747&\textcolor{blue}{\underline{.561}}\\
     \hline
    \end{tabular}
    }
    \caption[Empirical performance of \MostPop, \ProdVec, \MF, \batch, \BPR{}, {\GRU}$+$, \SASR, \caser, and \SO{}]{Comparison between \MostPop, \ProdVec, \MF, \batch, \BPR{}, {\GRU}$+$, \SASR, \caser, and \SO{} approaches in terms of \ndcgfive{} and \ndcgten (top), and \mapfive{} and \mapten (down). Best performance is in bold and the second best is underlined.}
    \label{tab:online_vs_all_ndcg_1h_new}
\end{table*}

With respect to the updated table \ref{tab:online_vs_all_ndcg_1h_new}, results justify the power of the proposed approach. Even the comparison of $\SO_{}$ with modern \GCN{} shows very promising results. \GCN{} also is trained on the triplets but it sampled negative interactions from all set of items for positive interactions, that's why in case of very imbalanced dataset, such as \kasandr, the model has so big improvement under \SO{} (because when the number of positive interactions is very small we almost for sure will sample negative). But if the data is not so imbalanced with respect to the number of positive/negative interactions, our approach is better because the sampling in \GCN{} for this case will bring the noise to the data.

\paragraph{Identifying stationary users.} We keep only users whose embeddings have four stationary components, using a preliminary estimation of the memory parameter. In table \ref{tab:datasets_filtering} it could be found that the output subset is much smaller for Kassandr and  Pandor than the full dataset whereas for \ML-1M and \Out{} we succeed in keeping a large part of the full dataset. Our filtering approach is then expected to be more successful on the latter. 
\begin{table}[!ht]
    \centering
    \begin{adjustbox}{width=0.4\textwidth,center}
    \begin{tabular}{lcccccc}
    \hline
    Data&$|U|$ & $|Stat\_U|$ \\
    \hline
    Kassandr&2,158,859& 26,308 \\
    Pandor&177,366&9,025\\
    ML-1M&6,040&5,289\\
    Outbrain&49,615&36,388\\
    \hline
    \end{tabular}
    \end{adjustbox}
    \caption{Statistics on datasets before and after filtering. Among these, the remaining number of users after filtering based on stationarity in embeddings is denoted as $|Stat\_U|$}
    \label{tab:datasets_filtering}
\end{table}

\begin{table}[t]
    \centering
    \begin{adjustbox}{width=1.0\textwidth,center}
     \begin{tabular}{c|cccc|cccc}
    \hline
      &\multicolumn{4}{c|}{\mapfive}&\multicolumn{4}{c}{\mapten}\\
     \cline{2-9}
     &{\ML-1M}&{\kasandr}&{Pandor}&\Out &{\ML-1M}&{\kasandr}&{Pandor}&\Out\\
     \hline
    \BPR{}  &.826 & .522 & .702 & .573 & .797 & .538 & .706 & .537\\
    \caser & .718 & .130 & .459 & .393 & .694 & .131 & .464 & .397\\
    \GRU & .777 & .689 & .613 & .477 & .750 & .688 & .618 & .463\\
    \SO  & .832 & .705 & .710 & .600 & .808 & .712 & .714 & .563\\
    \MOSAIC{}  & \textbf{.842} & \textbf{.706} & \textbf{.711} & \textbf{.613} & \textbf{.812} & \textbf{.713} & \textbf{.715} & \textbf{.575}\\
     \hline
    \end{tabular}
    \end{adjustbox}
    
~\\
~\\     
    \begin{adjustbox}{width=1.0\textwidth,center}
    \begin{tabular}{c|cccc|cccc}
    \hline
    \cline{2-9}
     &\multicolumn{4}{c|}{\ndcgfive}&\multicolumn{4}{c}{\ndcgten}\\
     \cline{2-9}
     &{\ML-1M}&{\kasandr}& Pandor&\Out
     &{\ML-1M}&{\kasandr}&Pandor&\Out\\
     \hline
     \BPR{}  & .776 & .597 & .862 & .560 & .863 & .648 & .878 & .663\\
     \caser & .665  & .163 & .584 & .455 & .787 & .198 & .605 & .570\\
     \GRU & .721 & .732 & .776 & .502 & .833 & .753 & .803 & .613\\
    \SO  & .788 & \textbf{.764} & \textbf{.863} & .589 & .874 & \textbf{.794} & .879 & .683\\
    \MOSAIC{} & \textbf{.794} & \textbf{.764} & \textbf{.863} & \textbf{.601} & \textbf{.879} & \textbf{.794} & \textbf{.880} & \textbf{.692}\\
     \hline
    \end{tabular}
    \end{adjustbox}
    
    \caption[Empirical comparison for \mapfive{}, \mapten (top), \ndcgfive{} and \ndcgten{} measures]{Comparison of different models in terms of \mapfive{} and \mapten (top), and \ndcgfive{} and \ndcgten (down). }
    \label{tab:measures_state_of_art}
\end{table}

Table~\ref{tab:measures_state_of_art} presents  the comparison of \MOSAIC{} with \BPR{}, {\caser} and \SO{}. These results suggest that compared to \BPR{} which does not model the sequence of interactions, sequence models behave generally better. Furthermore, compared to {\caser} and \GRU{} which only consider the positive feedback; our approach which takes into account positive interactions with respect to negative ones performs better.

Furthermore, as suspected results on \Out{} and \ML{} are better with \MOSAIC{} than \SO{} in these collections than the two other ones due to the fact that we have more LRD users. Keeping only in the dataset, {\it stationary} users, for which the behavior is consistent with time, is an effective strategy in learning  recommender systems. The predictable nature of the behavior of stationary users makes the sequence of their interactions much exploitable than those of generic users, who may be erratic in their feedback and add noise in the dataset.

\section{Conclusion}\label{sec:Conclusion}

In this chapter, we presented two variants of the \SO{} approach presented in chapter \ref{ch:2}; in the first model parameters are updated user per user over blocks of items constituted by a sequence of unclicked items followed by a clicked one. The parameter updates are discarded for users who interact very little or a lot with the system. The second variant, is based on the momentum technique as a means of damping oscillations.  The second contribution is a theoretical analysis of the proposed approach which bounds the deviation of the ranking loss concerning the sequence of weights found by both variants of the algorithm and its minimum in the general case of non-convex ranking loss.  Empirical results conducted on six real-life implicit feedback datasets support our founding and show that the proposed approach is significantly faster than the common batch and online optimization strategies that consist in updating the parameters over the whole set of users at each epoch, or after sampling random pairs of preferred and no-preferred items. The approach is also shown to be highly competitive concerning state of the art approaches on \texttt{MAP} and \texttt{NDCG} measures.

In addition, we introduced a strategy to filter the dataset with respect to homogeneity of the behavior in the users when interacting with the system, based on the concept of memory. From our results, it comes out that taking into account the memory in the case where the collection exhibits long range dependency allows to enhance the predictions of the proposed sequential model. As future work, we propose to encompass the analysis of LRD and the filtering phase in the training process.

\chapter{Faulted Lines Detection with ranking-based approach}
\label{ch:5}
\section*{Introduction}
\addcontentsline{toc}{section}{Introduction}



The climate change and global warming results in an increased number of extreme weather events \cite{sillmann2008indices} that compromises security and reliability of critical infrastructure (power and gas grids, telecommunications, transportation systems) \cite{birkmann2016extreme}. According to the recent statistics of the National Center for Environmental Information\footnote{\url{https://www.ncdc.noaa.gov/billions/}}, the total cost of 310 recent major weather events exceeds \$2.155 trillion dollars and projected to increase in the near future~\cite{smith2013us}. Power grids are responsible for a substantial part of this~cost~\cite{stern2007economics}. 

One of the major challenges in protecting a grid from impending a cascading blackout after a line failure is \textcolor{red}{a} real-time localization of the faulted line followed activating emergency controls \cite{begovic2005wide, zhang2016remedial}. Traditional data-driven methods for fault localization, such as travelling-wave \cite{parsi2020wavelet} and impedance based ones \cite{aucoin1996high}, require high grid observability and sampling rates that are technically challenging and expensive for bulky systems~\cite{sundararajan2019survey} or even known distribution of renewables~\cite{owen2019importance,lukashevich2021importance,lukashevich2021power}. Another line of algorithms leverages deep neural networks capabilities \cite{li2019real,li2021physicsA,zhang2020novel,misyris2020physics}; however, these methods suffer from high requirements on the amount of phasor-measurement unit data. The latter lead to inability to make a accurate and timely detection in time-changing environment that is intrinsic for extreme weather events and, therefore, compromises power grid security.

The chapter addresses power grid reliability during extreme events, such as wildfires, hurricanes and extreme winds, when multiple line failures may happen. The latter failures must be detected in real-time to preserve secure and reliable operations and prevent the grid from impending energy blackout. The most common failure type is a line failure, when power supply through a specific line (lines) is interrupted for a few seconds or permanently. 

\textbf{Contribution.} Our contribution is as follows. First, we propose Topology-Aware failure Localization Detector (TALD), a neural-network based algorithm for detecting line faults in real-time. A particular advantage of our approach, that lead to a higher detection accuracy and lower data requirements, is leveraging grid topology information.

Second, the algorithm estimates the conditional probability that the fault has happened on this line. This allows not only estimate the detection confidence, but also efficiently utilize prior information on line vulnerability. The latter is often accessible for power grid operators as a result of earlier failures or maintenance information. 

Finally, we provide empirical support for TALD showing its superior performance over simulated data.



{\bf Paper structure.} The paper is organized as follows. Section \ref{ch5:problem} contains problem setup and provides necessary background information. Section \ref{ch5:exp} provides empirical results and discussion about it's role. Short conclusion is given in Section \ref{ch5:concl}. 
\section{Problem Statement}\label{ch5:problem}

\subsection{Notation.}
Let $E$, $|E| = m$, be a set of lines and $V$, $|V| = n$, is a set of buses in a power grid $G = \langle V, E\rangle$. Let $p,q \in\mathbb{R}^n$ be vectors of active and reactive power, $v\in\mathbb{R}^n$ be a vector of voltage magnitudes, and $\theta \in \mathbb{R}^n$ be a vector of voltage phases. We denote phase angle differences as $\theta_{ij}$, $(i,j)\in E$. Power grid buses consist of PQ (load) buses, PV (generation) buses, and a slack bus that often stands for the largest and slowest generator in the grid. We assume below that the phase angle $\theta_i = 0$ for the slack bus $i$.

\begin{table}[th!]
    \centering
    \begin{adjustbox}{width=\linewidth}
    \begin{tabular}{l|l|l|l}
         $E$ & set of lines & $V$ & set of buses\\
         $m$ & number of lines& $n$ & number of buses\\
         $v$ & bus voltages, $v\in \mathbb{R}^n$ & $\theta$ & phase angles, $\theta\in\mathbb{R}^n$\\
         $p,q $ & \multicolumn{3}{l}{vector of active/reactive power injections} \\
         $d$ & number of PMUs & $V^d$ & set of nodes with PMUs\\
         $t$ & time index & & \\
         $y_{i}^t$ & \multicolumn{3}{l}{failure indicator at time $t$ at line $i$}\\
         $x^t = (\{\theta^t_i, v^t_i\}_{i=1}^d)$ & \multicolumn{3}{l}{a set of PMU measurements at time $t$} \\
         $\nb_E(\cdot), \nb_V(\cdot)$ & \multicolumn{3}{l}{list of adjacent edges, vertices}\\
         $\nb^k_E(\cdot), \nb^k_V(\cdot)$ & \multicolumn{3}{l}{$\nb^k_E(\cdot) = \underbrace{\nb_E((\dots \nb_E(\cdot))}_{k \text{ times}}, \nb^k_V(\cdot) = \underbrace{\nb_V(\dots \nb_V(\cdot))}_{k \text{ times}}$} \\
    \end{tabular}
    \end{adjustbox}
    \caption[Chapter notation]{Chapter notation.}
    \label{tab:notation}
\end{table}
The chapter notation is summarized in Table~\ref{tab:notation}. 

\subsection{Background.} Phasor Measurement Units (PMUs) enable high-resolution situational awareness of power grid state by providing information about voltage magnitude $v_i$, $i\in V$ and phase angle $\theta_{ij}$, $(i,j)\in E$ using a common time source for synchronization. Often PMUs are required at tap-changing transformers, complex loads, and PV (generation) buses. Despite of wide-spread of PMUs and their role in grid monitoring, power grids remain covered only in part because of privacy and budget limitations. 

For notation simplicity, we assume w.l.o.g. that PMUs are places at the first $d$ buses $V_d$ of the grid, $V_s\subseteq V$, and this placement does not change during the observation time. We refer $V_d$ as a set of observable buses. Furthermore, we receive a set of PMUs measurements $x^t = (\{\theta_i^t, v_i^t\}_{i=1}^d)$ for each time~$t$, $0\le t \le T$. Let $y^t \in \mathbb{R}^n$ be a an indicator of faulted lines, e.g. $y_{ij}^t = 1$ iff line $(i,j)$ is faulted at time $t$, $0\le t \le T$. 

The ability of PMU to measure the voltage phasor at the installed bus and the current phasor of all the branches connected to the PMU installed bus can help determine the remaining parameters to use for indirect measurements.

A particular advantage of PMU technology is high sampling rate that dramatically increase situational awareness and allows to detect grid failures in nearly real-time. For instance, for 60 Hz systems, PMUs must deliver between 10 and 30 synchronous reports per second depending on the application.

\begin{table}[th]
    \centering
    \begin{tabular}{l|l}
        Event & Time, sec. \\
        \hline\hline
         {\bf Transient Voltage Stability}& 0.2 -- 10 \\
         \hline 
         Line trip & 0.1 -- 1.5\\
         Static VAR Compensator (SVC) & 0.1 -- 1\\
         DC compensator & 0.1 -- 1\\
         Generator Inertial Dynamics & 0.5--5\\
         Undervoltage Load Shedding & 1--9\\
         Mechanically Switched Capacitors Dynamics & 0.15--2\\
         Generator/Excitation Dynamics & 0.15--3\\
         Induction Motor Dynamics & 0.1--2\\
         DC Converter LTCs & 4--20\\
         \hline
         {\bf Long-term Voltage Stability} & 20 -- 10000 \\
         \hline 
         Protective Relaying Including Overload Protection & 0.1 -- 1000\\
         Prime Mover Control & 1--100\\
         Auto-Reclosing & 15--150\\
         Excitation Limiting & 9--125\\
         Boiler Dynamics & 20--300\\
         Generator Change/AGC & 20--800\\
         Power Plant Operator & 40--1000\\
         Load Tap Changers and Dist, Voltage Reg. & 20--200\\
         System Operator & 60--10000\\
         RAS & 150--300\\
         RAP & 350--1000\\
         Gas Turbine Start-Up & 250--900\\
         Load Diversity/Thermostat & 200--2000\\
         Line/Transformer Overload & 600--2500\\
         Load/Power Transfer Increase & 250--7000 \\
         \hline
    \end{tabular}
    \caption{Timeline of events in a power grid}
    \label{tab:my_label}
\end{table}

To consider the topology of power grid we transform the binarized targets (fault or non-fault), that we used during training into two vectors: the first one includes the information about the faulted line and the second one consists of the information about the neighbours of the faulted line. In more details, suggest we have a sample $(x,y)$ with the features $x \in R^d$. Then the first vector of targets is defined as $y = [y_1, \cdots,y_i, \cdots, y_n]^T \in R^n$, where in case of faulted line at the location $j$, $y_j =1$ and $y_i=0$ for $i!=j$. For the second vector of targets, let $nb(j)$ denote the neighborhood of the $j$th line, including the lines connected with $j$, and then $\hat{y}_i = 1/\text{nb}(j)$ only if $i \in \text{nb}(j)$. The definition of $\hat{y}$ is formalized at the equation defined at \ref{eq:multilabel_targets}:

\begin{equation}
\label{eq:multilabel_targets}
\hat{y}(i) = 
\begin{cases}
1/|\text{nb}(j)|, & \text{if i} \in \text{nb}(j): \text{neighbor set of $j$} \\
0, & \text{else}\\
0 & i = j$~$ \text{the true location has weight 0}
\end{cases}
\end{equation}

For the remains line target is equal to zero. Then the loss function $Loss(f(x),y,\hat{y})$ for the proposed model (architecture is presented on the Fig. \ref{fig:energy_arch}) over the samples $(x,y)$, where $f(x)$ are the predicting probabilities of the proposed model is defined as the sum of two terms of cross-entropy functions (here CE). The definition of CE is given below at the Eq. \ref{eq:cross entropy}:

\begin{equation}
CE(y,f(x)) = \sum_{i=1}^{n}y_i\cdot\log\left(\frac{\exp^{f_i(x)}}{\sum_{i=1}^{n}\exp^{f_{i}(x)}}\right)
\label{eq:cross entropy}
\end{equation}

\begin{figure}[H]
\centering
\vspace*{-10pt}
\includegraphics[width=0.7\textwidth]{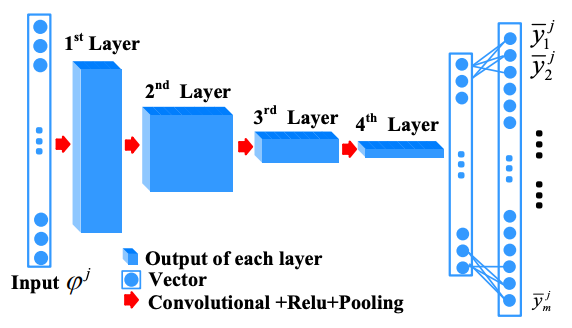}
\caption[Convolutional Network Architecture]{Architecture of the applied model proposed in \cite{DBLP:journals/corr/abs-1810-05247}}
\label{fig:energy_arch}
\end{figure}

Then, we could express the loss function more formally by the next equation \ref{eq:loss}:

\begin{equation}
Loss(y,\hat{y},f(x)) = CE(y,f(x))\cdot(1-\epsilon)+CE(\hat{y},f(x))\cdot \epsilon
\label{eq:loss}
\end{equation}

The architecture of the baseline model presented on the Fig. \ref{fig:energy_arch} is described in details in the paper of authors \cite{DBLP:journals/corr/abs-1810-05247}. It's suggested to use the convolution-based neural network with the information about the bus voltages and prepared features with a physical interpretation to make the predictions about fault location. To make the model more interpretable and to improve the output accuracy we modified the loss function to the explained in the equation \ref{eq:loss} by including the network topology in the model and then provide empirical evaluation of both approaches presented in section \ref{ch5:exp}.

\section{Experimental part}\label{ch5:exp}

\subsection{Dataset}
To estimate the approaches we apply two benchmarks: SIM\_LARGE and SIM\_SMALL. SIM\_SMALL was provided us by authors of \cite{DBLP:journals/corr/abs-1810-05247} for 68-bus power system.
The second dataset SIM\_LARGE we simulated in the power system toolbox, based on nonlinear models \cite{PST}, a three-phase
short circuit fault lasting 0.2 seconds at the line 5-6 also in the IEEE 68-bus power system as in the SIM\_SMALL. The main differences between two benchmarks are the number of samples simulated for train, test and validation sets, where the new simulated set is about 10 times bigger. The second point is that the test set for SIM\_LARGE is generated simultaneously for all fault types, as the train set for both datasets, whereas in SIM\_SMALL benchmark there are separate test sets for each fault. This new simulation allows us to estimate the generalization property of the model to distinguish between different fault types. Also it let us to avoid the overfitting of the model on one particular class. 

The feature vector $\psi$ is computed based on the idea lies in the baseline approach \cite{DBLP:journals/corr/abs-1810-05247}. Represented by the feature vectors faulted lines in power grid are then labeled by their locations; in case of $m$ lines in the power grid, the number of output classes are equal to $m+1$, where additional class is for the normal condition, that means there are no faults in the system. Below the statistics regarding the size of simulated data for train, test and validation evaluations are represented in the table \ref{tab:sim_grid_info}:

\begin{table}[ht!]
    \centering
    \begin{tabular}{ccc}
    \hline
    Dataset & Set & Size\\
    \hline
    \multirow{6}{*}{SIM\_SMALL} & Train & 1210\\
    & \begin{tabular}{c}
    \hline
  TP - Test \\ 
  DLG - Test \\ 
  LG - Test \\ 
  LL - Test \\ \hline
\end{tabular} & \begin{tabular}{c}
    \hline
  71 \\ 
  71 \\ 
  70 \\
  71 \\ \hline
\end{tabular}\\
    & Validation & 1210\\
    \hline
   \multirow{3}{*}{SIM\_LARGE} & Train & 14413 \\
    & Test & 994\\
    & Validation & 1207\\
    \hline
    \end{tabular}
    \caption{Size of the train, test and validation parts.}
    \label{tab:sim_grid_info}
\end{table}


The fault-cases provided in data are simulated by changing the line impedance, depending on the type. For simulation we consider a power grid of $n$ buses with a single line fault that may either be one of the following: three-phase short circuit (TP), line to ground (LG), double line to ground (DLG) and line to line (LL) faults for SIM\_SMALL and LG, DLG and LL for SIM\_LARGE. To characterize the location of the faults in power grid, the authors of \cite{DBLP:journals/corr/abs-1810-05247} propose to apply the substitution theory \cite{J.Quanyuan_2014} for deriving the equations related to pre- and during-fault system variables to express feature vector. The feature vector $\psi \in C^{n\times 1}$ based on the substitution theory is defined then in terms of the bus voltages variations $\Delta U$ before and during the faults and the admittance matrix $Y_0$ before the faults:

\begin{equation}
   \psi=\Delta U\cdot Y_{0}
\end{equation}

To understand the distribution of generated data, we provide the statistics with respect to the size of groups regarding the number of neighbours over lines. The results of the calculated statistics are introduced in the Fig. \ref{fig:data_distr}:

\begin{figure}[!h]
   \begin{minipage}{0.48\textwidth}
     \centering
     \includegraphics[width=1.0\linewidth]{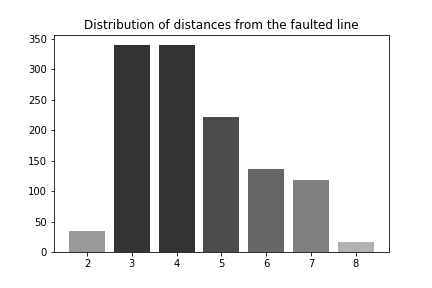}
     \caption[Dataset SIM\_SMALL]{SIM\_SMALL}
   \end{minipage}\hfill
   \begin{minipage}{0.48\textwidth}
     \centering
     \includegraphics[width=1.0\linewidth]{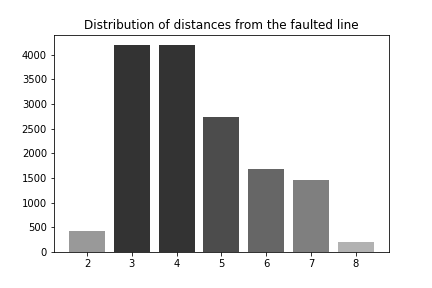}
     \caption[Dataset SIM\_LARGE]{SIM\_LARGE}
   \end{minipage}
\caption{Number of objects for corresponding group with the amount of neighbours for faulted line}
\label{fig:data_distr}
\end{figure}


\subsection{Signal to Noise Ratio}

SNR (signal-to-noise ratio) is a measure used in science and engineering that compares the level of a desired signal to the level of background noise. SNR is defined as the ratio between the output power of the transmitted signal and the power of the noise that distorts it.

\begin{equation}
    SNR = \frac{P_{signal}}{P_{noise}}=\frac{A^{2}_{signal}}{A^{2}_{noise}}
\end{equation}

$P$ here means average power and $A$ is mean-square amplitude. Because many signals have a very wide dynamic range, signals are often expressed using the logarithmic decibel scale.  Then SNR ratio is expressed in decibels(dB) is transformed into the form:

\begin{equation}
    SNR_{dB} = 10\log_{10}\frac{P_{signal}}{P_{noise}}= 20\log_{10}\frac{A_{signal}}{A_{noise}}
\end{equation}


The ratio of SNR can take zero, positive or negative values. An SNR over 0 dB indicates that the signal level is greater than the noise level. The higher the ratio, the better the signal quality. The SNR of PMU measurements in different
regions can vary. We additionally explore this parameter over the test evaluations in subsection \ref{subsect:exp} of present chapter.

\subsection{Empirical Evaluation}
\label{subsect:exp}
The proposed model was trained using RMSProp optimizer and for early-stopping criteria was suggested the next one: validation loss is computed over all validation data, then if $\min$ over last 100 validation losses $<$ best loss, where the best loss is the minimum between the current best loss and the average over the last validation losses for 100 steps, then we continue to train, otherwise - stop. All the parameters such as learning rate, batch size, ratio that responsible for how many information about the neighbours we take during the training and the remains parameters are set using cross-validation method. To estimate the model we apply accuracy measure that is defined as the relation between the number of correctly detected faulted lines and total number of faults. 
The first experiments are done on the small SIM\_SMALL dataset over the full and partial observability cases. The partial measures are range between $15\%$ and $30\%$ of buses and estimated over 4 test sets for each particular fault class. The analysis of the results for two models could be find in the table \ref{tab: small_data_partial_measures}.

\begin{table}[h!]
    \centering
    \resizebox{1.0\columnwidth}{!}{
    \begin{tabular}{ccc|cc|cc|cc}
    \\
    \hline
    \multicolumn{1}{c}{} & \multicolumn{2}{c}{TP fault} & \multicolumn{2}{c}{DLG fault} & \multicolumn{2}{c}{LG fault}& \multicolumn{2}{c}{LL fault}
    \\
    \hline
    \% buses & No-neigbhbors & With-neighbors & No-neigbhbors & With-neighbors& No-neigbhbors & With-neighbors& No-neigbhbors & With-neighbors  \\
    \hline
100 & 98.59 & \bf{100.0} & \bf{100.0} & \bf{100.0} & \bf{100.0} & \bf{100.0} & \bf{100.0} & \bf{100.0}\\
30 & 91.55 & \bf{97.18} & 95.77 & \bf{98.59} & \bf{97.14} & \bf{97.14} & 98.59 & \bf{100.0}\\
25 & 78.87 & \bf{92.96} & 92.96 & \bf{97.18} & 94.29 & \bf{97.14} & 95.77 & \bf{98.59}\\
20 & 91.55 & \bf{94.36} & 90.14 & \bf{97.18} & 84.29 & \bf{94.29} & \bf{95.77} & \bf{95.77}\\
15 & 73.24 & \bf{88.73} & 95.77 & \bf{97.18} & 88.57 & \bf{92.86} & 88.73 & \bf{90.14}\\
\end{tabular}}
\caption{Comparison of the approaches based on partial observability without noise, SIM\_SMALL data}
\label{tab: small_data_partial_measures}
\end{table}

Based on the experiments it could be said that more measured buses improve the predictability of fault locations. Also it should be noticed that information about the grid topology also improve the final results on $2\%$-$18\%$ in comparison the case without taking into account the neighbours during training in the loss function.

The results of the estimation the generalisation property to distinguish the faults over different types are done on SIM\_LARGE dataset and presented on the table \ref{tab:SIM_BIG_diff_samples} for range of train samples between $10$ and $100$ percentages with step $10$. For the most cases we could see the profit for the model with neighbours topology. This results support the property of the generalisation the fault classes.

\begin{table}[htb!]
    \centering
    \resizebox{1.0\columnwidth}{!}{
    \begin{tabular}{ccccccccccc}
    \\
    \hline
    +/- neighbors & 100 \% & 90\% & 80 \% & 70 \% & 60 \% & 50 \% & 40 \% & 30 \% & 20 \% & 10 \% \\
    \hline
 no neighbors & 95.07 & 93.86 & 93.66 & 92.76 & \bf{95.07} & 88.33 & \bf{91.44} & 93.16 & \bf{89.03} &  85.11\\
 with neighbors & \bf{95.57} & \bf{95.27} & \bf{95.07} & \bf{95.47} & 94.67 & \bf{91.95} & 90.74 & \bf{94.16} & 88.63 & \bf{89.64}  \\
\end{tabular}}
\caption{Estimation for different sizes of training set on SIM\_LARGE data}
\label{tab:SIM_BIG_diff_samples}
\end{table}

For SIM\_LARGE data we also provide the experiments for partial bus observations as it was done for SIM\_SMALL. The results are presented in the table \ref{tab:sim_big_partial_measures}. What could be seen from here is that as in the table \ref{tab: small_data_partial_measures}, the tendency between ratio of measured buses and accuracy is preserved; it means that more observations usually provide more accurate predictions of faulted line locations that could be explained by the bigger amount of input information provided for the model. Regarding the comparison between two models, for all ratio values topology-based model outperform the second one, that proves it's less sensitivity to the lack of information, where for the topology-based model the result sank around on $9\%$, where for another model it sank on $13\%$.

\begin{table}[h!]
    \centering
    \resizebox{0.5\columnwidth}{!}{
    \begin{tabular}{ccc}
    \\
    \hline
    \multicolumn{3}{c}{LG fault}
    \\
    \hline
    \% buses & No-neigbhbors & With-neighbors\\
    \hline
100 & 95.07 & \bf{95.57}\\
30 & 85.41 & \bf{87.32}\\
25 & 78.27 & \bf{82.09}\\
20 & 79.48 & \bf{80.28}\\
15 & 71.93 & \bf{76.25}\\
\end{tabular}}
\caption{Comparison of the approaches based on partial observability on SIM\_LARGE dataset}
\label{tab:sim_big_partial_measures}
\end{table}

The test evaluations over SNR parameter are done by ranging approximation value of noise from 40dB to 100dB with the step size 10. The Gaussian noise of the same SNR was added both to the training and testing parts of datasets. The structure of the CNN was kept the same but the hyperparameter as ratio $\epsilon$ in Eq. \ref{eq:loss} was additionally set up in
the noisy regime. Other parameters are the same. Results in Fig. \ref{fig:snr_estimation} indicate that the sensitivity
of both models to noise is different, and that model based on neighbours topology is relatively more robust to the noise.

\begin{figure}[H]
\centering
\vspace*{-10pt}
\includegraphics[width=0.75\textwidth]{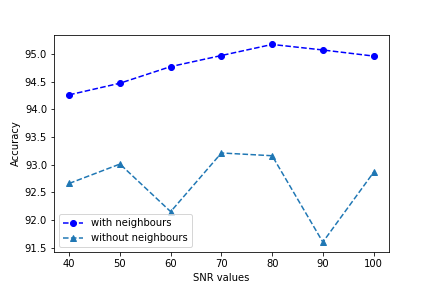}
\caption[SNR for various datasets]{Estimation of SNR approximation over the approaches on SIM\_LARGE dataset. The results are provided for both models: with and without neighbours topology term in loss function}
\label{fig:snr_estimation}
\end{figure}




\subsection{U-Mann-Whitney Test}

Because of some instability in results from table \ref{tab:SIM_BIG_diff_samples}, we suggest to compare them on the basis of statistical criterion of Mann-Whitney-Wilcoxon. The distributions of the output samples of accuracy's are illustrated on the Fig. \ref{fig:samples_dist_mannwhitney}. This U-criterion is used to assess the differences between two independent samples by the level of a feature measured quantitatively. 

\begin{figure}[H]
\centering
\vspace*{-10pt}
\includegraphics[width=0.75\textwidth]{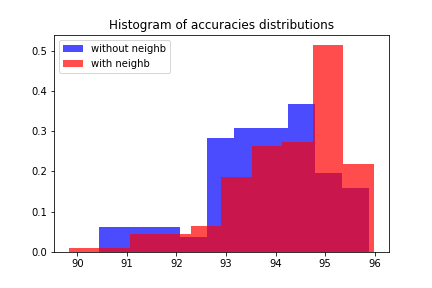}
\caption[Normalized histogram for samples distributions]{Normalized histogram for samples distributions. The histogram provide information about output accuracy for two models for different randomization.}
\label{fig:samples_dist_mannwhitney}
\end{figure}

This method determines whether the zone of overlapping values between two rows is small enough. The lower the value of the criterion, it is more likely that the differences between the parameter values in the samples are significant. U-Mann-Whitney test step-by-step:
\begin{itemize}
    \item To make a single ranked series from both compared samples, placing their elements according to the degree of increase and assigning a lower rank to a lower value with number of elements in the first sample $n_1$ and $n_2$ in the second one.
    \item Divide a single ranked series into two, consisting of units of the first and second samples, respectively. Calculate separately the sum of ranks for each sample $R_1$ and $R_2$, then calculate:
    \begin{equation}
        U_1 = n_1 \cdot n_2 + \frac{n_1 \cdot (n_1+1)}{2}-R_1
    \end{equation}
    
    \begin{equation}
        U_2 = n_1 \cdot n_2 + \frac{n_2 \cdot (n_2+1)}{2}-R_2
    \end{equation}
    \item Determine the value of the Mann-Whitney U-statistics by the formula $U=\max\{U_{1},U_{2}\}$.
    \item Using the table for the selected level of statistical significance, determine the critical value of the criterion for the data. If the resulting value of $U$ is greater than or equal to the tabular one, then it is recognized that there is a significant difference between the samples and an alternative hypothesis is accepted. If the resulting value of $U$ is less than the table value, the null hypothesis is accepted.
\end{itemize}

In our case, as null hypothesis we consider the equivalence of the mean for both samples, as alternative hypothesis we suggest that the mean of the model that takes into accounts the neighbours topology is greater than of the second one. The statistical significance, also denoted as $\alpha$, is the threshold probability of rejecting the null hypothesis when it is true. $p_{value}$ - is actual probability (calculated from the resulting value of $U$) of rejecting the null hypothesis when it is true. So when $p_{value} < \alpha$, we assume that we reject the null hypothesis correctly.

The result of Mann-Whitney statistical test is presented on the Fig. \ref{fig:mannwhitney_res}. For the comparison, a critical region of $2\sigma$ is given. The Fig. \ref{fig:mannwhitney_res} shows that the value of p-value is significantly less than alpha, so, we reject the null hypothesis. Therefore, the mean of the model that takes into account the topology among neighbour lines in power grid exceeds the mean of the baseline model, and then we could consider obtained results as statistically significant.

For power systems operational practice, it might be beneficial to present the solution in a simple logical form~\cite{boros2000implementation,hammer2006logical,maximov2012comparative,maximov2013implementation,maximov2012simple} conventional for interpretation by a power system operator. 

\begin{figure}[H]
\centering
\vspace*{-10pt}
\includegraphics[width=0.7\textwidth]{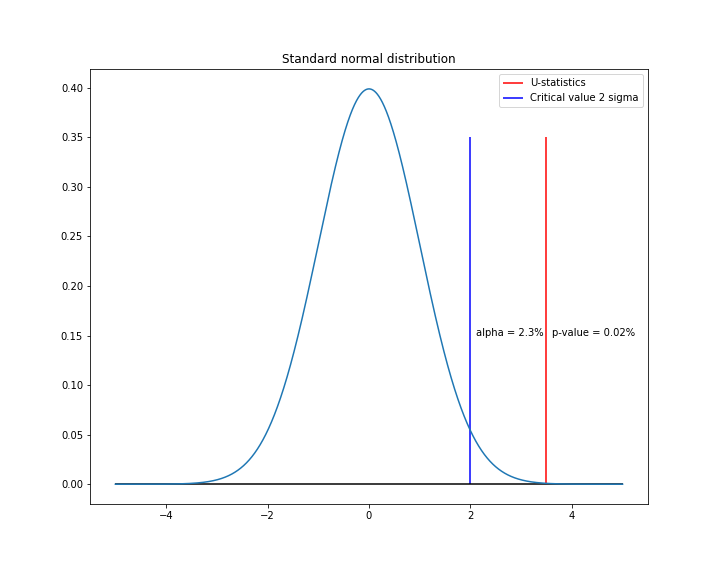}
\caption[The Mann-Whitney Statistics]{Computed Mann-Whitney Statistics for normal distribution.}
\label{fig:mannwhitney_res}
\end{figure}


\section{Conclusion}\label{ch5:concl}

The problem of predicting power grid faults with a convolutional neural network is discussed in this chapter. Simulated datasets SIM\_SMALL and SIM\_BIG containing four and three types of errors, respectively, were used to address the problem. By improving the loss function of the previously presented model \cite{DBLP:journals/corr/abs-1810-05247}, we were able to achieve minor gains in the accuracy measure. We added an additional term accounting neighbor information to the loss function to account for the neighbors of the line with a failure throughout the learning phase. To evaluate the statistical significance of the suggested technique, we used a statistical Mann-Whitney test to corroborate our findings. The test validated the approach's static significance. Also the modified model demonstrates it is better robustness to the noise conditions and partial observability. A similar approach can be used for analysis of power generation reliability~\cite{stulov2020learning,mikhalev2020bayesian}.

\chapter{Conclusion and Future Perspectives}\label{ch:6}
\chaptermark{Conclusion}

\section{Concluding remarks}
In this thesis, the problem of ranking was considered in relation to different fields, in particular, to recommender systems and power systems. 
The first and the main part of the thesis is devoted to recommender systems. As a solution to the problem regarding effective recommendations in the case of implicit feedback, we propose an approach \SO{} \citepers{Burashnikova19} for effective sequential training of hidden representations of users and objects that take into account their interactions in the system. The approach uses both types of feedback, positive and negative, organized in blocks so that, according to our assumption, the algorithm pays more attention to positive interactions during training. The proposed algorithm has proven itself in practice relative to other popular approaches, where the most of them use only one type of feedback. This confirmed the importance of using both types of interaction's output, as well as the effectiveness of the proposed block structure of training. And also provided ways to improve it and speed it up. It is important to note that the proposed algorithm was theoretically justified. First we proved its convergence for the case of a convex loss function, and then we extend the theory of convergence on the general case \citepers{DBLP:journals/jair/BurashnikovaMCL21}. 

We also suggest the ways to speed up the algorithm in conditions of preserving the quality of predictions \citepers{DBLP:journals/corr/abs-2112-02242}. The proposed method considers the time series of user interactions with objects and filters out those objects that do not keep the long memory. The experiments showed that the effective part of the training data keeps memory. Thus, by filtering out users with short memory, we have preserved the high quality of the model, slashed the size of the input data, and reduced the time for processing and training them.

As for the second part about the application of ranking problems in engineering systems, the existing model for ranking the faulted lines in power grid was considered. In order to improve it, we proposed to take into account the network topology and changed the loss function by adding the term that takes into account the neighbors of the broken line. Unfortunately, we didn't appreciably improve the results, but according to the Mann-Whitney statistical test, our results are statistically significant.

\section{Future perspectives}

\paragraph{Recommender Systems.}

Generally speaking, it is a quite difficult task to adapt recommendations to different data. Everything that we proposed to do on open-source data may not work in reality. This may be a problem from two sides. The first reason are users, as because their behavior may not be the same as that of those users on whom we configured the algorithm. Another side is time: even if we recommend good objects, then users will see and click only those objects that we recommend to them and it will be more profitable for the company to recommend the same objects (feedback loop task, \cite{10.1145/3340531.3412152,NIPS2016_962e56a8}). Therefore, it would be interesting to try to run the proposed algorithm on some real data and adapt the approach to it.

\paragraph{Power Systems.}

As for power grids, it would be interesting to try graph neural networks for more advanced work with the network topology. Recently, graph neural networks have shown impressive results in the power systems tasks \cite{DBLP:journals/corr/abs-2101-10025} due to their ability to capture dependencies in the graph-structured systems.

\begin{singlespace}
\nocite*
\bibliographystylepers{habbrvyr}
\bibliographypers{personal}
\bibliographystyle{apalike}
\bibliography{main}
\end{singlespace}

\end{document}